\documentclass{article}
\usepackage{booktabs}
\usepackage{amsfonts}
\usepackage{multirow}
\usepackage{titletoc}
\usepackage{wrapfig}
\usepackage{algorithm}
\usepackage{algpseudocode}
\usepackage[preprint]{corl_2026} %

\title{Inference-time Policy Steering via Vision and Touch}

\author{
  Yilin Wu\\
  Carnegie Mellon University\\
\texttt{yilinwu@andrew.cmu.edu} \And 
Zilin Si \\
 Carnegie Mellon University\\
\texttt{zsi@andrew.cmu.edu}
\And 
Zeynep Temel \\
 Carnegie Mellon University\\
\texttt{ztemel@andrew.cmu.edu}
\And 
Oliver Kroemer \\
 Carnegie Mellon University\\
\texttt{okroemer@andrew.cmu.edu}
\And 
Andrea Bajcsy \\
 Carnegie Mellon University\\
\texttt{abajcsy@andrew.cmu.edu}
}

\usepackage{xspace}
\usepackage{bbm}
\usepackage{graphicx}
\usepackage{subfig} 
\usepackage{annotate-equations}
\definecolor{basepolicy}{HTML}{7A6E5D}
\definecolor{visualc}{HTML}{4F93D2}
\definecolor{tactilec}{HTML}{5EAA5E}
\definecolor{oursc}{HTML}{E98A4E}

\definecolor{dark_red}{RGB}{122, 0, 0}
\definecolor{coral}{RGB}{255, 119, 94}
\definecolor{pink_orange}{RGB}{255, 72, 126}
\definecolor{vibrant_pink}{RGB}{255, 0, 104}
\definecolor{pink_pink}{RGB}{255, 37, 153}
\definecolor{wine}{RGB}{204, 0, 102}

\definecolor{light_orange}{RGB}{255, 198, 107}
\definecolor{orange(sae/ece)}{rgb}{1.0, 0.49, 0.0}
\definecolor{dark_orange}{RGB}{216,92,0}

\definecolor{org-purp-0}{RGB}{165, 76, 0}
\definecolor{org-purp-1}{RGB}{250, 130, 28}
\definecolor{org-purp-2}{RGB}{226, 89, 68}
\definecolor{org-purp-3}{RGB}{206, 92, 124}
\definecolor{org-purp-4}{RGB}{116, 80, 146}
\definecolor{org-purp-5}{RGB}{110, 78, 157}

\definecolor{teal(sae/ece)}{rgb}{0, 0.47, 0.52}
\definecolor{aqua}{RGB}{52,172,139}
\definecolor{dark_aqua}{RGB}{35,115,93}
\definecolor{dark_green}{RGB}{0, 92, 34}
\definecolor{solid_green}{RGB}{39, 163, 50}

\definecolor{grape}{RGB}{112,48,160}
\definecolor{purple}{rgb}{0.74, 0.65, 1.0}
\definecolor{dark_purple}{rgb}{0.58, 0.0, 0.82}
\definecolor{periwinkle}{RGB}{191, 140, 230}

\definecolor{light_gray}{rgb}{0.9, 0.9, 0.9}
\definecolor{medium_gray}{rgb}{0.6, 0.6, 0.6} 
\definecolor{dark_gray}{rgb}{0.2, 0.2, 0.2} 

\definecolor{sky_blue}{RGB}{37, 166, 213}
\definecolor{light_blue}{rgb}{0.33, 0.80, 1}
\definecolor{dark_blue}{rgb}{0.098, 0.239, 0.52}
\definecolor{ocean}{RGB}{13, 121, 202}
\definecolor{light_ocean}{RGB}{18, 178, 235}
\definecolor{dark_ocean}{RGB}{10, 89, 148}
\definecolor{vibrant_blue}{RGB}{14, 120, 255}

\definecolor{dark_brown}{rgb}{0.3255, 0.004, 0.001}

\newcommand{\para}[1]{\textbf{#1. }} 
\newcommand{\paras}[1]{\smallskip\noindent\textbf{#1. }} 

\newcounter{qnum}
\newcounter{tnum}
\newcommand{\ours}{\textcolor{black}{\textbf{ViTaL}}\xspace}

\newcommand{\obs}{o}
\newcommand{\obsv}{o^v}
\newcommand{\obst}{o^\tau}

\newcommand{\proprio}{q}
\newcommand{\propriotraj}{\mathbf{q}}

\newcommand{\obstraj}{\mathbf{o}}
\newcommand{\obsvtraj}{\mathbf{o}^v}
\newcommand{\obsttraj}{\mathbf{o}^\tau}
\newcommand{\obsSpace}{\mathcal{O}}

\newcommand{\enc}{\mathcal{E}_\psi}
\newcommand{\dec}{\mathcal{D}_\varphi}

\newcommand{\worldmodel}{\mathcal{P}}

\newcommand{\encv}{\enc^{v}}
\newcommand{\enct}{\enc^{\tau}}
\newcommand{\decv}{\dec^v}
\newcommand{\dect}{\dec^\tau}

\newcommand{\task}{\mathcal{L}}
\newcommand{\lang}{\ell}

\newcommand{\latent}{z}
\newcommand{\latentv}{z^{v}}
\newcommand{\latentt}{z^{\tau}}
\newcommand{\latenttraj}{\mathbf{z}}

\newcommand{\latenttrajtpred}{\mathbf{\hat{z}}^{\tau}}
\newcommand{\latenttrajvpred}{\mathbf{\hat{z}}^{v}}
\newcommand{\latentSpace}{\mathcal{Z}}

\newcommand{\policy}{\pi_{\theta}}
\newcommand{\action}{a}
\newcommand{\acttraj}{\mathbf{\action}}
\newcommand{\actchunk}{a_{t:t+h}}

\usepackage{amsmath}

\begin{document}
\maketitle

\begin{abstract}
Inference-time steering adapts pre-trained generative robot policies during deployment by verifying candidate actions before execution. While prior methods typically perform this verification only with visual observations, vision alone is often insufficient for contact-rich manipulation, where success depends on both global task progress and subtle local interactions such as contact force. We introduce \ours, a visuo-tactile inference-time steering framework that formulates multimodal guidance as a bi-level optimization problem. At the high level, visual sampling-and-verification performs long-horizon mode selection, deciding \textit{what} behavior the robot should execute. At the low level, tactile-guided diffusion editing refines the selected action sequence over a shorter horizon to satisfy local contact requirements. To support outcome-based steering, \ours learns a visuo-tactile latent world model and employs semantically aligned visual and tactile verifiers, including a novel text-conditioned tactile reward that scores predicted tactile futures directly in latent space. Across three real-world contact-rich manipulation tasks, \ours improves overall success by $\mathbf{51}\%$ over the base policy, outperforms unimodal steering by at least $\mathbf{33}\%$, and exceeds naive multimodal fusion by at least $\mathbf{20}\%$. Website: \href{https://yilin-wu98.github.io/vital_website/}{yilin-wu98.github.io/vital\textunderscore website/}.

\end{abstract}


\section{Introduction}
\label{sec:introduction}

Inference-time steering~\cite{wu2025foresight, du2025dynaguidesteeringdiffusionpolices, sun2026latent, kwok2025robomonkey, wang2024inference, hansteer2026, wagenmaker2025steering, yuan2026actasklearnuncertaintyaware} is an emerging way to adapt pre-trained generative robot policies~\cite{chi2024diffusionpolicy, black2024pi_0} at deployment-time and without re-training. 
By treating the frozen policy (e.g., a diffusion policy) as an action proposal generator, inference-time steering focuses on verifying or refining candidate actions before execution. 
Recent work has largely focused on visual verification: candidate actions are rolled out into predicted visual futures (images or latents) and then evaluated with a reward to pick the best action for the task~\cite{wu2025foresight,du2025dynaguidesteeringdiffusionpolices, sun2026latent,yuan2026actasklearnuncertaintyaware}. 
This has proven effective when task success is fully observable by visual cues, like picking the right object or placing it in the right goal.

\begin{figure}
    \centering
    \includegraphics[width=\linewidth]{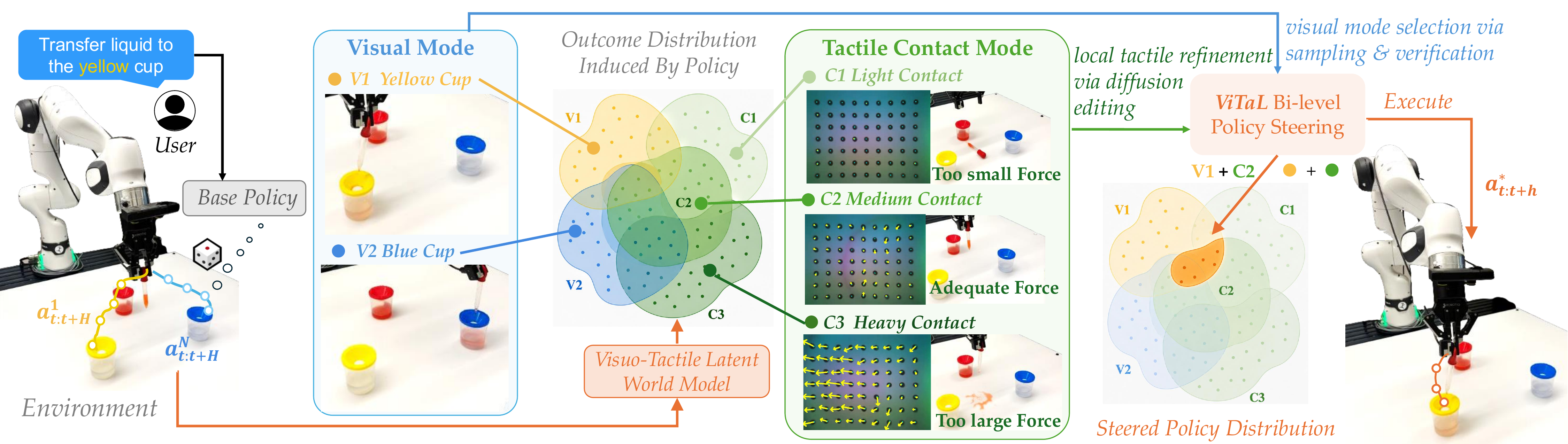}
    \caption{\textbf{Multi-Modal Policy Steering}. Our method, \ours, selects visual modes and refines local contact, steering the base policy toward actions satisfying both task goals and contact constraints.}
    \vspace{-0.4cm}\label{fig:front_figure}
\end{figure}

However, in contact-rich manipulation tasks like pipetting (Fig.~\ref{fig:front_figure}) vision-only steering is insufficient: whether a robot is about to dispense the contents of a pipette mid-transport, or whether its grasp will produce the right force at the goal location, is not observable from wrist or third-person images alone. 
This motivates \textit{multimodal steering}: using vision to guide \textit{which} behavior to pursue, and touch to govern \textit{how} to realize it through contact. 
Yet integrating these two signals during inference-time steering is non-trivial for two key reasons. 
First, visual outcomes and tactile outcomes operate on different time scales: candidate action sequences need to be rolled out long enough to reveal their semantically distinct visual outcomes (e.g., to distinguish which cup the robot is moving towards), whereas tactile outcomes are often local and transient, capturing contact events 
that may be obscured in long-horizon visual predictions (e.g., how much force is applied while squeezing the dropper head). 
Second, even with the perfect outcome predictions, its extremely challenging to have a single unified reward function for verification:
the relative importance of visual and tactile outcomes shifts across task phases, requiring rewards that are themselves phase-conditioned.

To address these challenges, we present
\ours, a visuo-tactile inference-time steering framework for contact-rich manipulation.
Our key idea is a bi-level 
optimization that decomposes multimodal guidance into visual mode selection followed by tactile contact refinement. 
In the visual stage, we perform sampling-and-verification: candidate actions are rolled out with a visuo-tactile latent world model, and a visual verifier selects the sequence that best satisfies the visual objective. In the tactile stage, we treat this selected sequence as an action anchor and apply tactile-guided diffusion editing~\cite{meng2021sdedit} over a short horizon to improve contact execution while remaining close to the visual plan. This decomposition matches each modality with its strength: vision guides the policy toward the intended behavioral mode, while touch refines the physical execution of that mode.

To specify rewards in \ours's bi-level optimization, we use text as a shared task representation to bridge visual and tactile reasoning: the high-level language instruction is decomposed into phase-level visual and tactile objectives, which are then evaluated by semantically-aligned verifiers for each modality.
A visual verifier scores decoded visual predictions for semantic progress, while our text-conditioned tactile verifier scores predicted tactile latents against textual contact objectives. This enables contact-sensitive steering that is \textit{discriminative}, \textit{flexible}, and \textit{efficient}: the verifier can distinguish contacts, adapt to different task phases through language, and guide actions directly in latent space without hand-designing task-specific tactile rewards or decoding tactile predictions.

\para{Statement of Contributions} Our contributions are threefold: (i) we propose a bi-level multimodal guidance framework that uses vision for long-horizon semantic mode selection and touch for targeted contact refinement; (ii) we introduce, to our knowledge, the first language-conditioned tactile reward for robotic manipulation, which scores predicted tactile futures in an aligned latent space, and use it alongside a separate language-conditioned visual reward and a visuo-tactile latent world model for outcome-based policy steering without task-specific reward learning; and (iii) we conduct extensive real-world experiments across three contact-rich manipulation tasks, where \ours improves overall success by $\mathbf{51\%}$ over the base policy, exceeds unimodal steering by at least $\mathbf{33\%}$, and outperforms naive multimodal fusion by at least $\mathbf{20\%}$.

\section{Related Work}

\para{Inference-time Steering of Robot Policies}
Recent work on inference-time adaptation for robot policies typically follow three paradigms: sample-then-verify~\cite{wu2025foresight,kwok2025robomonkey,hansteer2026,yuan2026actasklearnuncertaintyaware}, noise-space optimization~\cite{wagenmaker2025steering}, and classifier-based guidance~\cite{sun2026latent,du2025dynaguidesteeringdiffusionpolices,wang2024inference}. While effective, they all have focused on visual guidance signals that provide global task context but often miss fine-grained contact details in manipulation~\cite{zhang2026touchguide,xue2025reactive,higuera2026visuo,cui2026vitacman}. Some tactile steering works instead use touch to provide direct contact information~\cite{tian2019manipulation,zhang2026touchguide}, but do not capture global progress. Specifically, TouchGuide~\cite{zhang2026touchguide} learns action-space verifiers via contrastive learning, but it requires costly demonstrations to obtain positive samples and does not explicitly model dynamics, limiting its scalability.
Instead, we propose a visuo-tactile steering approach that operates at different temporal scales with a learned latent world model. In our work, visual lookahead provides global long-horizon steering, while tactile refinement enables short-horizon, contact-sensitive corrections.

\para{Tactile Sensing for Robot Decision Making}
We use vision-based tactile sensors, specifically GelSight Mini~\cite{yuan2017gelsight} due to its dense measurements of contact geometry, deformation, and force-related signals. The existing datasets with GelSight~\cite{yuan2017gelsight,yuan2015measurement,helmut2025learning} have been used to pretrain encoders for tactile images~\cite{fenganytouch,fenganytouch2,higuerasparsh,higuera2025tactile,xie2026universal}, providing semantically meaningful latent spaces for tactile reasoning. Tactile sensing has also been widely used for contact-rich manipulation~\cite{pmlr-v78-calandra17a,calandra2018more,lee2019making,chen2023visuo, zhu2025touch}, and recent work incorporates tactile feedback into learned policies for fine-grained reactive control~\cite{huang20253d,heng2025vitacformerlearningcrossmodalrepresentation,xue2025reactive,chen2026multimodal}, primarily during policy training or low-level control. In contrast, we use tactile as an inference-time guidance signal: candidate actions are scored by aligning predicted tactile embeddings with textual contact goals, enabling scalable tactile verification without demonstrations.

\para{World Modeling in Robot Decision Making}
Dynamics models have been studied in planning, control, and policy learning for robotics~\cite{sutton1991dyna,deisenroth2011pilco,chua2018deep}. Recent visual world models focus on learning latent dynamics ~\cite{zhou2025dino,huang2026vjepa,assran2025v}, adapting diffusion or flow-based video models~\cite{guo2025ctrl,alhaija2025cosmos,gao2026dreamdojo} or training with model-based reinforcement learning~\cite{hafner2019dream,hafner2023mastering,wu2023daydreamer, hafner2025training}. Meanwhile, recent visuo-tactile world models incorporate touch into dynamics modeling~\cite{higuera2026visuo,tian2019manipulation,zheng2026omnivta,yuan2026vtam}.
In this work, we train a visuo-tactile world model to evaluate the multimodal outcomes of actions directly in the latent space. 
Therefore, actions can be scored directly in the latent space without decoding images, enabling efficient text-conditioned 
guidance.

\section{Problem Formulation: Visuo-Tactile Inference-Time Steering }
\label{sec:problem_formulation}

\para{Notation \& Setup}
At any real time $t$, the robot observes \textit{multimodal} data in the form of RGB camera images $\obsv_t \in \obsSpace^v$ from $C^v$ cameras (e.g., a third person and a wrist camera), vision-based tactile images $\obst_t \in \obsSpace^\tau$ from $C^\tau$ sensors (e.g., one GelSight Mini~\cite{yuan2017gelsight} attached to the gripper fingertip), and 
proprioceptive states (e.g. end-effector pose and gripper width), $\proprio_t \in \mathcal{Q}$. 
Let the concatenated vision, tactile, and proprioceptive observation be denoted by $\obs_t := [\obsv_t,\obst_t,\proprio_t]$. 
We denote h-step \textit{future observation sequences} from time $t$ by $\obstraj_t := \obs_{t:t+h}$ with vision, tactile, and proprioceptive sequences denoted by $\obsvtraj_t$, $\obsttraj_t$, and $\propriotraj_t$ respectively.
Let the robot's low-level action space  be $\action \in \mathcal{A}$  (e.g., end-effector cartesian pose and gripper control). A $h$-step action sequence starting from time $t$ is denoted by $\acttraj_t := \actchunk$. 
The robot generates low-level ``action chunks'' via a pre-trained\footnote{While our method is agnostic to the base policy, we focus on vision-only policies for their prevalence and strong off-the-shelf performance, with the visuo-tactile variant in App~\ref{app:visuo_tactile_policy}.} diffusion policy, $\acttraj_t \sim \policy(\cdot \mid \obs_t)$. 

\para{Problem Formulation} Let $\task$ denote a textual description for the high-level goal of the task (e.g. ``transfer liquid to the blue cup''). 
Inference-time policy steering can be formalized as a model predictive control problem: the base policy $\policy$ generates candidate actions, each sample is forward simulated into future observations which capture \textit{outcomes} of the action sequence, and these outcomes are scored by a verifier for its alignment with task description. 
When the robot uses multimodal observations,  future outcomes require a \textit{visuo-tactile world model} which yields a distribution over vision, tactile, and proprioceptive sequences induced by any action sample:
$\worldmodel(\obsvtraj_t,\obsttraj_t,\propriotraj_t \mid \obsv_t,\obst_t,\proprio_t,\acttraj_t)$.
Multimodal inference-time steering requires selecting the action sample $\acttraj_t \sim \policy(\cdot \mid \obs_t)$ whose future observations maximize a task-conditioned reward $R(\obsvtraj_t, \obsttraj_t; \task)$, The reward $R$ scores how well the predicted visual and tactile futures satisfy the task instruction $\task$.

\para{Open Challenges}
\label{sec:open_challenges}Solving the general multimodal steering problem introduces two key challenges.

\textit{Modality-dependent Observability Across Temporal Scales.}
Vision and touch provide complementary observations of the task state. 
Vision captures global scene structure and semantic progress, but the relevant visual cues may only emerge over a sufficiently long horizon: for example, in the pipetting task, candidate trajectories must be rolled out far enough to determine whether the robot is moving towards the right cup. 
In contrast, touch provides direct evidence about physical interactions, such as grasp stability or wiping pressure. These contact-critical events are often local and transient, so they can be difficult to identify in long-horizon vision-only predictions. 
Thus, a single shared prediction and verification horizon is suboptimal: the horizon needed to distinguish visual behavioral modes can differ from the horizon needed to assess contact quality from touch.

\textit{Multi-modal Reward Modeling.}
Even with access to visual and tactile observations, defining rewards for contact-rich manipulation is difficult because success depends on both semantic task progress and local physical interaction. A task instruction may specify the final goal, such as transferring liquid, but it often leaves intermediate visual subgoals and contact requirements implicit. Moreover, tactile rewards must distinguish subtle contact states, such as gentle grasping versus dropping, or sufficient versus excessive force. This makes it challenging to construct a unified multi-modal reward that is both informative about global task progress and sensitive to contact quality.

\section{\ours: \textbf{Vi}suo-\textbf{Ta}ctile \textbf{L}atent Steering}
\label{sec:method}
\ours addresses the challenges in Sec.~\ref{sec:open_challenges} by decomposing multimodal steering into long-horizon visual mode selection and short-horizon tactile refinement. The visual stage selects an action mode based on predicted semantic progress, while the tactile stage refines it using predicted contact quality, matching the complementary timescales of vision and touch. We formulate this bi-level objective in Sec.~\ref{subsec:method_bilevel}, introduce a visuo-tactile latent world model for outcome prediction in Sec.~\ref{subsec:method_world_model}, define visual and tactile verifiers in Sec.~\ref{subsec:method_reward_modeling}, and describe the overall steering approach in Sec.~\ref{subsec:method_bilevel_execution}.

\subsection{Bi-level Visuo-Tactile Steering Formulation}
\label{subsec:method_bilevel}
We formulate the multimodal steering problem as a bi-level optimization problem.
For both levels of the optimization, we use a latent visuo-tactile world model to predict the outcomes of any action sequence, $\acttraj$, across both modalities:
$\latenttrajvpred, \latenttrajtpred \sim  p_\phi\!\left(
\cdot,\cdot
\mid
\latentv, \latentt, \acttraj
\right)$ where $\latent \in \latentSpace$ are the latent states obtained from the respective modality's encoder $\enc^m(\obs), m \in \{v, \tau\}$. 
The \textit{inner} optimization selects a globally promising action mode $\bar{\acttraj}_{t:t+H}$ with visual task reward $R^v(\latenttrajvpred; \task)$. 
The \textit{outer} optimization stays close to, but refines, the first $h$ steps of the visual plan $\bar{\acttraj}_{t:t+h}$ using the tactile latents and reward $R^\tau(\latenttrajtpred; \task)$.
The final action chunk informed by both vision and touch is denoted by $\acttraj^\star_{t:t+h}$:
\begin{equation}
\begin{aligned}
\mathbf{a}^*_{t:t+h}
&=
\arg\max_{\acttraj_{t:t+h}}
\overbrace{
\log p_\theta\!\left(
\acttraj_{t:t+h}
\mid
\textcolor{ocean}{\bar{\mathbf{a}}_{t:t+h}}, o_t
\right)
}^{\text{visual prior around selected mode}}
+
\beta \cdot
\overbrace{
R^\tau\!\left(
\textcolor{solid_green}{\latenttrajtpred_{t:t+h}}; \mathcal{L}
\right)
}^{\text{\textcolor{solid_green}{\textbf{tactile}}: local contact reward}}
\\
&\text{s.t.}\quad
\textcolor{ocean}{\bar{\mathbf{a}}_{t:t+H}}
=
\underbrace{
\arg\max_{\mathbf{a}_{t:t+H}\sim \pi_\theta(\cdot\mid o_t)}
R^v\!\left(
\textcolor{ocean}{\latenttrajvpred_{t:t+H}}; \mathcal{L}
\right)
}_{\text{\textcolor{ocean}{\textbf{vision}:} select action mode via visual latents \& reward}},
\\
& \qquad \textcolor{ocean}{\latenttrajvpred_{t:t+h'}}, \textcolor{solid_green}{\latenttrajtpred_{t:t+h'}}
\sim
\underbrace{p_\phi\!\left(
\cdot,\cdot
\mid
\latentv_t, \latentt_t, \acttraj_{t:t+h'}
\right)}_{\text{latent visuo-tactile world model}},
\quad h' \in \{h, H\}, \\ 
&\qquad \latentv_t =\encv(\obsv_t), \quad \latentt_t =\enct(\obst_t).
\end{aligned}
\label{eq:bilevel_joint_annotated}
\end{equation}

In the \textit{inner} optimization, actions are sampled from the base policy, $\acttraj_{t:t+H} \sim \pi_\theta(\cdot \mid \obs_t)$, rolled out with the latent visuo-tactile dynamics model $p_\phi$, and the actions with the highest long-horizon visual reward $R^v(\latenttrajvpred_{t:t+H}; \task)$ are chosen to encourage global task progress. 
The \textit{outer} optimization uses the visually-selected action mode, $\bar{\acttraj}$, as an initial behavior proposal. 
The first term in the optimization, $p_\theta$, is a visual action prior (described in detail in Sec.~\ref{subsec:method_bilevel_execution}) centered on this action mode; this encourages the optimization to remain close to the globally desired behavior. 
The second term locally refines this visually selected mode by reweighting shorter $h$-step action chunks according to the tactile reward $R^\tau(\latenttrajtpred_{t:t+h}; \task)$. 
The hyperparameter $\beta$ controls the strength of tactile refinement relative to the visual prior. 
Intuitively, our formulation uses vision for global mode selection and then touch for local contact refinement.

\subsection{Visuo-Tactile Latent World Model for Multi-Modal Latent Outcome Prediction}
\label{subsec:method_world_model}

Eq.~\ref{eq:bilevel_joint_annotated} uses a latent world model $p_\phi$ to predict the visual and tactile futures induced by candidate action sequences. 
We encode the current visual and tactile observations separately as
$
    \latentv_t = \encv(\obsv_t), \;
    \latentt_t = \enct(\obst_t), \;
$
with joint vision-tactile state $\latent_t = (\latentv_t, \latentt_t)$. In our implementation, $\encv$ is the pretrained DINOv3 encoder~\cite{simeoni2025dinov3} and $\enct$ is the pretrained AnyTouch2 encoder~\cite{fenganytouch2} with structured latent spaces for scene understanding and tactile reasoning.
On top of the frozen encoders, we train an action-conditioned latent dynamics model inspired by~\cite{zhoudino}. Starting from $\hat{\latent}_t=\latent_t$, the model recursively predicts future joint visuo-tactile latents over action chunks:
$    \hat{\latent}_{t+h}
    =
    p_\phi
    \left(
        \hat{\latent}_{t},
        \action_{t:t+h}
    \right).$
We train the model 
with a multi-step latent prediction objective following~\cite{higuera2026visuo} with details in App~\ref{app:implement_visuo_tactile_world_model}.
We additionally train decoders
$
    \hat{\obs}^v = \decv(\latentv), \;
    \hat{\obs}^\tau = \dect(\latentt)
$
for visualization.
Together, this multimodal model enables long-horizon visual lookahead as well as short-horizon tactile prediction.

\subsection{Multi-modal Reward Modeling for Verification}
\label{subsec:method_reward_modeling}
After the latent world model predicts multimodal (latent) outcomes, we need to verify how well these predictions align with global task progress and local contact quality, which may vary across task phases. 
We use language-conditioned visual and tactile verifiers, with phase-dependent textual objectives specifying the relevant criteria for each modality.

\para{Phase-Dependent Verification Objectives}
A single task instruction is often too coarse for long-horizon manipulation: e.g., $\task$=``transfer liquid to the blue cup and return'' does not explicitly specify intermediate visual objectives (such as moving back to the red cup) or contact requirements (such as squeezing only during dispensing). We therefore use a VLM offline to decompose the task into phase-level textual objectives,
    $\{\lang_p\}_{p=1}^{P} = g^{\mathrm{VLM}}(\task)$, 
where each phase provides a visual objective $\lang_p^v$ and a tactile objective $\lang_p^\tau$. The visual verifier determines when to switch to the next phase.

\para{Visual Verifier}
Visual success depends on if the imagined outcomes appear to be making progress toward the visual objective described in $\lang_p^v$. 
Recent work has developed strong, off-the-shelf, general-purpose reward models that take as input RGB images and textual task descriptions \cite{tan2025robo, liang2026robometer, ma2025vision}.
We leverage these models zero-shot, and we decode our predicted visual latents into RGB image observations $\hat{\obs}^v_{t+H} = \decv(\hat{\latent}^v_{t+H})$ and concatenate that with the previous history of observations we have obtained $\obs_{\leq t}$ as the input of ROBOMETER~\cite{liang2026robometer} reward model for online visual evaluation. 
This reward is used for long-horizon visual mode selection, where the goal is to distinguish globally meaningful task progress, like moving to the correct cup, mark, or hole.

\para{Tactile Verifier}
Tactile success depends on fine-grained, often visually ambiguous interactions, such as stable grasping versus incipient slip. 
We introduce a tactile verifier that is \textit{discriminative} over subtle contact outcomes, \textit{efficient} for inference-time  evaluation, and \textit{flexible} to adapt to phase-specific textual contact goals.
Specifically, we use a pretrained tactile encoder~\cite{fenganytouch2} with a semantically-aligned latent space and define the tactile reward as the cosine similarity between the phase-level contact description and the tactile embedding: 
$$R^\tau(\hat{\latent}^\tau_{t+h}, \lang_p^\tau)
=
\cos\left(
    \hat{\latent}^\tau_{t+h},
    \enc^\lang(\lang_p^\tau)
\right),$$
where $\enc^\lang(\cdot)$ is the corresponding text encoder. Computing rewards directly in this latent space makes the verifier discriminative over subtle contact states, efficient for steering without tactile image decoding, and flexible to language-specified contact goals such as ``grasp lightly'' or ``grasp heavily.'' To the best of our knowledge, this is the first general-purpose language-conditioned tactile reward.

\subsection{Inference-time Optimization for Bi-level Visuo-Tactile Steering Objective }
\label{subsec:method_bilevel_execution}
At inference time, we solve the bi-level optimization problem from Sec.~\ref{subsec:method_bilevel} by combining sampling-based visual mode selection with tactile-guided diffusion editing. We describe the implementation details behind this optimization in this section, with further details in Algo.~\ref{alg:vital} in App.~\ref{app:implement_visual_steering}.

\para{Sampling \& Verification as Visual Steering}
As shown in the inner optimization in Eq.~\ref{eq:bilevel_joint_annotated}, the visual stage selects a global behavior mode by evaluating long-horizon candidate action sequences. We obtain a set of long-horizon action sequences
$
\mathbb{A}_N=\{\acttraj_{t:t+H}^{(i)} \mid 
\acttraj_{t:t+H}^{(i)} \sim \pi_\theta(\cdot \mid \obs_t)\}_{i=1}^{N}
$
by rolling-out the short action samples obtained from the policy within the world model, decoding the visual observations, passing these into the policy, and repeating. We choose the sequence with the highest visual reward:
$
\bar{\acttraj}_{t:t+H}
=
\arg\max_{\acttraj_{t:t+H} \in \mathbb{A}_N}
R^v
\left(
    \hat{\latenttraj}^v_{t:t+H}; \lang_p^v
\right).$
We use sampling-based optimization here because of its simplicity and stability; alternatives like classifier-based guidance would  
require differentiating through  through the VLM, latent world model, and recursive policy generation, resulting in potentially unstable gradients.  
Thus, we use the visual verifier to rank candidate futures rather than guide denoising directly. 
The selected sequence $\bar{\action}_{t:t+H}$ serves as the visual anchor, and its first $h$ steps are passed to the outer optimization for tactile refinement.

\para{Diffusion Editing as Tactile Steering}
Conditioned on the visual anchor $\bar{\acttraj}_t :=\bar{\acttraj}_{t:t+h}$, the goal of tactile steering is to locally refine the action chunk to satisfy contact requirements while preserving the action mode selected by the visual steering. We evaluate tactile refinement using the short-horizon tactile reward in Sec.~\ref{subsec:method_reward_modeling}. The short horizon $h$ is important because tactile signals are most informative for immediate contact correction, such as adjusting force or stabilizing a grasp.

Following Eq.~\ref{eq:bilevel_joint_annotated} and prior diffusion editing methods~\cite{meng2021sdedit}, we define a reference-conditioned prior around visual anchor $\bar{\acttraj}_{t}$:
$
p_\theta(\acttraj_{t} \mid \bar{\acttraj}_{t}, \obs_t)
:=
\int
    \rho_\theta(\acttraj_{t} \mid \mathbf{x}_K, \obs_t)
\nu_K(\mathbf{x}_K \mid \bar{\acttraj}_t)
d\mathbf{x}_K,
$
where $\nu_K$ partially noises $\bar{\acttraj}_t$ to diffusion level $K$ and $\rho_\theta$ denoises it under the base policy. This prior captures plausible local edits of the visual anchor, with smaller $K$ enforcing closer refinements and larger $K$ allowing larger deviation. We then define the tactile-refined policy $p^\textrm{refine}$ by reweighting the policy distribution with predicted tactile reward,
$
\log p^{\mathrm{refine}}
=
\log p_\theta(\acttraj_t \mid \bar{\acttraj}_t, \obs_t)
+
\beta R^\tau(\hat{z}^\tau_{t+h}, \lang^\tau_p),
$
and approximate sampling from $p^\textrm{refine}$ via tactile-guided reverse diffusion: we partially noise $\bar{\acttraj}_t$ and bias denoising toward actions with higher tactile reward. Thus, the visual stage selects \emph{what} global behavior to execute, while the tactile stage refines \emph{how} to execute it with touch. Details are in App.~\ref{app:implement_visual_steering}-~\ref{app:implement_tactile_steering}.

\section{Experiments}
\label{sec:experiments}
We evaluate \ours on three real-world contact-rich manipulation tasks. 
Our experiments are designed to answer three questions: whether multimodal visuo-tactile steering improves policy over unimodal guidance (Sec.~\ref{sec:exp_multimodal_guidance}); whether a learned latent world model and semantically aligned verifier provides reliable rewards for predicted outcomes (Sec.~\ref{sec:experiment_verifier}); and whether the proposed bi-level optimization balances performance and efficiency compared to naive visual-tactile fusion (Sec.~\ref{sec:experiment_bilevel}).

\textbf{Robot Setup, Tasks, and Data Collection.}
We use a Franka Emika robot with RGB and tactile sensors to evaluate \ours on three real-world contact-rich manipulation tasks, including \textit{wiping}, \textit{insertion}, and \textit{pipette transfer}. These tasks require both global visual reasoning, such as selecting the correct mark, hole, or target cup, and local tactile feedback for maintaining contact, resolving alignment, or stabilizing grasp and dispensing behavior. 
 For each task, we collect $50$ expert demonstrations to train the base diffusion policy $\policy$ with additional $250$ policy rollouts from intermediate policy checkpoints to train the visuo-tactile world model.  Details are in App~\ref{app:real_world_setup}.

\textbf{Base Policy.}
We train a diffusion policy that predicts a 16-step action chunk and executes the first $h=8$ actions at each step. For long-horizon lookahead $H=16$, we recursively roll out the visuo-tactile world model and condition the policy on predicted future observations. Our main experiments use a vision-conditioned policy $\pi_\theta(\action \mid \obsv_t)$; App~\ref{app:visuo_tactile_policy} evaluates a visuo-tactile policy $\pi_\theta(\action \mid \obsv_t,\obst_t)$.

\textbf{World Model and Verifiers.}
We train separate visual and tactile decoders, and a transformer-based latent dynamics model for visuo-tactile prediction. For steering, we modify ROBOMETER~\cite{liang2026robometer} with KV caching for online visual reward estimation, and compute tactile rewards by comparing pretrained tactile embeddings~\cite{fenganytouch2} with CLIP text embeddings~\cite{radford2021learning}. Details are in App~\ref{app:implement_visuo_tactile_world_model} -~\ref{app:implement_tactile_steering}.

\begin{figure}[!ht]
\vspace{-0.4cm}
    \centering
    \includegraphics[width=\linewidth]{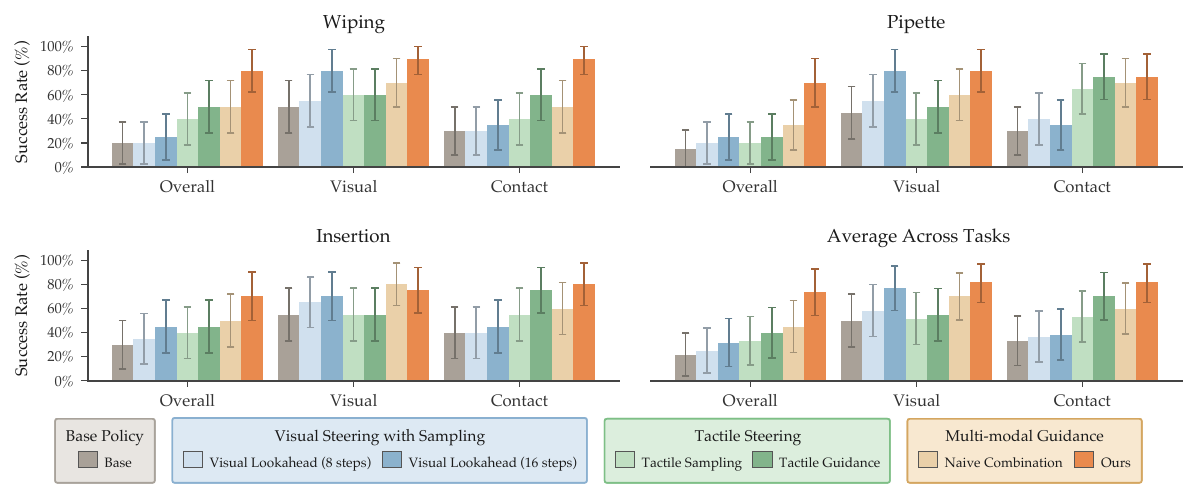}
   \caption{\textbf{Policy Steering Performance}. \ours outperforms vision-only, tactile-only, and naive multimodal baselines across three tasks. Error bars show binomial standard errors over $20$ trials.}
    \vspace{-0.2cm}
    \label{fig:policy_success_rate}
\end{figure}

\subsection{Multi-Modal Visuo-Tactile Steering Outperforms Single-Modality Steering}
\label{sec:exp_multimodal_guidance}
We compare our multi-modal guidance \ours against five methods in Fig.~\ref{fig:policy_success_rate}: 1) 
\textbf{Base Policy}; 2) visual steering, including
\textbf{Visual Lookahead (8 steps) vs (16 steps)}, which selects the best of $N=10$ with visual verifier over different prediction horizons; 3) tactile steering, including
\textbf{Tactile Sampling}, which selects the best of $N=10$ samples with tactile verifier; 
\textbf{Tactile Guidance}, classifier-based tactile guidance;  
We omit classifier-based visual guidance due to the reasons discussed in Sec.~\ref{subsec:method_bilevel_execution}.
We report \textbf{Overall}, \textbf{Visual}, and \textbf{Contact} success, measuring full task completion, correct visual goal following, and appropriate local physical interaction, respectively.

\paras{Results}
In Fig.~\ref{fig:policy_success_rate}, visual-only steering improves visual success but provides limited gains in overall task completion, while tactile-only steering improves contact behavior without reliably satisfying the task goal. In contrast, \ours jointly improves visual and contact success, leading to a $51\%$ increase in overall task success over base policy and at least 33\% gain over unimodal baselines.

\begin{figure}[t]
    \centering
    \includegraphics[width=0.95\linewidth]{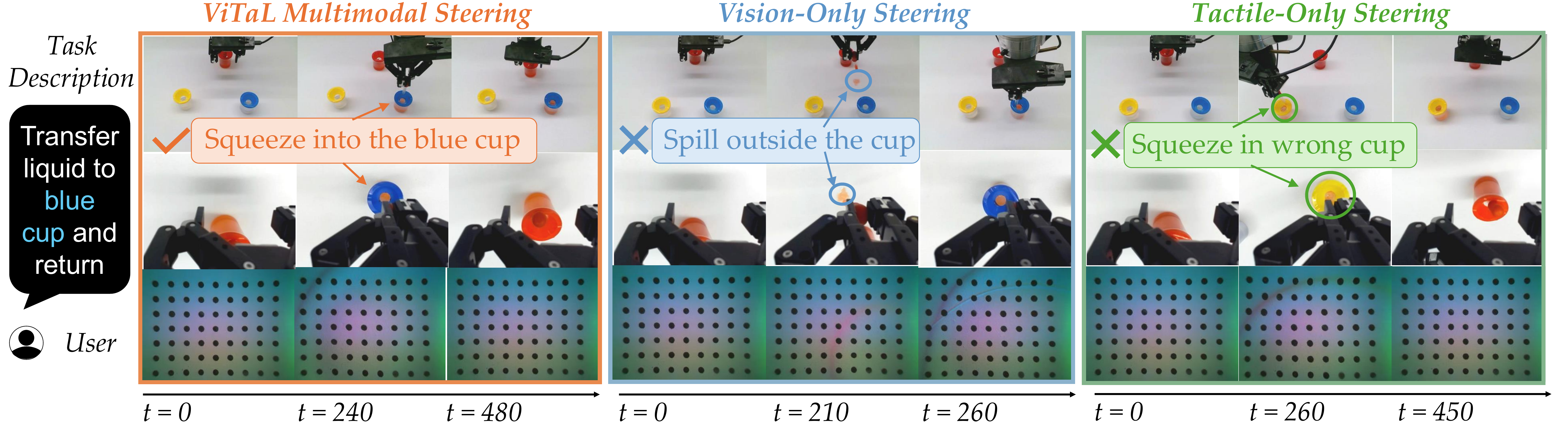}
    \caption{
    \textbf{Policy Steering with Different Guidance Modalities}. Vision-only Steering (middle) fails due to local contact errors, while tactile-only steering (right) can fail due to global task misalignment. \ours (left) with multimodal steering addresses both failure modes.
    }
    \vspace{-0.4cm}\label{fig:multimodal_failure_modes}
\end{figure}

Fig.~\ref{fig:multimodal_failure_modes} illustrates why multimodal steering is particularly helpful in the pipette task: vision-only steering can choose globally plausible but imprecise actions (e.g., spilling), while tactile-only steering can improve local contact with adequate grasp force but it misses the intended visual goal. 
Our multimodal method succeeds by satisfying both local and global performance requirements. More qualitative results for other tasks are provided in App~\ref{app:results_multi_modal}.

\begin{table}[t]
\centering
\tiny
\caption{
\textbf{Reward Accuracy}. We evaluate reward with preference-order accuracy. GT versus Pred uses ground-truth versus predicted futures. Error bars show binomial standard errors over $40$ trials.
}
\label{tab:reward_eval}
\resizebox{0.95\linewidth}{!}{
\begin{tabular}{lcccccccc}
\toprule
\multirow{2}{*}{\textbf{Src.}}
& \multicolumn{4}{c}{\textbf{Visual Reward (\%)}} 
& \multicolumn{4}{c}{\textbf{Tactile Reward (\%)}} \\
\cmidrule(lr){2-5}
\cmidrule(lr){6-9}
& \textbf{Wiping} & \textbf{Insertion} & \textbf{Pipette} & \textbf{Avg.}
& \textbf{Wiping} & \textbf{Insertion} & \textbf{Pipette} & \textbf{Avg.} \\
\midrule
GT
& $80.0{\pm}6.3$ & $70.0{\pm}7.2$ & $\mathbf{100.0{\pm}0.0}$ & $83.3{\pm}5.9$
& $85.0{\pm}5.6$ & $70.0{\pm}7.2$ & $\mathbf{80.0{\pm}6.3}$ & $78.3{\pm}6.5$ \\
Pred
& $\mathbf{82.5{\pm}6.0}$ & $\mathbf{72.5{\pm}7.1}$ & $\mathbf{100.0{\pm}0.0}$ & $\mathbf{85.0{\pm}5.6}$
& $\mathbf{90.0{\pm}4.7}$ & $\mathbf{77.5{\pm}6.6}$ & $77.5{\pm}6.6$ & $\mathbf{81.7{\pm}6.1}$ \\
\bottomrule
\end{tabular}
}
\vspace{-0.4cm}
\end{table}
\subsection{Semantically-Aligned Visual and Tactile Verifiers Enable Global and Local Adaptation}
\label{sec:experiment_verifier}

\begin{figure}
    \centering
    \includegraphics[width=0.9\linewidth]{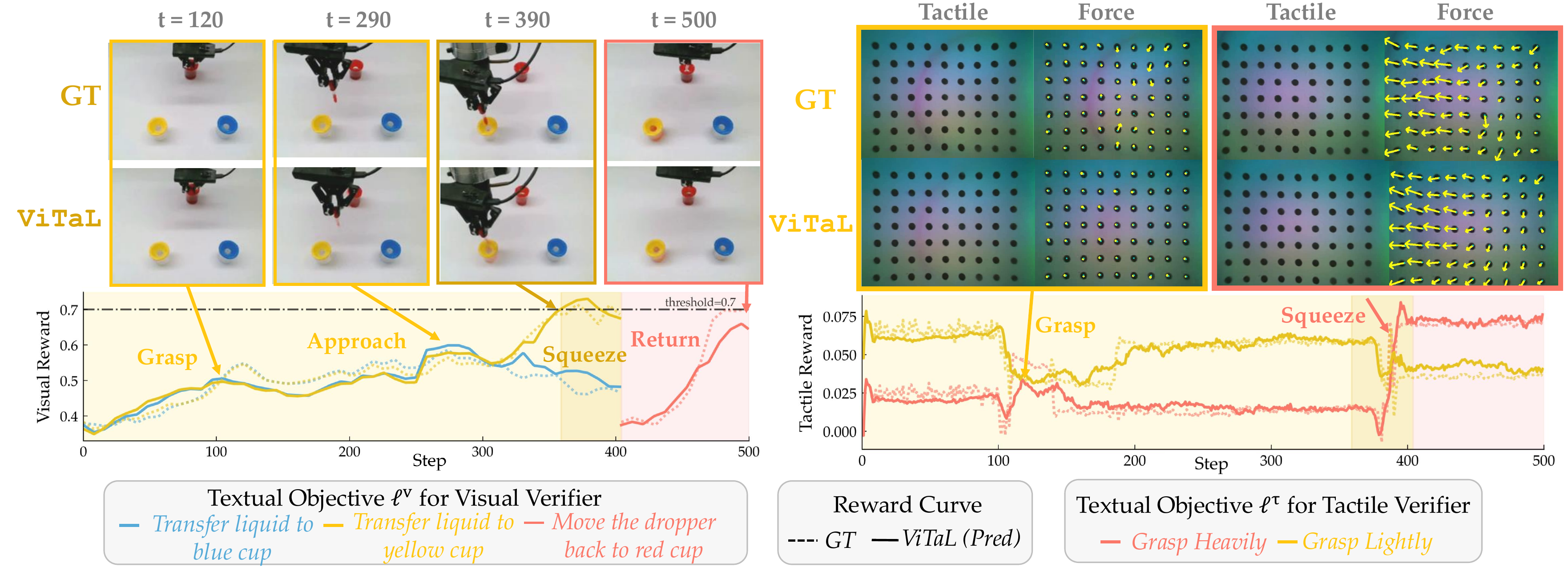}
    \caption{
\textbf{Visual and Tactile Verification.}
Our visual verifier (left) identifies whether the target cup is selected and the phase transition, while the tactile verifier (right) captures changes in grasp force.
}
  \vspace{-0.3cm}\label{fig:reward_qualitative}
\end{figure}

Effective steering requires accurately predicting \textit{both} visuo-tactile outcomes and their respective rewards. Thus, we evaluate whether our visual and tactile verifiers provide reliable semantic rewards for steering predicted futures. 
We first compare our joint visuo-tactile world model against separately trained unimodal ones with standard visual metrics and tactile-related metrics. For brevity, these world models results are in App~\ref{app:world_model_ablation} and we focus on the reward model results in this section. However, Fig.~\ref{fig:reward_qualitative} visualizes the decoded visual and tactile world model predictions in the pipette task.

\para{Metrics} We focus on evaluating reward quality of both ground-truth observations (\textbf{GT}) and world-model predictions (\textbf{Pred}) with preference-order accuracy~\cite{tian2026position} shown in App.~\ref{app:results_reward_world_model}. For each task, we select rollout pairs with different human preferences, rank each pair by the sum of predicted reward divided by rollout length, and measure how often the predicted order matches the human order.

\para{Results}
Table~\ref{tab:reward_eval} shows that both the ROBOMETER visual reward and our tactile reward achieve at least $70\%$ preference-order accuracy across tasks on both \textbf{GT} and \textbf{Pred} observations. In some cases, \textbf{Pred} rewards are slightly more accurate, likely because predicted latents smooth observation noise (Fig.~\ref{fig:reward_qualitative} right) and are easier to rank due to lower visual/tactile variations than real rollouts.

Fig.~\ref{fig:reward_qualitative} further illustrates the semantic alignment of both verifiers on the pipette task. The visual verifier assigns higher reward when the robot dispenses into the cup that matches the target cup specified in textual description (e.g, the yellow cup), and phase-level textual objectives focus on the alignment to the relevant subgoal (e.g. transfer to the target cup or move back to the red cup). The tactile verifier captures the transition from gentle to firm grasp, consistent with marker-tracking force estimates. Together, these results show that our visual rewards can identify global goal satisfaction, while our proposed tactile rewards provide meaningful local contact signals.

\begin{figure}
    \centering
    \includegraphics[width=0.9\linewidth]{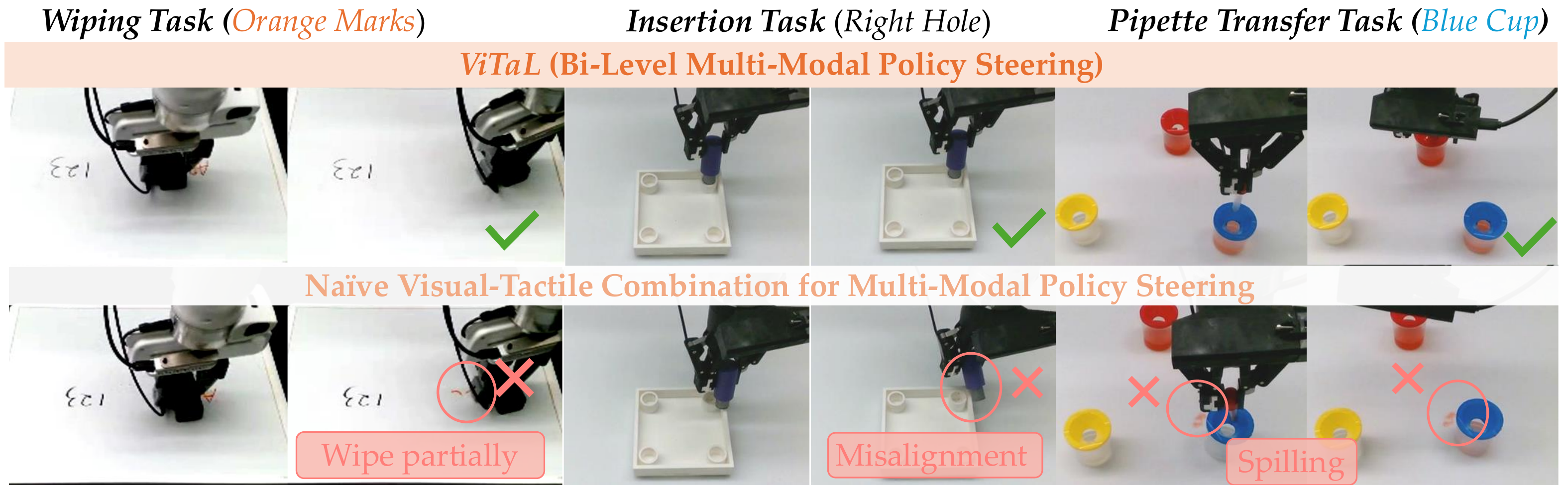}
    \caption{\textbf{Multi-Modal Policy Steering}. \ours (top) successfully completes three contact-rich tasks while naive combination (bottom) of both modalities during steering always fails.}
    \vspace{-0.cm}\label{fig:bilevel_qualitative}
\end{figure}

\subsection{Bi-Level Steering Achieves Balance Between Performance and Inference Speed}
\label{sec:experiment_bilevel}

We evaluate whether \ours achieves a better trade-off between task success and inference speed than naive multimodal fusion. We compare \textbf{Base Policy}; \textbf{Naive Combination}, which normalizes visual and tactile rewards separately across $N=10$ samples and sums them for action selection; and \ours, which first selects actions using long-horizon visual lookahead and then refines it with local tactile guidance. We report \textbf{Overall}, \textbf{Visual}, and \textbf{Contact} success as in Sec.~\ref{sec:exp_multimodal_guidance}, together with inference-time overhead, to assess whether \ours improves steering without sacrificing efficiency.

\para{Results}
As shown in Fig.~\ref{fig:policy_success_rate}, \ours outperforms \textbf{Naive Combination} by at least $20\%$ across all three tasks. Naive reward fusion often suffers from reward imbalance: the visual objective can dominate selection, leading to actions that appear globally plausible but fail to establish proper contact. Fig.~\ref{fig:bilevel_qualitative} illustrates these failures, where naive fusion can leave orange marks partially wiped, misalign the peg with the hole, or spill liquid. In contrast, \ours uses vision to efficiently select globally appropriate behaviors and touch to make targeted local corrections. As detailed in App.~\ref{app:results_bi_level}, \ours adds $0.05$ second per policy inference step (8 actions) over naive fusion while yielding substantially higher success. These results show that bi-level steering provides a stronger success--speed trade-off.

\section{Conclusion}
\label{sec:conclusion}

We presented \ours, a bi-level visuo-tactile inference-time steering framework for contact-rich manipulation. By combining long-horizon visual mode selection with short-horizon tactile refinement in a latent visuo-tactile world model, \ours improves performance across three real-world, contact-rich manipulation tasks: pipetting, wiping, and peg insertion.

\para{Limitations} Despite the improvement of \ours over the base policy, several limitations remain. \ours depends on the fidelity of the latent world model, where compounding prediction errors can affect verification, especially for subtle contact events. Its tactile verifier is also limited by tactile encoders pretrained on much smaller datasets than modern VLMs, suggesting that larger-scale tactile pretraining could enable richer contact reasoning. Furthermore, future work can also extend multimodal policy steering to more complex embodiments, such as dexterous hands.

\section{Acknowledgments}
The authors were partially supported by the National Science Foundation (NSF) award $[\#2246447]$, NSF CAREER award $[\#2441014]$, and Samsung Research through the LEAP-U program. The views expressed are those of the authors and do not necessarily reflect those of NSF or Samsung. We would like to thank Ruihan Gao, Jehan Yang and Hung-Jui Huang for sharing their Gelsight Mini sensors and Kensuke Nakamura for designing the custom mount of Gelsight Mini  for the Robotiq Gripper. We also appreciate the discussion of the project with Junwon Seo, Kensuke Nakamura and Hyun Joe Jeong as well as Zhanyi Sun's feedback for the manuscript.

\clearpage

\bibliography{reference}  %
\clearpage
\appendix
\section*{Appendix}
\startcontents[appendix]
\printcontents[appendix]{}{1}{\setcounter{tocdepth}{2}}
\clearpage
\section{Real Robot Experiment Setup}
\label{app:real_world_setup}
\subsection{Wiping Task}
\label{app:wipe_task}

\begin{wrapfigure}{r}{0.5\textwidth}
\vspace{-0.6cm}
  \begin{center}
    \includegraphics[width=0.48\textwidth]{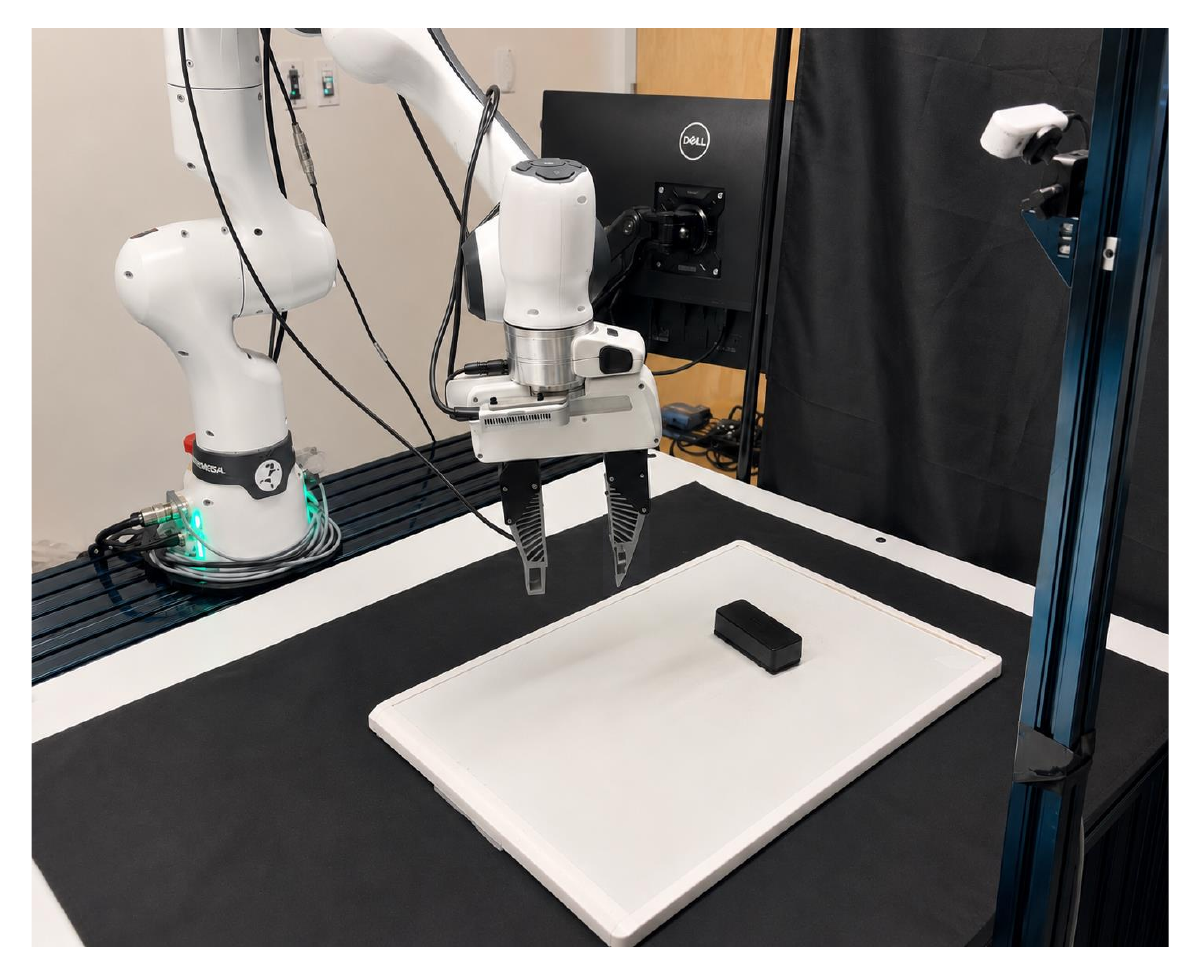}
  \end{center}
  \caption{\textbf{Wiping Task Robot Setup}}
  \label{fig:wiping_task}
\end{wrapfigure}

\para{Setup} We use a 7-DoF Franka Emika robot equipped with 3D-printed soft compliant fingers attached to the end-effector, following~\cite{chi2024diffusionpolicy}. We design a custom mount on one fingertip to attach a GelSight Mini tactile sensor~\cite{yuan2017gelsight}; the mount design is available on the project website. We use two RealSense D435 cameras: one mounted on the end-effector as a wrist camera and one placed in front of the robot as an external camera, as shown in Fig.~\ref{fig:wiping_task}. The action space of the robot is 8-dimensional including the delta cartesian pose of the end-effector (3-dimensional position and 4-dimensional quaternion) and 1-dimensional binary gripper control. The low-level controller is implemented using Deoxys: \href{https://github.com/UT-Austin-RPL/deoxys_control}{https://github.com/UT-Austin-RPL/deoxys\textunderscore control} with OSC POSE control mode. 

The wiping task requires the robot to grasp a black eraser and wipe marked regions on a whiteboard. The board contains black ``123'' marks on the left side and orange ``ABC'' marks on the right side. We evaluate two language instructions: ``wipe the orange marks'' and ``wipe the black marks.''

\para{Base Policy} We collect 25 demonstrations for wiping the orange marks and 25 demonstrations for wiping the black marks using the Meta Oculus Quest 3 as the teleoperation device. The demonstrations have mixed quality: some fully erase the marks, while others only partially erase them. For the visual policy, at each timestep, the policy receives two RGB observations, corresponding to the wrist and external views, together with the robot proprioceptive state, and predicts a 16-step action sequence. For the visuo-tactile policy, the policy receives an additional tactile image as input. All images are resized to $256 \times 256$. During execution, we apply only the first 8 steps of the predicted action sequence.

\para{Visuo-Tactile World Model} We collect 280 policy rollouts using different policy checkpoints. Together with the 50 demonstrations, this yields 330 trajectories, which we split into a training set of 300 trajectories and a validation set of 30 trajectories.

\para{Visual \& Tactile Verifiers} For the wiping task, the task goal is straightforward and consists of a single phase. We use either ``wipe the orange marks'' or ``wipe the black marks'' as the task description $\lang_1^v$ for the visual verifier, and we use ``wipe the board'' as the language input $\lang_1^\tau$ for the tactile verifier.

\para{Evaluation} During training and evaluation, we slightly randomize the eraser position within the region directly under the gripper. For each new trial, we redraw the orange and black marks and randomize their positions within the right and left regions of the board, respectively. To evaluate success rate, we run 10 trials with the instruction ``wipe the orange marks'' and 10 trials with the instruction ``wipe the black marks.'' For each trial, we count visual success if the robot moves to the correct region of the board. We count contact success if the robot completely wipes the black or the orange marks. A trial is counted as an overall task success only when both visual and contact success are achieved.

\subsection{Insertion Task}
\label{app:insertion_task}

\begin{wrapfigure}{r}{0.5\textwidth}
\vspace{-0.7cm}
  \begin{center}
    \includegraphics[width=0.48\textwidth]{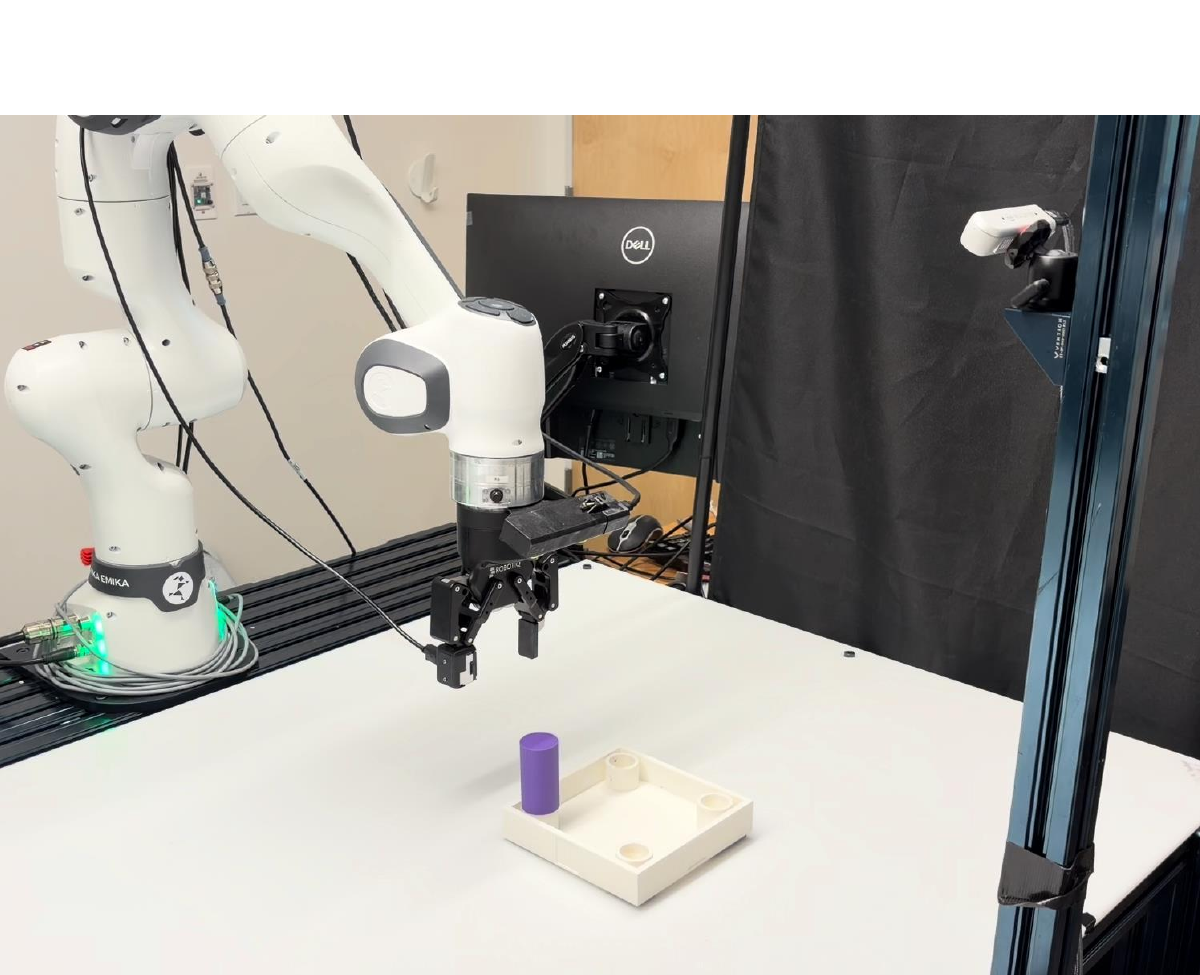}
  \end{center}
  \caption{\textbf{Insertion Task Robot Setup}}
  \label{fig:insert_task}
\end{wrapfigure}

\para{Setup} The insertion task uses the same overall hardware and data collection pipeline as the wiping task in Appendix~\ref{app:wipe_task}, with several task-specific modifications. We use a 7-DoF Franka Emika robot equipped with a Robotiq 2F-85 gripper. A GelSight Mini tactile sensor~\cite{yuan2017gelsight} is mounted to one of the fingertips of the Robotiq gripper and the custom mount is also available on the website. For visual observations, we use a ZED Mini camera mounted on the end-effector as the wrist camera, together with an external RealSense D435 camera placed in front of the robot as shown in Fig.~\ref{fig:insert_task}. We use a 8 dimension action to control the robot similar to the wiping task but the gripper control is a continuous value from $[0,1]$ instead of the binary control as wiping task.

The task involves inserting a purple cylindrical peg into a white tabletop fixture that is fixed to the table. The fixture contains four holes, one at each corner. We consider two target goals: inserting the peg into the top-left hole or the top-right hole. The clearance between the peg and each hole is approximately $1$ mm, requiring precise contact-rich manipulation.

\para{Base Policy} We collect 50 demonstrations using the same teleoperation device, with 25 demonstrations targeting the top-left hole and 25 targeting the top-right hole. As in the wiping task, the demonstrations have mixed quality, including both successful insertions and partial or imperfect attempts. The policy observations, action prediction horizon, and execution strategy follow the wiping-task setup in Appendix~\ref{app:wipe_task}, except that the wrist RGB observation is captured using the ZED Mini camera and the tactile observation comes from the GelSight Mini mounted on the Robotiq gripper.
\para{Visuo-Tactile World Model} We collect the world model data using the same procedure as in Appendix~\ref{app:wipe_task}. Policy rollouts collected from different checkpoints are combined with the demonstration data to form the training and validation sets for learning future latent imaginations.

\para{Visual \& Tactile Verifiers} The insertion task consists of a single phase. For the visual verifier, we use the task description $\lang_1^v$ corresponding to the commanded target hole: either ``the robot inserts the purple peg into the top left corner of the tabletop'' or ``the robot inserts the purple peg into the top right corner of the tabletop.'' For the tactile verifier, we use ``insert the peg'' as the language input $\lang_1^\tau$.

\para{Evaluation} For training and evaluation, we randomize the peg position within a $15cm \times 15cm$ region between the robot base and the gripper for each trial. During evaluation, we choose 10 trials with the left goal and 10 trials with the right goal and assess whether the robot selects the correct target side and whether it completes the insertion. We count visual success if the robot moves toward the correct side of the tabletop, corresponding to the commanded top-left or top-right target. We count contact success if the peg is inserted into one of the holes. A trial is counted as an overall task success only when both visual and contact success are achieved.

\subsection{Pipette Task}
\label{app:pipette_task}
\para{Setup} The pipette transfer task uses the same hardware setup as the insertion task in Appendix~\ref{app:insertion_task}, including the Franka Emika robot, Robotiq 2F-85 gripper, GelSight Mini tactile sensor with the task-specific custom mount, ZED Mini wrist camera, and external RealSense D435 camera. The action space is the same as the insertion task.

\para{Base Policy} We use the same policy architecture and demonstration collection procedure as in Appendix~\ref{app:wipe_task}. The policy takes wrist and external RGB observations, proprioceptive state, and, for the visuo-tactile policy, an additional GelSight tactile image as input. 

\para{Visuo-Tactile World Model} We collect policy rollouts for the visuo-tactile world model following the same procedure as the insertion task in Appendix~\ref{app:insertion_task}. Demonstrations and rollouts from different policy checkpoints are combined to train the model to generate future latent imaginations conditioned on memory, history, and candidate action sequences.

\para{Visual \& Tactile Verifiers} Unlike the wiping and insertion tasks, the pipette transfer task consists of two phases. The high-level task description is either ``transfer the liquid to the yellow cup and return'' or ``transfer the liquid to the blue cup and return.'' In the first phase, the visual verifier uses the goal $\lang_1^v$ corresponding to the commanded target cup: either ``transfer the liquid to the yellow cup'' or ``transfer the liquid to the blue cup.'' The tactile verifier initially uses ``grasp lightly'' as the tactile language input $\lang_1^\tau$. Once the visual reward exceeds $0.7$, we switch the tactile description to ``grasp heavily'' and maintain this tactile goal for 40 steps. After this transition period, we switch the visual verifier to the second-phase goal, ``put the dropper back to the red cup.''

\para{Evaluation} During both training and evaluation, we randomize the red cup position and the liquid volume in the dropper. We decompose success into visual and contact components. Visual success evaluates whether the robot first moves toward the correct target cup specified by the language instruction and then returns the dropper to the red cup. Contact success evaluates whether the robot transfers liquid into one of the cups without spilling. A trial is counted as an overall task success only when both visual and contact success are achieved.

\section{Implementation Details}
\label{app:implement_details}
\subsection{Base Policy}
\label{app:implement_base_policy}

\para{Architecture Overview}
  The base policy is a \emph{Diffusion UNet} backbone~\cite{chi2024diffusionpolicy}
  that conditions on multi-camera RGB images and robot proprioception to produce a chunk of future
  actions via DDIM denoising. We use the same Conditional 1D UNet from  prior work~\cite{chi2024diffusionpolicy}. Details of the architecture are shown in Table.~\ref{tab:dp_arch}.

  \para{Observation Encoders}
  Each camera image ($224\times224$) is processed independently by a
  ResNet-18~\cite{he2016identity} backbone (ImageNet pre-trained) with all batch normalization layers
  replaced by GroupNorm (groups $= C/16$, following the Diffusion Policy convention), with the
  global-average-pool and classification head removed.
  The resulting $512\times7\times7$ feature map is passed through a \emph{Spatial Softmax}
  pooling layer (64 keypoints, temperature 1.0) and flattened to $64\times2=128$ values, which are
  then projected to a 64-dimensional embedding by a linear layer. For visual policy, two cameras (wrist, front-view) are each encoded by a \emph{separate} copy of this
  network, so the per-timestep visual embedding is $64\times2=128$\,dim.
  For visuo-tactile policy, three cameras (wrist, front-view, tactile) are encoded so the per-timestep visual embedding is $64\times3=192$\,dim.

  \para{Observation encoding}
  Two consecutive image frames per camera (\texttt{img\_chunk}$=2$) are stacked, giving a total
  visual embedding of $64\times3\times2=384$\,dim.
  This is concatenated with the 8-dimensional proprioceptive state
  (end-effector 3D position + quaternion + gripper width), yielding a
  $384+8=392$\,dim global conditioning vector $\mathbf{s}_t$.
  No shared MLP is applied; a dropout layer (rate 0.1) is the only non-linearity.

  \begin{table}[h]
  \centering
  \caption{\textbf{Diffusion Policy Architecture Hyperparameters}.}
  \label{tab:dp_arch}
  \begin{tabular}{lc}
  \toprule
  \textbf{Hyperparameter} & \textbf{Value} \\
  \midrule
  \multicolumn{2}{l}{\textit{Visual encoder}} \\
  Backbone            & ResNet-18 (ImageNet pre-trained) \\
  Normalization       & GroupNorm ($C/16$ groups) \\
  Pooling             & Spatial Softmax (64 keypoints) \\
  Per-camera embed dim & 64 \\
  \midrule
  \multicolumn{2}{l}{\textit{Observation encoding}} \\
  Cameras             & 3 (wrist, front, tactile) \\
  Image frames per camera (\texttt{img\_chunk}) & 2 \\
  Visual embed dim    & 384 ($64\times3\times2$) \\
  Proprioception dim  & 8 \\
  Total conditioning dim & 392 \\
  \midrule
  \multicolumn{2}{l}{\textit{Diffusion U-Net}} \\
  Action chunk (\texttt{ac\_chunk})  & 16 \\
  Action dim (\texttt{ac\_dim})      & 7 \\
  U-Net channel dims                 & $[256,\;512,\;1024]$ \\
  Diffusion step embed dim           & 256 \\
  FiLM conditioning dim              & 648 (step emb.\ $+$ obs.) \\
  Kernel size                        & 3 \\
  GroupNorm groups                   & 8 \\
  Activation                         & Mish \\
  \midrule
  \multicolumn{2}{l}{\textit{DDIM noise schedule}} \\
  Beta schedule                      & Squaredcos cap v2 \\
  $\beta_{\min}$ / $\beta_{\max}$    & $10^{-4}$ / $0.02$ \\
  Training diffusion steps           & 100 \\
  Inference diffusion steps          & 16 (DDIM) \\
  Prediction type                    & Epsilon ($\hat{\boldsymbol\epsilon}$) \\
  Clip sample                        & Yes ($[-1, 1]$) \\
  \bottomrule
  \end{tabular}
  \end{table}
  
  \para{Training}
  Images are augmented on-GPU with random-crop shifts (ImageNet normalization applied after).
  diffusion timestep $k \sim \mathrm{Uniform}[0, 100)$ is sampled, the action chunk is noised to k steps and the U-Net predicts the added noise. The details of the training parameters are in Table.~\ref{tab:dp_train}.
  
  \begin{table}[h]
  \centering
  \caption{\textbf{Diffusion Policy Training Hyperparameters}.}
  \label{tab:dp_train}
  \begin{tabular}{lc}
  \toprule
  \textbf{Hyperparameter} & \textbf{Value} \\
  \midrule
  Optimizer               & AdamW \\
  Learning rate           & $1\times10^{-4}$ \\
  Betas                   & $(0.95,\;0.999)$ \\
  Weight decay            & $1\times10^{-6}$ \\
  Epsilon                 & $1\times10^{-8}$ \\
  LR schedule             & Cosine decay \\
  Warmup steps            & 500 \\ 
  Training iterations     & 80{,}000 \\
  Batch size              & 100 \\
  Image augmentation      & random crop (shift) + ImageNet norm \\
  Dropout                 & 0.1 \\
  \bottomrule
  \end{tabular}
\label{tab:base_policy_training}
  \end{table}
\subsection{Visuo-Tactile World Model}
\label{app:implement_visuo_tactile_world_model}
 \para{Observation Encoders}
  We use frozen pre-trained encoders to extract patch-level embeddings from each observation stream.
  RGB images from the wrist camera  and front camera are encoded using
DINOv3~\cite{simeoni2025dinov3} with a ViT-B/16 backbone, producing a 14$\times$14 grid of 196 patch tokens each
  of dimension 768.
  Tactile images are encoded with AnyTouch2~\cite{fenganytouch2}, a TactileVideoMAE model
  that processes a sliding window of the last 4 consecutive tactile frames and outputs
  512-dimensional patch embeddings; the spatial grid is standardized to 196 patches by truncation or
  zero-padding.
  Images are resized to $224\times224$ before encoding; tactile frames additionally undergo
  background subtraction (offset 0.130) and CLIP normalization.
  All encoder weights are \emph{frozen} during world model training. The specifications of the encoders are in Table.~\ref{tab:observation_encoders}.

  \begin{table}[h]
  \centering
  \caption{\textbf{Observation Encoder Specifications.}}
  \label{tab:encoders}
  \begin{tabular}{llcc}
  \toprule
  \textbf{Camera} & \textbf{Model} & \textbf{Emb.\ dim} & \textbf{Patches} \\
  \midrule
  wrist-view RGB & DINOv3 ViT-B/16 & 768 & 196 ($14{\times}14$) \\
  front-view RGB & DINOv3 ViT-B/16 & 768 & 196 ($14{\times}14$) \\
  tactile image   & AnyTouch2 TactileVideoMAE & 512 & 196 ($14{\times}14$) \\
  \bottomrule
  \end{tabular}
  \label{tab:observation_encoders}
  \end{table}
  
  \para{Latent Dynamics Model Architecture}
  The world model is a causal patch-level transformer that maps
  one history frame of multi-modal observations and the corresponding collapsed action to a
  predicted next-frame embedding for every camera.

  Each camera's patch embeddings are projected from their native dimension to a common dimension of
  384 via a linear layer followed by LayerNorm and ReLU.
  Actions are represented as a block of $F=8$ consecutive raw actions per frame, each of dimension 7
  (6-DoF end-effector delta $+$ gripper), giving a per-frame action vector of dimension 56.
  This is encoded by a two-layer MLP to a 64-dimensional action embedding.
  The end-effector state (3D position, quaternion, gripper; dimension 8) is used directly.
  The projected camera tokens, action embedding, and state embedding are concatenated along the
  feature axis for each patch position, yielding tokens of total dimension $D_{\text{total}} =
  3{\times}384 + 64 + 8 = 1224$.
  Learned spatial positional embeddings of the same
  dimension are added.

  The transformer backbone consists of 6 identical blocks, each comprising multi-head self-attention
  (16 heads, 64 dimensions per head) with a causal mask that prevents cross-frame look-ahead,
  followed by a two-layer feed-forward network (hidden dimension 2048) with GELU activation and
  residual connections; dropout of 0.1 is applied throughout.
  Only the last context frame's output tokens are passed to the prediction heads.
  Separate two-layer MLP heads decode the transformer output to each camera's native embedding
  dimension; a mean-pooling state head predicts the next robot state. Details of the architecture of the latent dynamics model are in Table~\ref{tab:latent_dyn_arch}.

  \begin{table}[h]
  \centering
  \caption{\textbf{Latent Dynamics Model Architecture Hyperparameters}.}
  \label{tab:wm_arch}
  \begin{tabular}{lc}
  \toprule
  \textbf{Hyperparameter} & \textbf{Value} \\
  \midrule
  Camera projection (common) dim & 384 \\
  Action encoding dim             & 64 \\
  State dim                       & 8 \\
  Frame skip $F$                  & 8 \\
  Action input dim (per slot)     & 56 ($F{\times}7$) \\
  Total token dim $D_{\text{total}}$ & 1224 \\
  Transformer depth               & 6 \\
  Attention heads                 & 16 \\
  Dim per head                    & 64 \\
  MLP hidden dim                  & 2048 \\
  Dropout                         & 0.1 \\
  History frames (\texttt{num\_hist}) & 1 \\
  Prediction frames (\texttt{num\_pred}) & 1 \\
  \bottomrule
  \end{tabular}
  \label{tab:latent_dyn_arch}
  \end{table}

  \para{Latent Dynamics Model Training}
  Actions are normalized per-dimension using dataset mean and standard deviation before collapsing.
  We use the \emph{multi-step rollout loss}: at each training iteration a rollout horizon
  $k \sim \mathrm{Uniform}\{3, 4, 5\}$ is sampled; the model is unrolled autoregressively for $k$
  steps by feeding each predicted embedding back as the next observation, and MSE loss is accumulated
  at every step over all cameras and the robot state.
To improve robustness to noised actions during inference time for classifier-based guidance, we follow prior work~\cite{du2025dynaguidesteeringdiffusionpolices} to additionally apply DDIM-schedule noise to the
  input action tokens: a per-sample diffusion timestep is drawn from $\mathrm{Geometric}(p{=}0.05)$
   and a linearly-ramped masking probability (reaching 0.5 at the final epoch)
  randomly replaces clean actions with their noised version.

  Observations are subsampled at every $F{=}8$ raw steps following prior work~\cite{sun2026latent}, so each subsampled
  frame represents 8 environment steps.
  The train/test split holds out 30 trajectories; the remaining trajectories are used for training.
  Training uses mixed-precision (FP16 AMP) with torch compile (reduce-overhead mode). The details of the training are in Table~\ref{tab:latent_dyn_training}.

  \begin{table}[h]
  \centering
  \caption{\textbf{Latent Dynamics Model Training Hyperparameters}.}
  \label{tab:wm_train}
  \begin{tabular}{lc}
  \toprule
  \textbf{Hyperparameter} & \textbf{Value} \\
  \midrule
  Optimizer                          & AdamW \\
  Transformer LR                     & $5{\times}10^{-5}$ \\
  Action encoder \& positional emb.\ LR & $5{\times}10^{-4}$ \\
  Projection \& prediction head LR   & $5{\times}10^{-5}$ \\
  Training iterations                & 100{,}000 \\
  Batch size                         & 16 \\
  Test split trajectories            & 50 \\
  Loss                               & Multi-step MSE \\
  Rollout horizon $k$                & Uniform$\{3,4,5\}$ \\
  Action noise scheduler             & DDIM \\
  Action noise $p$ (geometric)       & 0.05 \\
  DDIM timesteps                     & 100 \\
  Frame skip $F$                     & 8 \\
  Temporal sampling                  & uniform subsampling \\
  Mixed precision                    & FP16 (AMP) \\
  Random seed                        & 42 \\
  \bottomrule
  \end{tabular}
  \label{tab:latent_dyn_training}
  \end{table}

  \para{Image Decoder Training}
  We train a convolutional image decoder to reconstruct RGB/tactile images from predicted patch
  embeddings, used only for visualization and evaluation.
  The decoder is a VQVAE-style network  with a stride-4
  convolutional decoder: a $3{\times}3$ input projection, 4 residual blocks (channel width 384,
  residual channels 128), followed by two transposed convolutions (stride 2) that upsample the
  $14{\times}14$ patch grid to $56{\times}56$ and then $224{\times}224$.
  Separate decoders are trained for the vision cameras (wrist-view and front-view share one decoder with
  $d_{\text{emb}}{=}768$) and the tactile camera uses $d_{\text{emb}}{=}512$.

  Training uses the RAE loss~\cite{zheng2025diffusion}: $\mathcal{L} = \mathcal{L}_{L1} + \lambda_{\text{LPIPS}}
  \mathcal{L}_{\text{LPIPS}} + \lambda_{\text{GAN}} \mathcal{L}_{\text{GAN}}$.
  An adaptive GAN weight (gradient norm balancing between reconstruction and adversarial gradients,
  clamped to $[0.01, 100]$) is applied per camera using per-camera PatchGAN discriminators
  (ndf$=$64, 3 layers) with hinge loss; the discriminator is warmed up for 200 iterations before
  the adversarial loss enters the generator objective. We use RAE loss instead of L1 loss or L2 loss because empirically we find RAE loss is good at reconstructing the fine-grained details such as the marks on the board compared to L1 loss or L2 loss because marks only exist in a few pixels in the front-view image and therefore has very low penalty if the prediction is wrong. Details of the Decoder Architecture and Training are in Table.~\ref{tab:decoder_training}.

  \begin{table}[h]
  \centering
  \caption{\textbf{Image Decoder Architecture and Training Hyperparameters}.}
  \label{tab:decoder}
  \begin{tabular}{lc}
  \toprule
  \textbf{Hyperparameter} & \textbf{Value} \\
  Architecture            & VQVAE decoder  \\
  Vision decoder emb.\ dim & 768 \\
  Tactile decoder emb.\ dim & 512 \\
  Channel width           & 384 \\
  Residual blocks         & 4 \\
  Residual channels       & 128 \\
  Spatial stride          & 4 (two ${\times}2$ upsample stages) \\
  Output resolution       & $224{\times}224$ \\
  \midrule
  Loss                    & RAE (L1 + LPIPS + Adversarial) \\
  $\lambda_{L1}$          & 1.0 \\
  $\lambda_{\text{LPIPS}}$ & 1.0 \\
  $\lambda_{\text{GAN}}$  & 0.75 (adaptive per camera) \\
  Discriminator warm-up   & 600 iters (disc.\ train) / 800 iters (gen.\ adv.) \\
  Discriminator           & PatchGAN (ndf$=$64, 3 layers, hinge loss) \\
  Optimizer (decoder)     & AdamW, lr$=3{\times}10^{-4}$ \\
  Optimizer (discriminator) & AdamW, lr$=3{\times}10^{-4}$, $\beta=(0.5, 0.9)$ \\
  Batch size              & 64 \\
  Training iterations     & 5{,}000 \\
  \bottomrule
  \end{tabular}
  \label{tab:decoder_training}
  \end{table}

\begin{algorithm}[t]
\caption{\ours: Bi-level Visuo-Tactile Policy Steering}
\label{alg:vital}
\scriptsize
\begin{algorithmic}[1]
\Require Policy $\pi_\theta$, latent world model $p_\phi$, encoders $\enc^v,\enc^\tau$,
decoders $\dec^v,\dec^\tau$, verifiers $R^v,R^\tau$, instruction $\task$
\Require Visual samples $N$, visual horizon $H$, execution horizon $h$,
edit level $K$, guidance scale $\lambda$
\State Decompose $\task$ into phase objectives
$\{(\ell^v_p,\ell^\tau_p)\}_{p=1}^P \gets g_{\mathrm{VLM}}(\task)$
\State Initialize phase $p\gets 1$, number of visual chunks $M\gets H/h$
\While{task not finished}
    \State Observe $o_t=(o^v_t,o^\tau_t,q_t)$ and encode
    $z_t=(\enc^v(o^v_t),\enc^\tau(o^\tau_t))$

    \Statex \textbf{Outer level: batched visual mode selection}
    \State Initialize $N$ imagined rollouts:
    $\hat o^{(i)}_t\gets o_t$, $\hat z^{(i)}_t\gets z_t$,
    $A^{(i)}\gets [\,]$ for all $i\in\{1,\dots,N\}$
    \For{$j=0,\dots,M-1$}
        \State Sample a batch of action chunks:
        $a^{(1:N)}_j \sim \pi_\theta(\cdot \mid \hat o^{(1:N)}_{t+jh})$
        \State Roll out each chunk with the latent world model:
        $\hat z^{(1:N)}_{t+(j+1)h}
        \sim p_\phi(\cdot \mid \hat z^{(1:N)}_{t+jh}, a^{(1:N)}_j)$
        \State Decode the terminal imagined observation for the next policy call:\\
        \qquad $\hat o^{(i)}_{t+(j+1)h}
        \gets (\dec^v(\hat z^{v,(i)}_{t+(j+1)h}),
        \dec^\tau(\hat z^{\tau,(i)}_{t+(j+1)h}),  \dec^q(\hat z^{q,(i)}_{t+(j+1)h})$
        \State Append chunk to candidate sequence:
        $A^{(i)}\gets A^{(i)} \Vert a^{(i)}_j$ for all $i$
    \EndFor
    \State Score each candidate using the decoded final visual frame:
    $s_i^v\gets R^v(\hat o^{v,(i)}_{t+H};\ell^v_p)$ for all $i$
    \State Select visual anchor:
    $\bar a_{t:t+H}\gets A^{(i^\star)}$,
    where $i^\star=\arg\max_i s_i^v$

    \Statex \textbf{Inner level: tactile contact refinement by diffusion editing}
    \State Let $\bar a\gets \bar a_{t:t+h}$ be the first $h$ actions of the visual anchor
    \State Partially noise the visual anchor:
    $x_K\gets \sqrt{\bar\alpha_K}\bar a+
    \sqrt{1-\bar\alpha_K}\epsilon$, where $\epsilon\sim\mathcal N(0,I)$
    \For{$k=K,\dots,1$}
        \State Predict diffusion noise:
        $\epsilon^k_\theta\gets \epsilon_\theta(x_k,k,o_t)$
        \State Estimate clean action:
        $\hat a^k_0\gets
        (x_k-\sqrt{1-\bar\alpha_k}\epsilon^k_\theta)/\sqrt{\bar\alpha_k}$
        \State Predict tactile future under the current clean-action estimate:
        $\hat z^\tau_{t:t+h}\sim p_\phi^\tau(\cdot \mid z_t,\hat a^k_0)$
        \State Compute phase-conditioned tactile reward:
        $r^\tau_k\gets R^\tau(\hat z^\tau_{t:t+h};\ell^\tau_p)$
        \State Compute normalized tactile guidance:
        $g_k\gets \nabla_{\hat a^k_0} r^\tau_k /
        (\|\nabla_{\hat a^k_0} r^\tau_k\|_2+\varepsilon)$
        \State Bias the reverse diffusion direction:
        $\tilde\epsilon_k\gets
        \epsilon^k_\theta-\lambda\sqrt{1-\bar\alpha_k}\,g_k$
        \State DDIM update:
        $x_{k-1}\gets
        \sqrt{\bar\alpha_{k-1}}\hat a^k_0+
        \sqrt{1-\bar\alpha_{k-1}}\tilde\epsilon_k$
    \EndFor
    \State Execute refined action chunk $a^\star_{t:t+h}\gets x_0$
    \State Update phase $p$ when the visual verifier satisfies $\ell^v_p$
\EndWhile
\end{algorithmic}
\end{algorithm}

\subsection{Visual Steering as Sampling \& Verification}
\label{app:implement_visual_steering}

Algo.~\ref{alg:vital} specifies the bi-level policy steering algorithm. We begin by discussing about the outer level of visual steering.

At each policy query step, we sample $N$ candidate long-horizon action sequences from the base diffusion policy,
\begin{equation}
\mathbb{A}_N
=
\left\{
\action_{t:t+H}^{(i)}
\mid
\action_{t:t+H}^{(i)}
\sim
\pi_\theta(\cdot \mid \obs_t)
\right\}_{i=1}^{N}.
\label{eq:visual_candidate_set}
\end{equation}
The candidates are generated in parallel using DDIM denoising. Each candidate is then evaluated by rolling it out with the latent visuo-tactile world model. Specifically, given the current visual and tactile embeddings
$\latentv_t=\encv(\obsv_t)$ and $\latentt_t=\enct(\obst_t)$, the world model predicts the future visual and tactile latent trajectories:
\begin{equation}
\latent_{t:t+H}^{v,(i)}, \latent_{t:t+H}^{\tau,(i)}
\sim
p_\phi
\left(
    \cdot,\cdot
    \mid
    \latentv_t,
    \latentt_t,
    \action_{t:t+H}^{(i)}
\right).
\label{eq:visual_wm_rollout}
\end{equation}
The visual reward model then scores each predicted visual future against the phase-specific visual language goal $\lang_p^v$. We select the candidate with the highest visual reward:
\begin{equation}
\bar{\action}_{t:t+H}
=
\arg\max_{\action_{t:t+H}^{(i)} \in \mathbb{A}_N}
R^v
\left(
    \latent_{t:t+H}^{v,(i)};
    \lang_p^v
\right).
\label{eq:visual_candidate_selection}
\end{equation}
This selected sequence defines the visual anchor. Its first $h$ steps,
$\bar{\action}_{t:t+h}$, are passed to the tactile diffusion editing stage in
Appendix~\ref{app:implement_tactile_steering}.

In implementation, we use recurrent world-model imagination and policy inference to obtain longer-horizon predictions. The policy predicts the first action chunk from the current observations and the world model consumes the current embedding history together with the first action chunk and predicts future latent embeddings. We decode the future latent embeddings into the observations and query the policy with the decoded observations to get the next action chunk. We retain the final-slot prediction, shift the context window forward by replacing the oldest context frame with the predicted embedding, and query the world model again with the next action chunk. This recurrent rollout produces the long-horizon latent future used by the visual verifier.

For visual scoring, we decode the predicted camera-view latent corresponding to the final imagined step into a $224 \times 224$ RGB image using the frozen VQ-VAE decoder. The decoded image is sent to a ROBOMETER~\cite{liang2026robometer}, which scores each candidate image against the task description using a 4B vision-language model. Since direct gradient-based guidance through the VLM, decoder, recurrent world model, and policy sampling process would be expensive and potentially unstable, we use the VLM only as a verifier to rank sampled futures.

We made the modifications to the original ROBOMETER code for online evaluation of the visual reward.  To reduce per-step latency during online evaluation, we introduce some optimization to the visual reward inference. We introduce a sliding-window context eviction policy: rather than
  accumulating the full observation history in the model's context, only the most recent $K$ frames are retained, bounding per-step inference latency to a constant independent of trajectory length.

We also support a two-phase task structure. The robot begins in phase~1 with the visual language goal $\lang_1^v$, such as ``transfer the liquid to the yellow cup.'' Once the maximum visual reward across candidates exceeds a threshold, the task switches to phase~2 with a new visual language goal $\lang_2^v$, such as ``put the dropper back to the red cup.'' A new ROBOMETER session is created for the second phase. Details of the hyperparameters used in visual steering are shown in Table~\ref{tab:visual_guidance}.

\begin{table}[h]
\centering
\caption{\textbf{Visual Steering Hyperparameters.}}
\label{tab:visual_guidance}
\begin{tabular}{lc}
\toprule
\textbf{Hyperparameter} & \textbf{Value} \\
\midrule
Candidates sampled ($N$) & 10 \\
Action horizon per chunk ($h$) & 8 \\
Number of recurrent chunks & 2 \\
Total visual lookahead horizon ($H$) & 16 \\
WM reward camera & front-view camera (RGB) \\
Visual reward model & ROBOMETER-4B \\
Phase-switch visual reward threshold & 0.70 \\

\bottomrule
\end{tabular}
\end{table}

\subsection{Tactile Steering as Diffusion Editing}
\label{app:implement_tactile_steering}
\label{app:tactile_guidance}

After the visual stage selects the best long-horizon candidate 
$\bar{\action}_{t:t+H}$ from Eq.~\ref{eq:bilevel_joint_annotated}, we begin the inner level optimization in Algo.~\ref{alg:vital} and take its first
$h$ steps, denoted as the visual anchor $\bar{\action}_{t:t+h}$, and refine it
using tactile-guided diffusion editing. The goal of this refinement is to improve
short-horizon contact quality while preserving the global behavior mode selected
by visual steering.

Following Eq.~\ref{eq:bilevel_joint_annotated} and prior diffusion editing
methods~\cite{meng2021sdedit}, we define a reference-conditioned action prior
around the visual anchor:
\begin{equation}
p_\theta
\left(
    \action_{t:t+h}
    \mid
    \bar{\action}_{t:t+h}, \obs_t
\right)
:=
\int
\rho_\theta
\left(
    \action_{t:t+h}
    \mid
    \mathbf{x}_{K}, \obs_t
\right)
\nu_K
\left(
    \mathbf{x}_{K}
    \mid
    \bar{\action}_{t:t+h}
\right)
d\mathbf{x}_{K},
\label{eq:reference_conditioned_prior}
\end{equation}
where $\nu_K$ partially noises the visual anchor to diffusion level $K$, and
$\rho_\theta$ denotes the reverse denoising process induced by the base diffusion
policy. The diffusion level $K$ controls the edit strength: smaller $K$ keeps the
sample closer to the visual anchor, while larger $K$ permits larger local
deviations.

\paragraph{Partial re-noise.}
We first map the visual anchor $\bar{\action}_{t:t+h}$, represented in the
policy's normalized $[-1,1]$ action space, to a mildly noised action latent at the
$K$-th-from-last DDIM timestep $K$:
\begin{equation}
\mathbf{x}_{K}
=
\sqrt{\bar\alpha_{K}}\,
\bar{\action}_{t:t+h}
+
\sqrt{1-\bar\alpha_{K}}\,
\varepsilon,
\qquad
\varepsilon \sim \mathcal{N}(0,I).
\label{eq:renoise}
\end{equation}
This step initializes the reverse process near the visually selected action mode
rather than sampling from pure noise.

\paragraph{Tactile-guided denoising.}
At each of the $K$ guided DDIM steps, we reconstruct the clean action estimate
from the current noisy action $\mathbf{x}_{K}$ using the stop-gradient U-Net
noise prediction $\hat{\varepsilon}_\theta$:
\begin{equation}
\hat{\action}_{0}^{(K)}
=
\frac{
    \mathbf{x}_{K}
    -
    \sqrt{1-\bar\alpha_{K}}\,
    \hat{\varepsilon}_\theta
    \left(
        \mathbf{x}_{K}, K, \obs_t
    \right)
}{
    \sqrt{\bar\alpha_{K}}
}.
\label{eq:x0hat}
\end{equation}
We then pass the estimated clean action chunk through the tactile branch of the
latent world model, conditioned on the current visual and tactile embeddings, to
predict the short-horizon tactile future:
\begin{equation}
\tilde{\latent}^{\tau}_{t:t+h}
\sim
p_\phi
\left(
    \cdot
    \mid
    \latentv_t,
    \latentt_t,
    \hat{\action}_{0}^{(n)}
\right),
\qquad
\latentv_t=\encv(\obsv_t),
\quad
\latentt_t=\enct(\obst_t).
\label{eq:tactile_future_prediction}
\end{equation}
The predicted tactile sequence is scored by the tactile reward model using the
phase-specific tactile language input $\lang_p^\tau$:
\begin{equation}
r^\tau
=
R^\tau
\left(
    \tilde{\latent}^{\tau}_{t:t+h};
    \lang_p^\tau
\right).
\label{eq:tactile_reward_guidance}
\end{equation}
Equivalently, this implements the refinement objective
\begin{equation}
\log p^{\mathrm{refine}}
\left(
    \action_{t:t+h}
\right)
=
\log p_\theta
\left(
    \action_{t:t+h}
    \mid
    \bar{\action}_{t:t+h}, \obs_t
\right)
+
\beta
R^\tau
\left(
    \tilde{\latent}^{\tau}_{t:t+h};
    \lang_p^\tau
\right),
\label{eq:refined_policy_distribution}
\end{equation}
where the first term keeps the edited action close to the visual anchor and the
second term biases the sample toward higher predicted tactile reward.

To approximately sample from this refined distribution, we compute the gradient
of the tactile reward with respect to the clean action estimate,
$\nabla_{\hat{\action}_{0}^{(K)}} r^\tau$, normalize it to unit norm, and use it
to modify the diffusion noise prediction:
\begin{equation}
\hat{\varepsilon}_{\mathrm{guided}}
=
\hat{\varepsilon}_\theta
-
\lambda
\sqrt{1-\bar\alpha_K}
\frac{
    \nabla_{\hat{\action}_{0}^{(K)}} r^\tau
}{
    \left\|
    \nabla_{\hat{\action}_{0}^{(K)}} r^\tau
    \right\|
}.
\label{eq:guided_eps}
\end{equation}
The scale factor $\sqrt{1-\bar\alpha_n}$ follows LPB guidance~\cite{sun2026latent} and
bounds the guidance magnitude across noise levels, avoiding the high-noise
gradient explosion that can arise in DynaGuide-style guidance~\cite{du2025dynaguidesteeringdiffusionpolices}.

We take one DDIM step using $\hat{\varepsilon}_{\mathrm{guided}}$ and repeat
Eqs.~\eqref{eq:x0hat}--\eqref{eq:guided_eps} for all $K$ reverse steps. The final
denoised action $\action^\star_{t:t+h}$ is unnormalized and executed on the robot.
In this way, the visual stage determines \emph{what} global behavior mode to
execute, while tactile diffusion editing refines \emph{how} to execute the first
action chunk with better contact. The detailed hyperparameters used in this process are shown in Table~\ref{tab:tactile_guidance}.
  \begin{table}[h]
  \centering
  \caption{\textbf{Tactile Guidance Hyperparameters.}}
  \label{tab:tactile_guidance}
  \begin{tabular}{lc}
  \toprule
  \textbf{Hyperparameter} & \textbf{Value} \\
  Re-noise / guided denoising steps ($K$)  & 4 \\
  Stochastic-sampling repetitions per step ($ss$) & 1 \\
  Guidance scale ($\lambda$)               & 10 \\
  Gradient normalization                   & Unit norm (FGSM) \\
  Gradient scale                           & $\sqrt{1-\bar\alpha_t}$ \\
  
  \bottomrule
  \end{tabular}
  \end{table}

\section{Ablations}
\label{app:ablations}
\subsection{Visuo-Tactile World Model}
\label{app:world_model_ablation}
\para{Methods} We compare three world model variants in Table~\ref{tab:world_model_eval_wipe},~\ref{tab:world_model_eval_insert},~\ref{tab:world_model_eval_pipette}: a tactile-only model, an vision-only model, and a visuo-tactile model. All unimodal baselines are trained with the same multi-step prediction loss as \ours. The multi-step loss supervises not only one-step dynamics prediction, but also autoregressive predictions over several future steps, encouraging the model to remain accurate under rollout. This is crucial for visual prediction since we need to predict longer horizon beyond a single action chunk. 

For the quantitative and qualitative evaluations in this section, predicted videos from the visuo-tactile world model are generated with sliding-window prediction rather than full-trajectory autoregressive rollout. At each time step, we condition the model on the current visual observation, the previous four tactile observations, and the next eight actions to predict the observation at the eighth future frame. We then advance one step and repeat the procedure with the next observation-action window. This protocol matches how the world model is used in \ours: after each steered eight-step action chunk is executed, the model receives a new real observation and predicts the next future outcome from that updated state. Thus, we evaluate the model under the same receding-horizon setting used for steering, rather than under a fully autoregressive full-trajectory generation setting.

\para{Visual Prediction Metrics} We evaluate predicted camera observations using six complementary metrics. Mean	
Squared Error (\textbf{MSE}) and Mean Absolute Error (\textbf{MAE}) measure pixel-level reconstruction accuracy, with MAE being			
less sensitive to large outliers. Peak Signal-to-Noise Ratio (\textbf{PSNR}, in dB) expresses MSE on a logarithmic scale,			
where higher values indicate better fidelity. Structural Similarity Index (\textbf{SSIM}) captures perceptual quality by			
jointly assessing luminance, contrast, and local structure, with values closer to 1 denoting greater similarity.			
Learned Perceptual Image Patch Similarity (\textbf{LPIPS}) computes perceptual distance using deep VGG features,			
correlating more closely with human judgement than pixel-level metrics; lower scores indicate greater perceptual			
similarity. Fréchet Inception Distance (\textbf{FID}) evaluates distributional alignment between predicted and			
ground-truth frames using Inception-V3 embeddings, where lower values indicate the predicted frames are			
statistically closer to the real distribution. For \textbf{MSE, MAE, LPIPS}, and \textbf{FID}, lower is better; for \textbf{PSNR} and \textbf{SSIM},			
higher is better.			
			
\para{Tactile Prediction Metrics} We evaluate predicted GelSight tactile observations using five metrics that target			
distinct aspects of visuotactile fidelity. Surface deformation patterns are assessed via optical flow: mean			
End-Point Error (\textbf{Flow EPE}) measures the average Euclidean distance between predicted and ground-truth flow vectors,			
and flow angular error in degrees (\textbf{Flow AE}) measures the mean directional deviation of the predicted deformation field; both			
are lower-is-better. Gradient cosine similarity (\textbf{Grad. Cos.}) computes the cosine similarity between the spatial gradient			
fields of predicted and ground-truth tactile images, reflecting the accuracy of contact edges and surface texture			
patterns, where higher values (maximum 1) are better. Finally, height map MAE (\textbf{Height MAE}) and MSE (\textbf{Height RMSE}) measure the pixel-wise			
reconstruction error of the 3D surface depth maps recovered via photometric stereo from the tactile images, with			
lower values indicating more accurate tactile shape prediction.	

\para{Results}
Across all three tasks, the visuo-tactile world model achieves comparable or better prediction performance than the unimodal baselines, indicating that joint visual-tactile modeling does not degrade either modality and often improves both. For RGB prediction, \ours{} consistently improves the distributional and perceptual metrics over the RGB-only world model. In particular, \ours{} obtains lower FID and LPIPS on both front and wrist cameras across all tasks, suggesting that incorporating tactile observations helps the model generate visually more realistic future predictions, even when pixel-level metrics such as MSE or PSNR are similar. This is especially important for visual steering, where candidate actions are ranked by a VLM over decoded future images rather than by pixel reconstruction alone.

For tactile prediction, the benefit of joint modeling is more pronounced. Compared with the tactile-only world model, \ours{} improves most tactile metrics across all tasks, including flow endpoint error, flow angular error, and height reconstruction error. The gains are particularly clear in the insertion and pipette transfer tasks, where precise contact dynamics are central to task success. In the insertion task, \ours{} substantially reduces both flow error and height-map error, indicating that visual context helps predict how the peg will make contact with the tight-clearance hole. In the pipette transfer task, \ours{} also improves all tactile metrics, suggesting that visual information about the dropper, cups, and liquid-transfer configuration provides useful context for forecasting contact and grasp behavior.

The wiping task shows smaller but still consistent improvements for most tactile and visual metrics. This is expected because wiping involves broader surface contact and less geometrically constrained contact than insertion, making the tactile future somewhat easier to predict from tactile history alone. Even in this case, the visuo-tactile model improves flow and height prediction while maintaining strong RGB prediction quality.

\begin{figure}
    \centering
    \includegraphics[width=\linewidth]{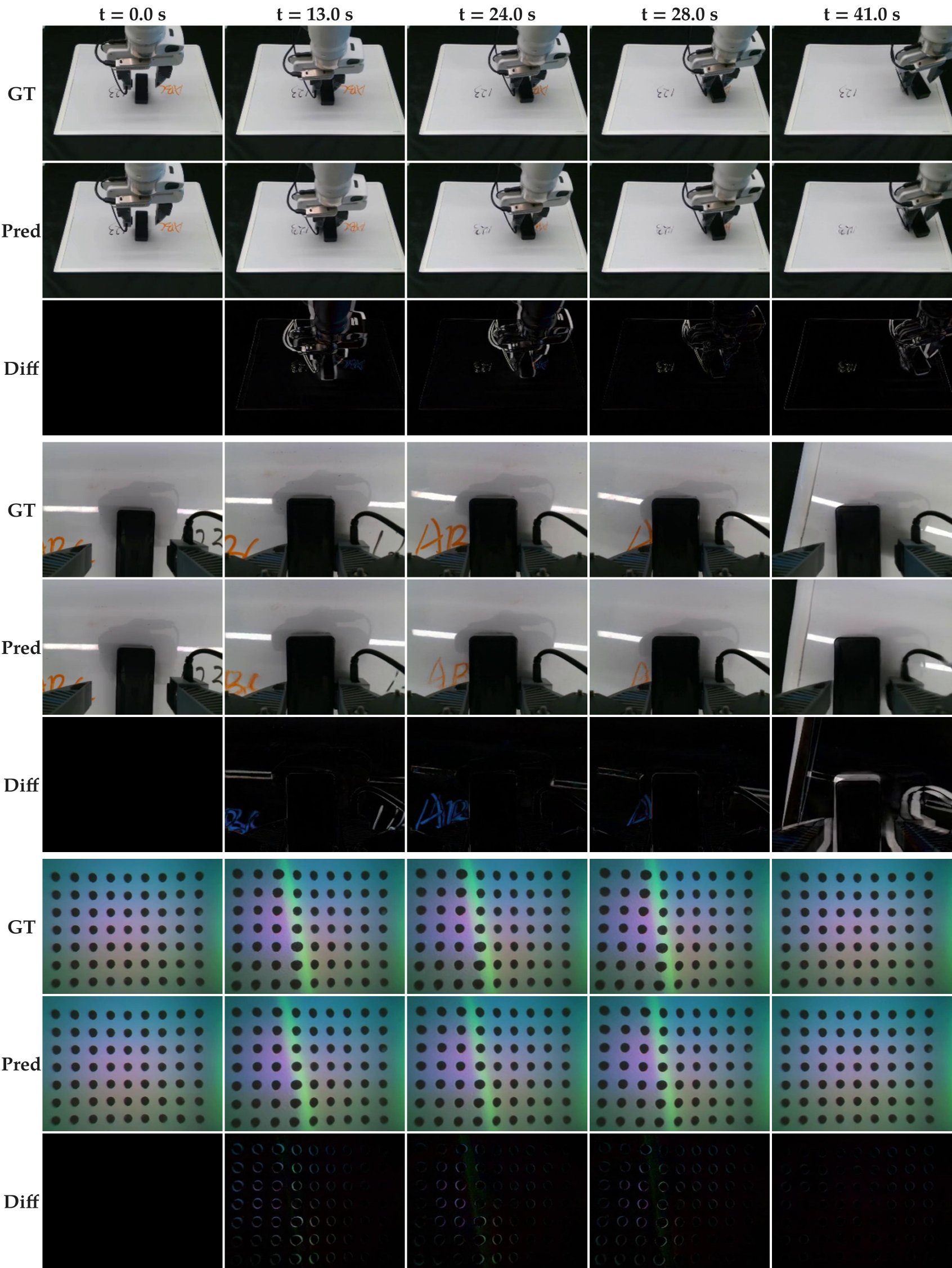}
    \caption{\textbf{\ours Visuo-Tactile World Model Rollouts For Successfully Wiping the Orange Marks From the Board}. We show the ground-truth rollouts, predicted rollouts for front view (top), wrist view (middle) and the gelsight Mini tactile image (bottom) as well as the differences between ground-truth and predictions.
    }
    \label{fig:world_model_wipe_orange_success}
\end{figure}

\begin{figure}
    \centering
    \includegraphics[width=\linewidth]{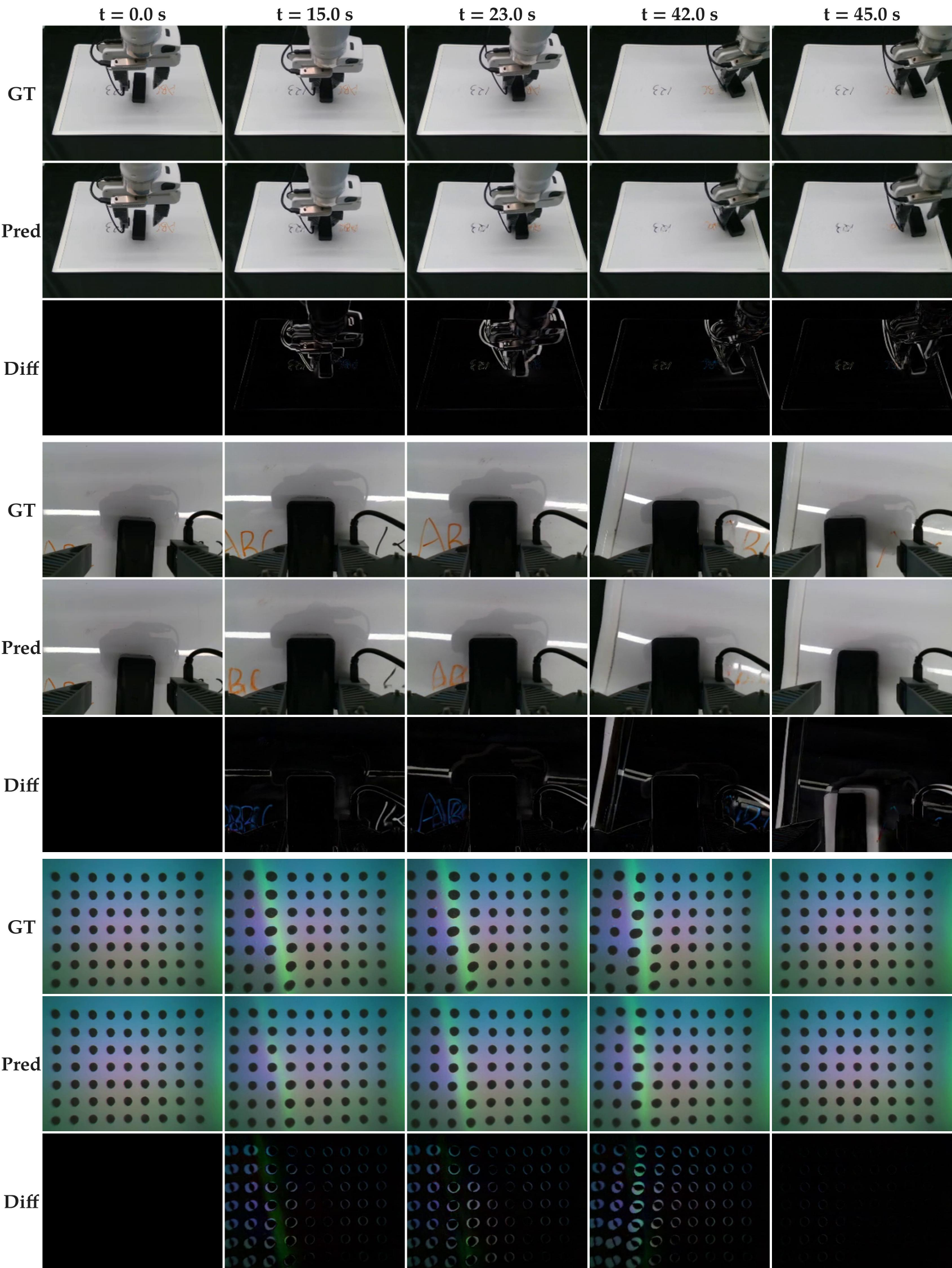}
    \caption{\textbf{\ours Visuo-Tactile World Model Rollouts For Failing to Completely Wipe the Orange Marks From the Board}. We show the ground-truth rollouts, predicted rollouts for front view (top), wrist view (middle) and the gelsight Mini tactile image (bottom) as well as the differences between ground-truth and predictions.
    }
    \label{fig:world_model_wipe_orange_failure}
\end{figure}

\begin{figure}
    \centering
    \includegraphics[width=\linewidth]{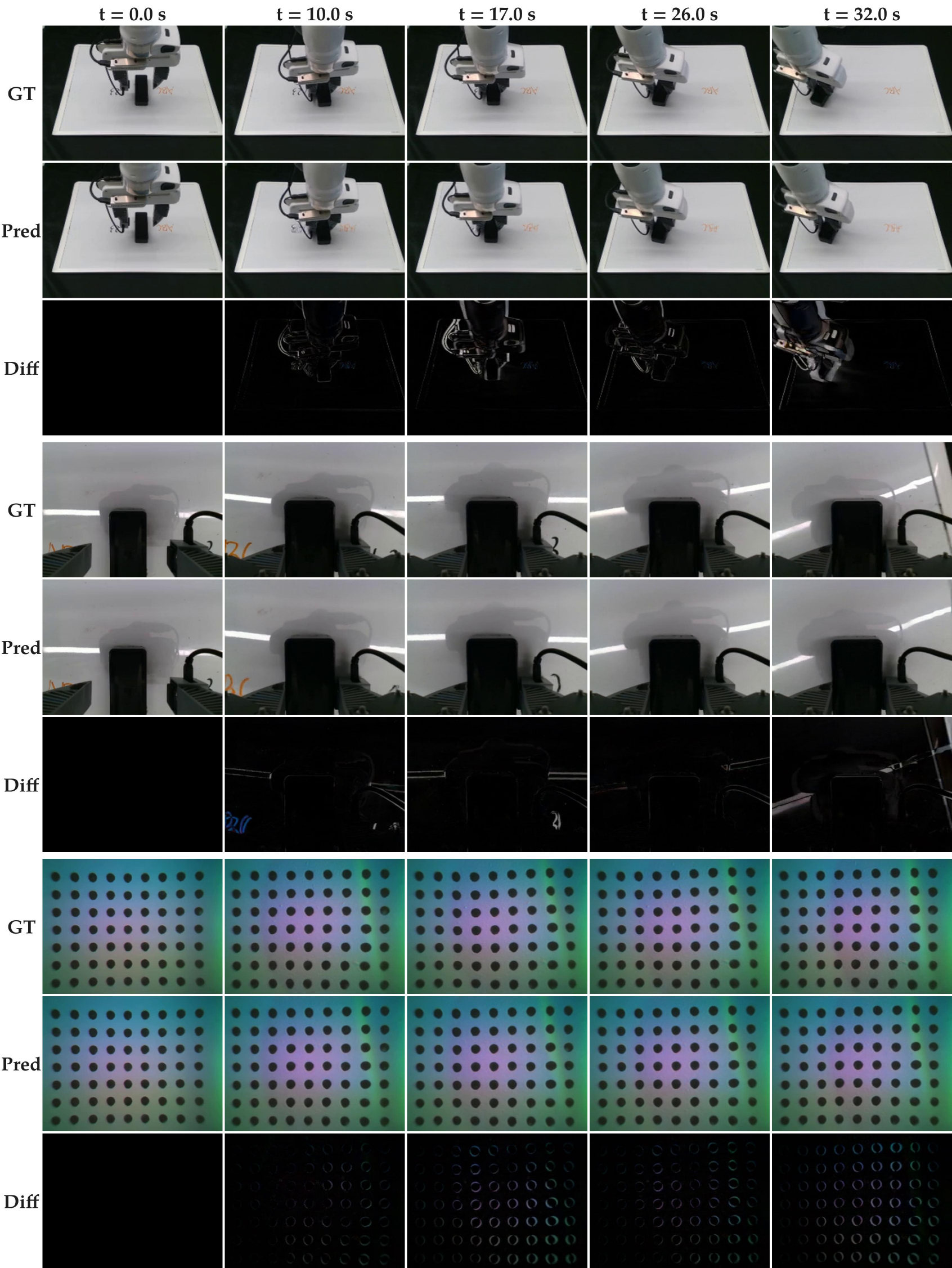}
    \caption{\textbf{\ours Visuo-Tactile World Model Rollouts For Successfully Wiping the Black Marks From the Board}. We show the ground-truth rollouts, predicted rollouts for front view (top), wrist view (middle) and the gelsight Mini tactile image (bottom) as well as the differences between ground-truth and predictions.
    }
\label{fig:world_model_wipe_black_success}
\end{figure}

\begin{figure}
    \centering
    \includegraphics[width=\linewidth]{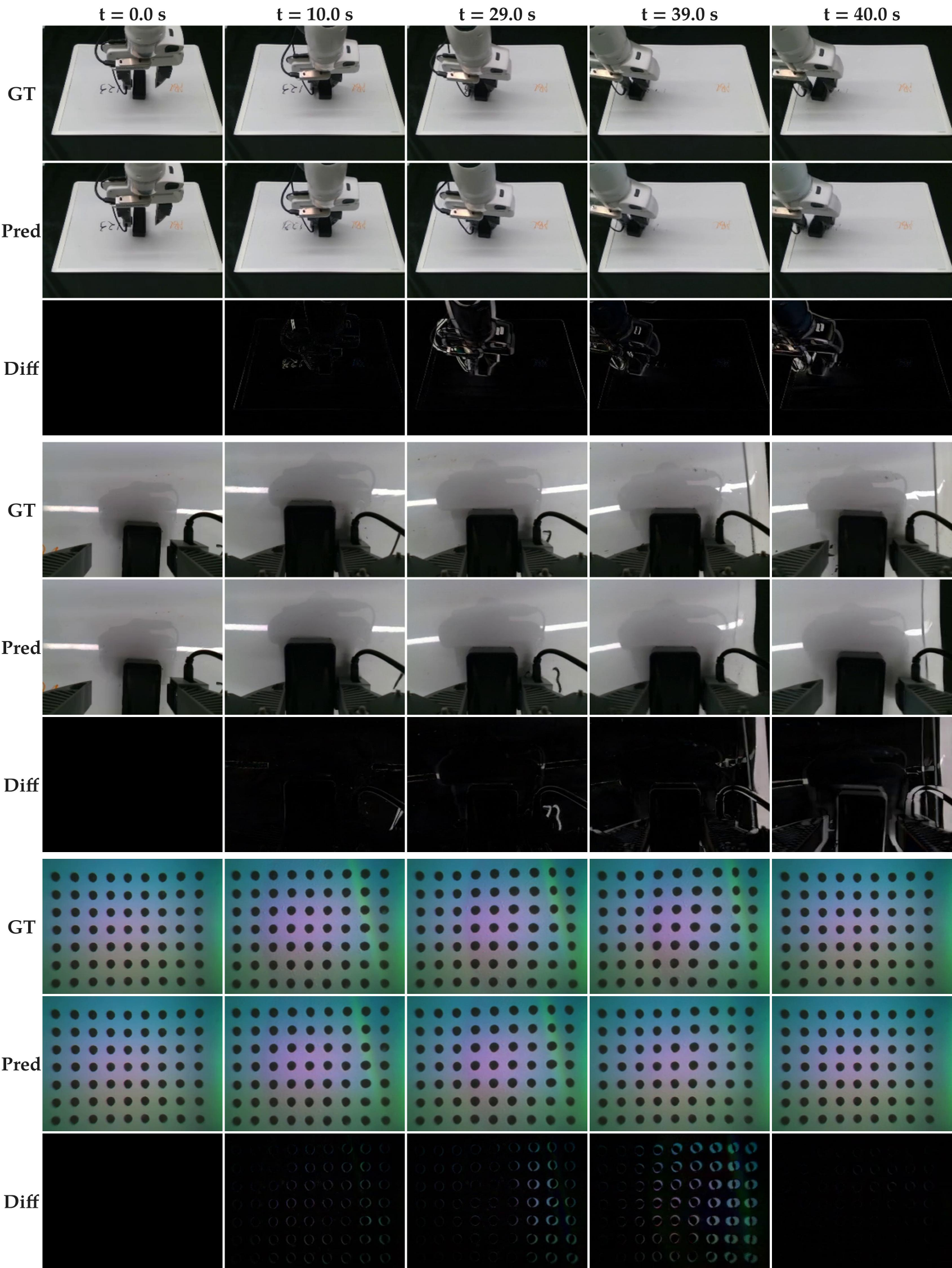}
    \caption{\textbf{\ours Visuo-Tactile World Model Rollouts For Failing to Wipe the Black Marks From the Board}. We show the ground-truth rollouts, predicted rollouts for front view (top), wrist view (middle) and the gelsight Mini tactile image (bottom) as well as the differences between ground-truth and predictions.
    }
    \label{fig:world_model_wipe_black_failure}
\end{figure}

\begin{figure}
    \centering
    \includegraphics[width=\linewidth]{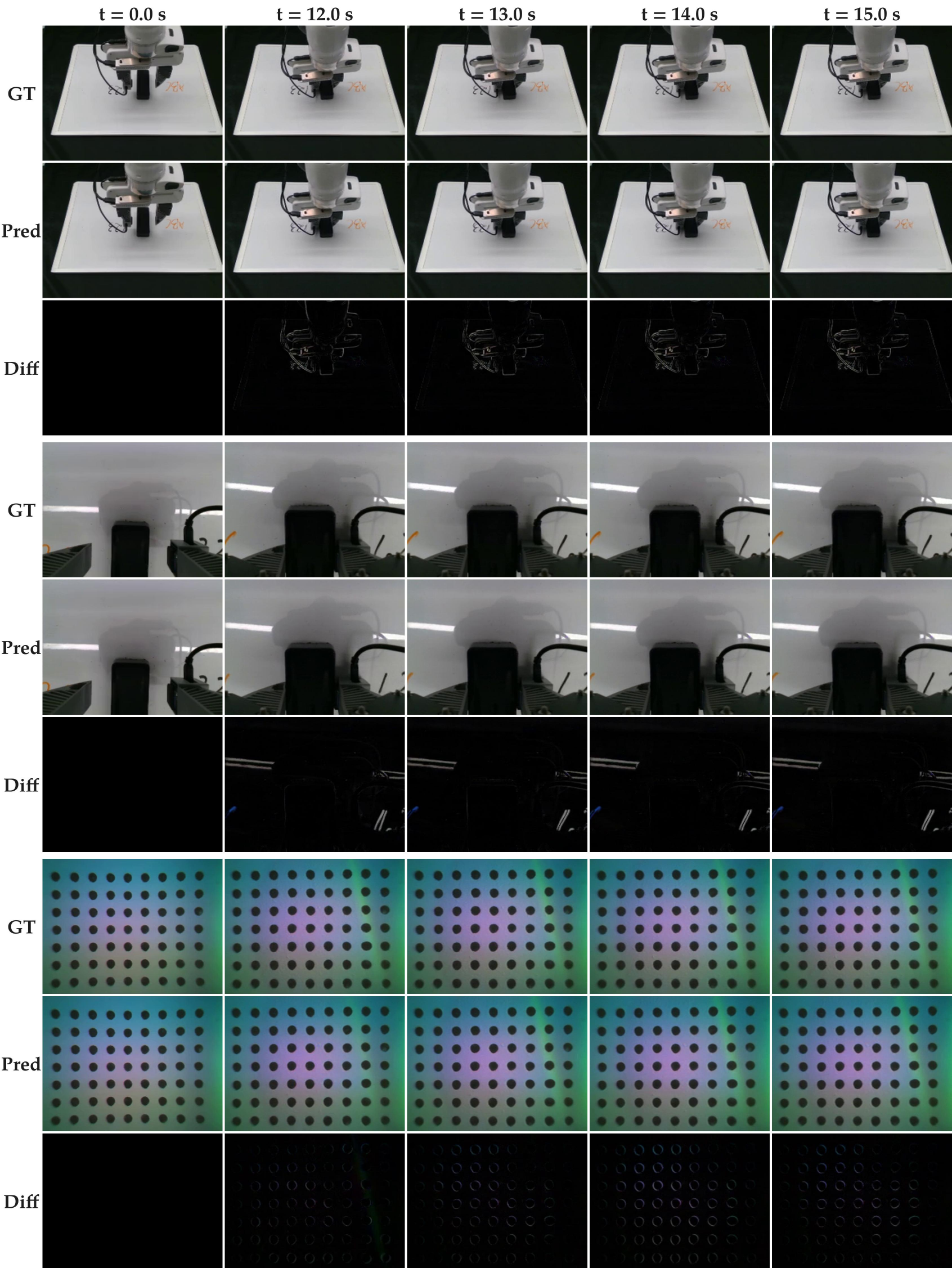}
    \caption{\textbf{\ours Visuo-Tactile World Model Rollouts For Getting Stuck in the Middle of the Board}. We show the ground-truth rollouts, predicted rollouts for front view (top), wrist view (middle) and the gelsight Mini tactile image (bottom) as well as the differences between ground-truth and predictions.
    }
    \label{fig:world_model_wipe_failure_stuck}
\end{figure}

\begin{figure}
    \centering
    \includegraphics[width=\linewidth]{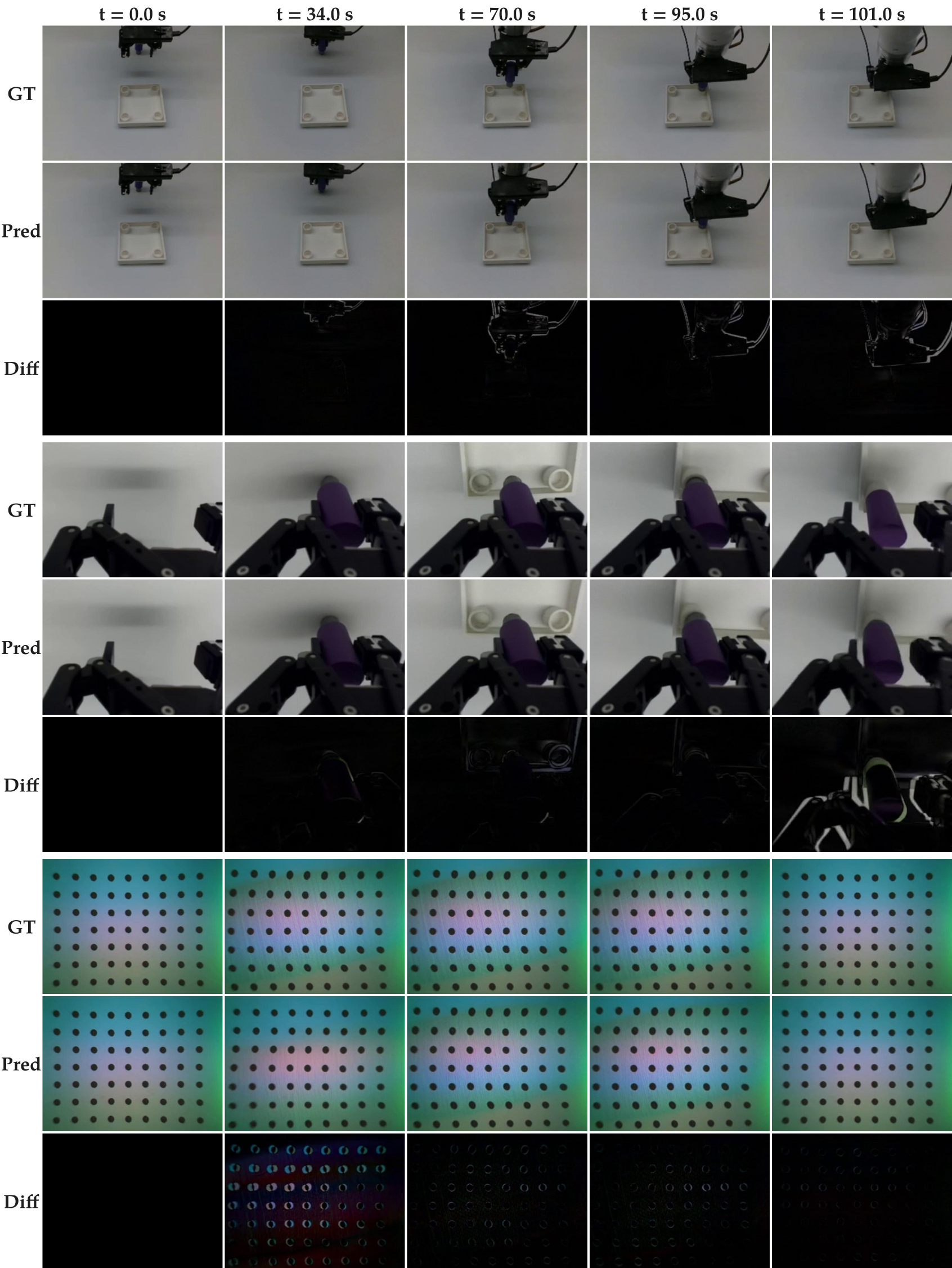}
    \caption{\textbf{\ours Visuo-Tactile World Model Rollouts For Successfully Inserting the Peg into the Right Hole}. We show the ground-truth rollouts, predicted rollouts for front view (top), wrist view (middle) and the gelsight Mini tactile image (bottom) as well as the differences between ground-truth and predictions.
    }
    \label{fig:world_model_insert_right_success}
\end{figure}

\begin{figure}
    \centering
    \includegraphics[width=\linewidth]{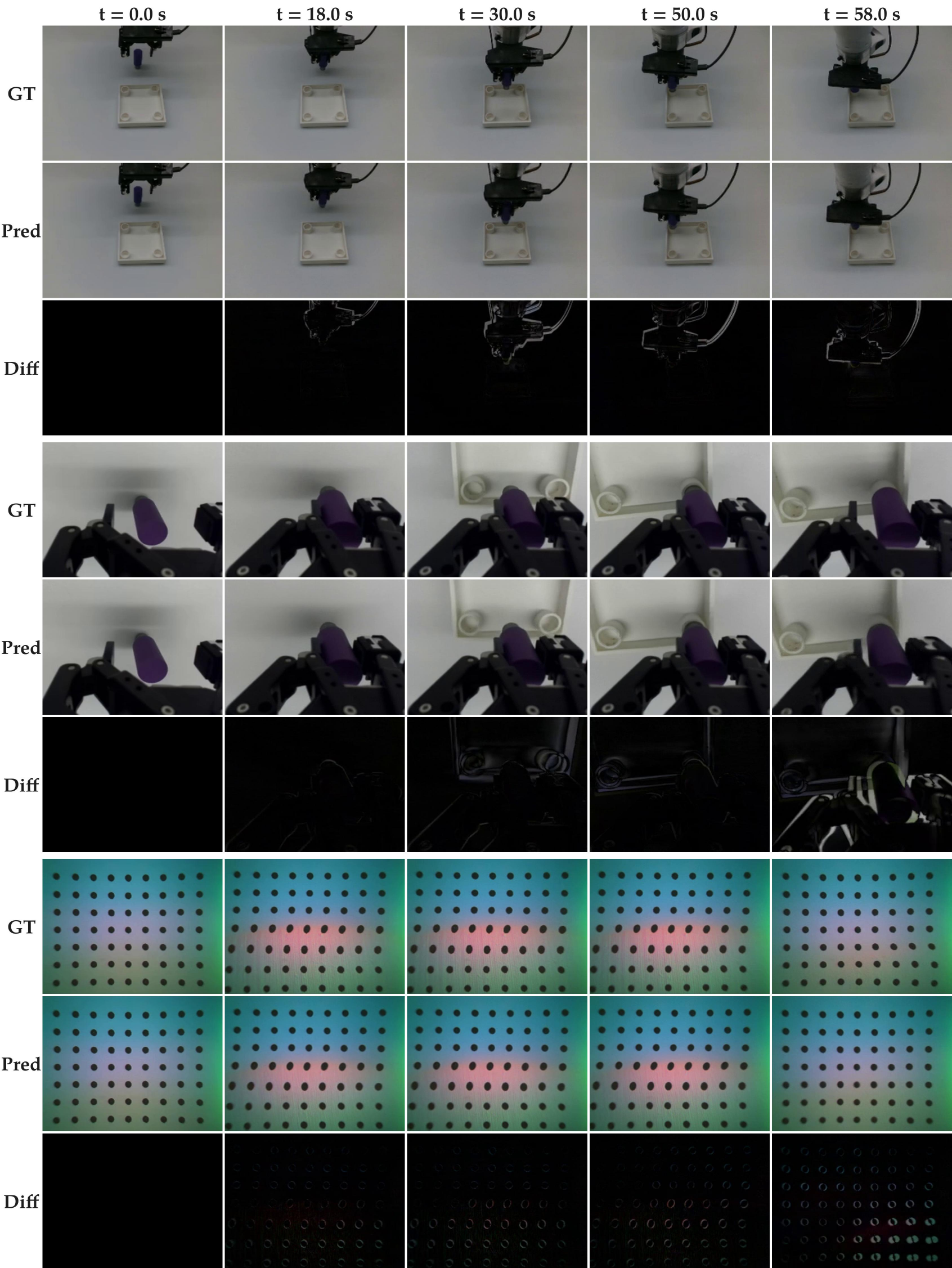}
    \caption{\textbf{\ours Visuo-Tactile World Model Rollouts For Successfully Inserting the Peg into the Left Hole}. We show the ground-truth rollouts, predicted rollouts for front view (top), wrist view (middle) and the gelsight Mini tactile image (bottom) as well as the differences between ground-truth and predictions.
    }
    \label{fig:world_model_insert_left_success}
\end{figure}

\begin{figure}
    \centering
    \includegraphics[width=\linewidth]{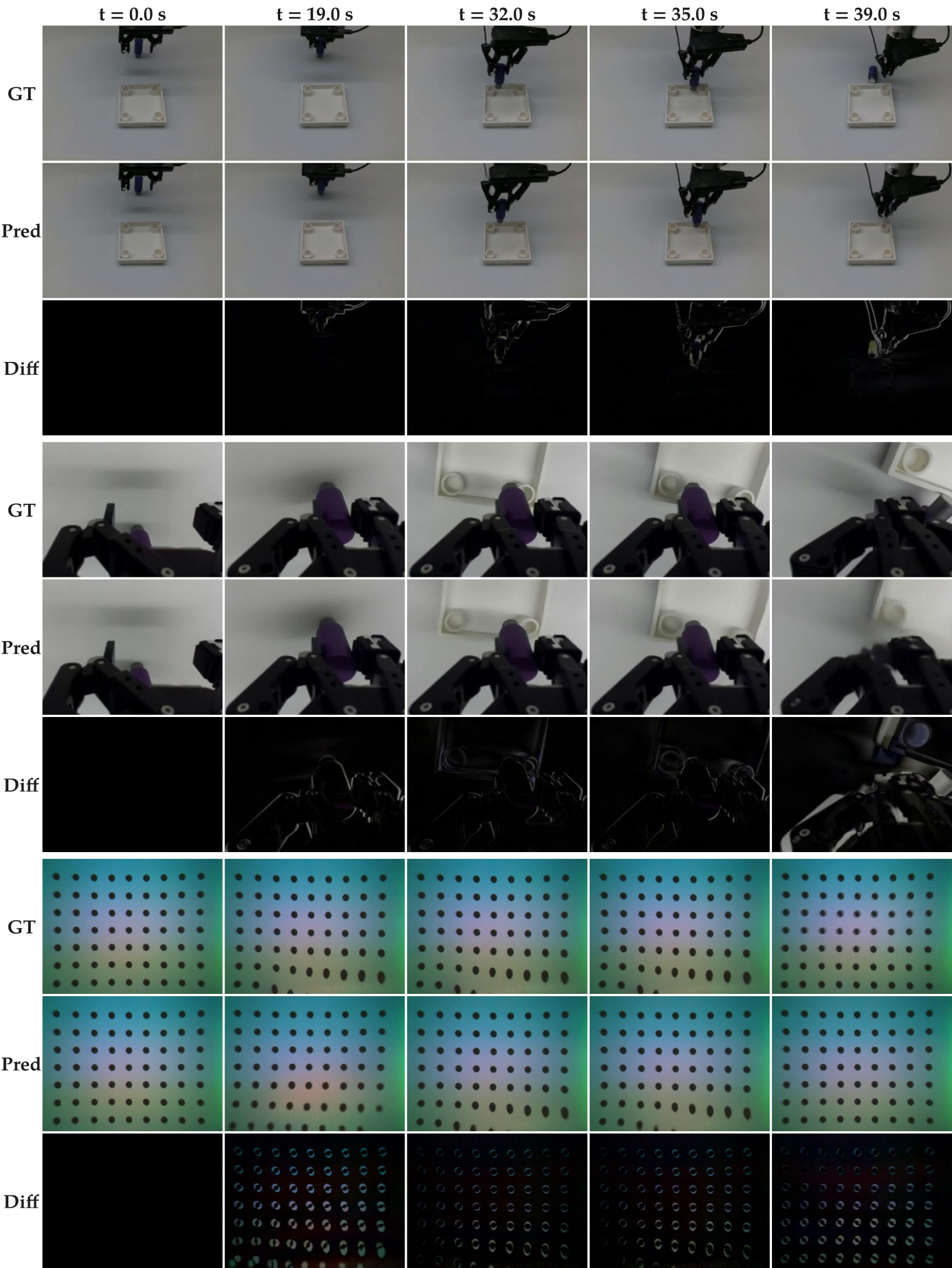}
    \caption{\textbf{\ours Visuo-Tactile World Model Rollouts For Failing to Insert the Peg into the Right Hole}. We show the ground-truth rollouts, predicted rollouts for front view (top), wrist view (middle) and the gelsight Mini tactile image (bottom) as well as the differences between ground-truth and predictions.
    }
    \label{fig:world_model_insert_right_failure}
\end{figure}

\begin{figure}
    \centering
    \includegraphics[width=\linewidth]{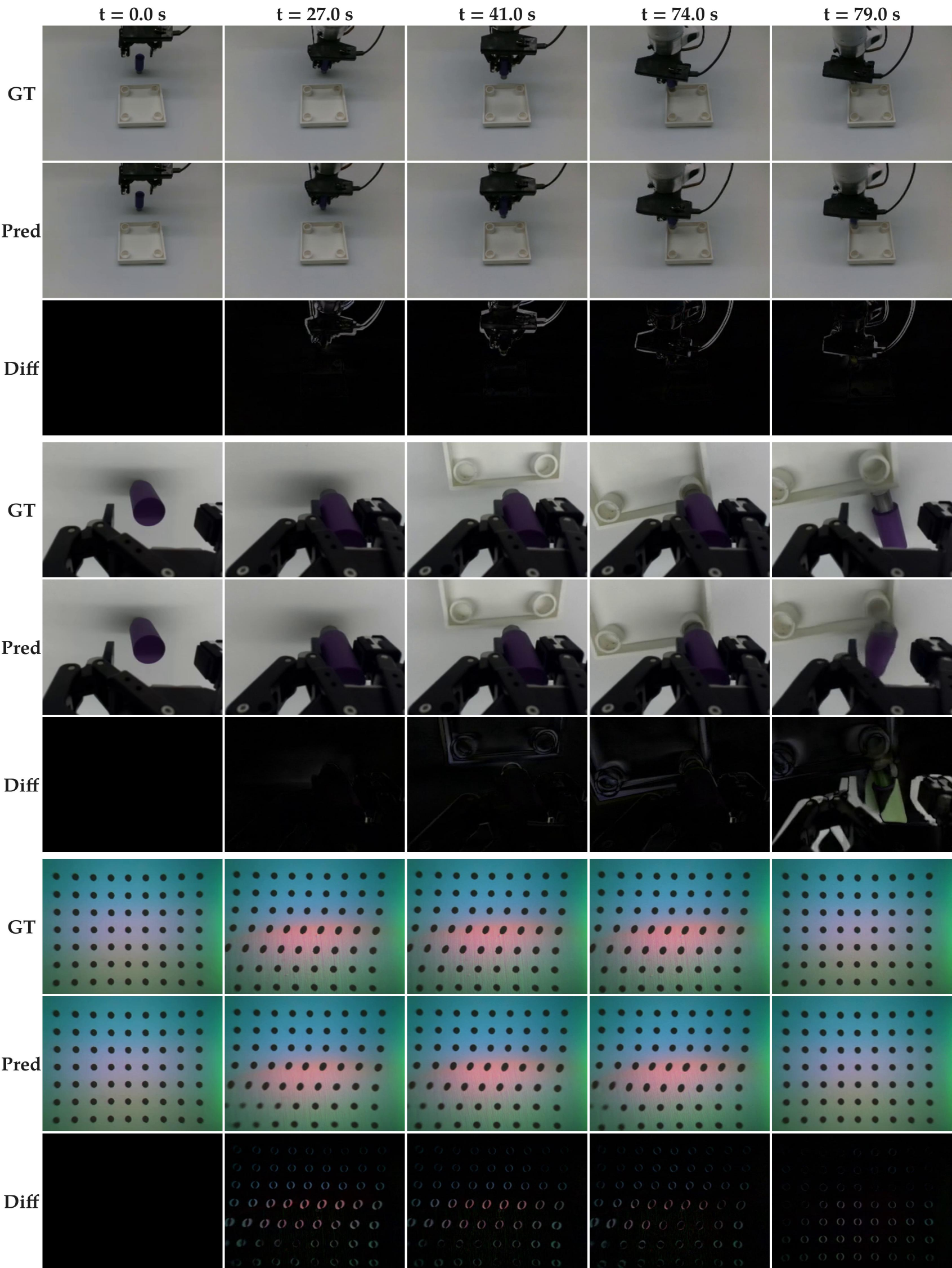}
    \caption{\textbf{\ours Visuo-Tactile World Model Rollouts For Failing to Insert the Peg into the Left Hole}. We show the ground-truth rollouts, predicted rollouts for front view (top), wrist view (middle) and the gelsight Mini tactile image (bottom) as well as the differences between ground-truth and predictions.
    }
    \label{fig:world_model_insert_left_failure}
\end{figure}

\begin{figure}
    \centering
    \includegraphics[width=\linewidth]{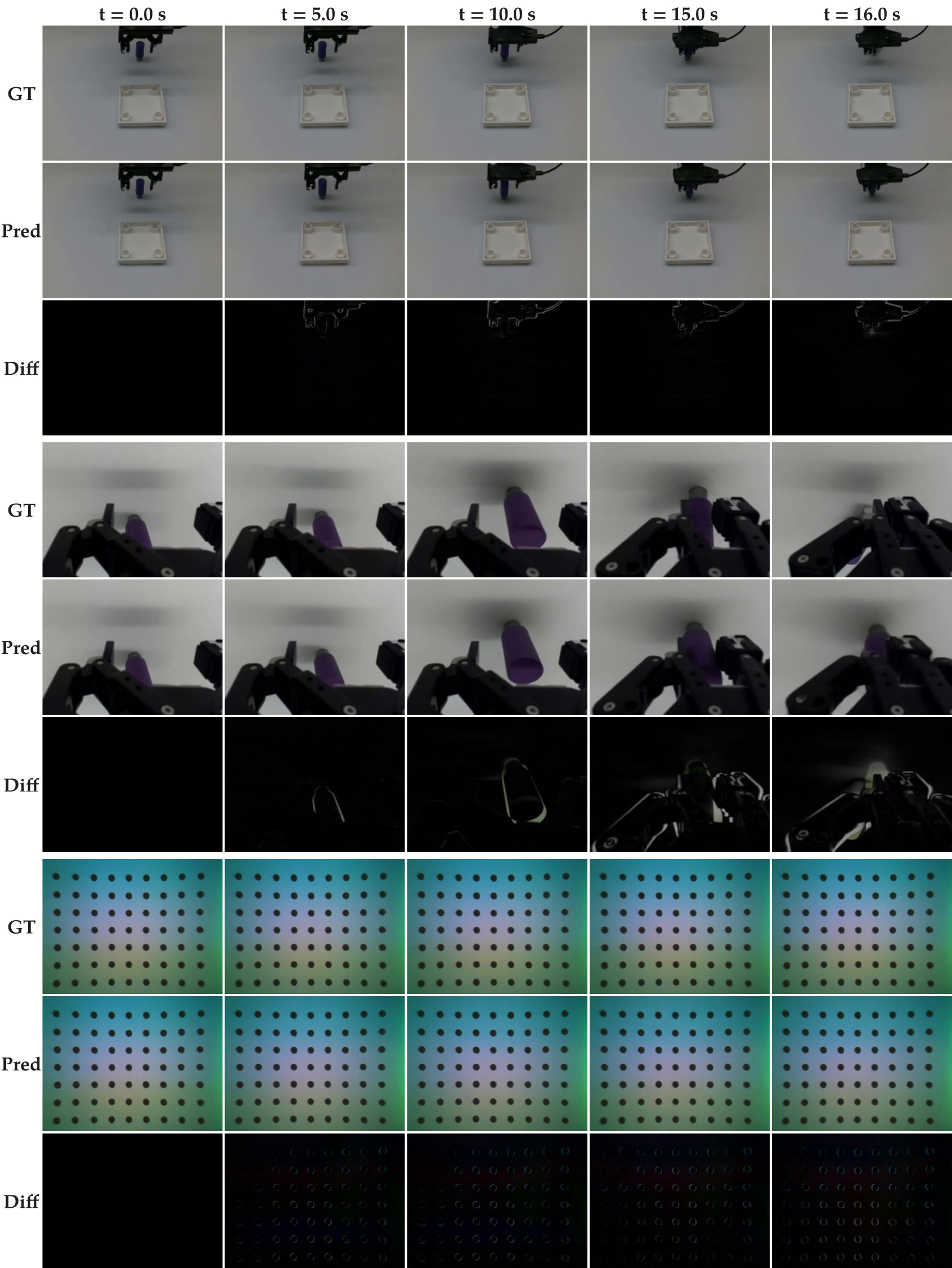}
    \caption{\textbf{\ours Visuo-Tactile World Model Rollouts For Failing to Grasp the Peg}. We show the ground-truth rollouts, predicted rollouts for front view (top), wrist view (middle) and the gelsight Mini tactile image (bottom) as well as the differences between ground-truth and predictions.
    }
    \label{fig:world_model_insert_grasp_failure}
\end{figure}

\begin{figure}
    \centering
    \includegraphics[width=\linewidth]{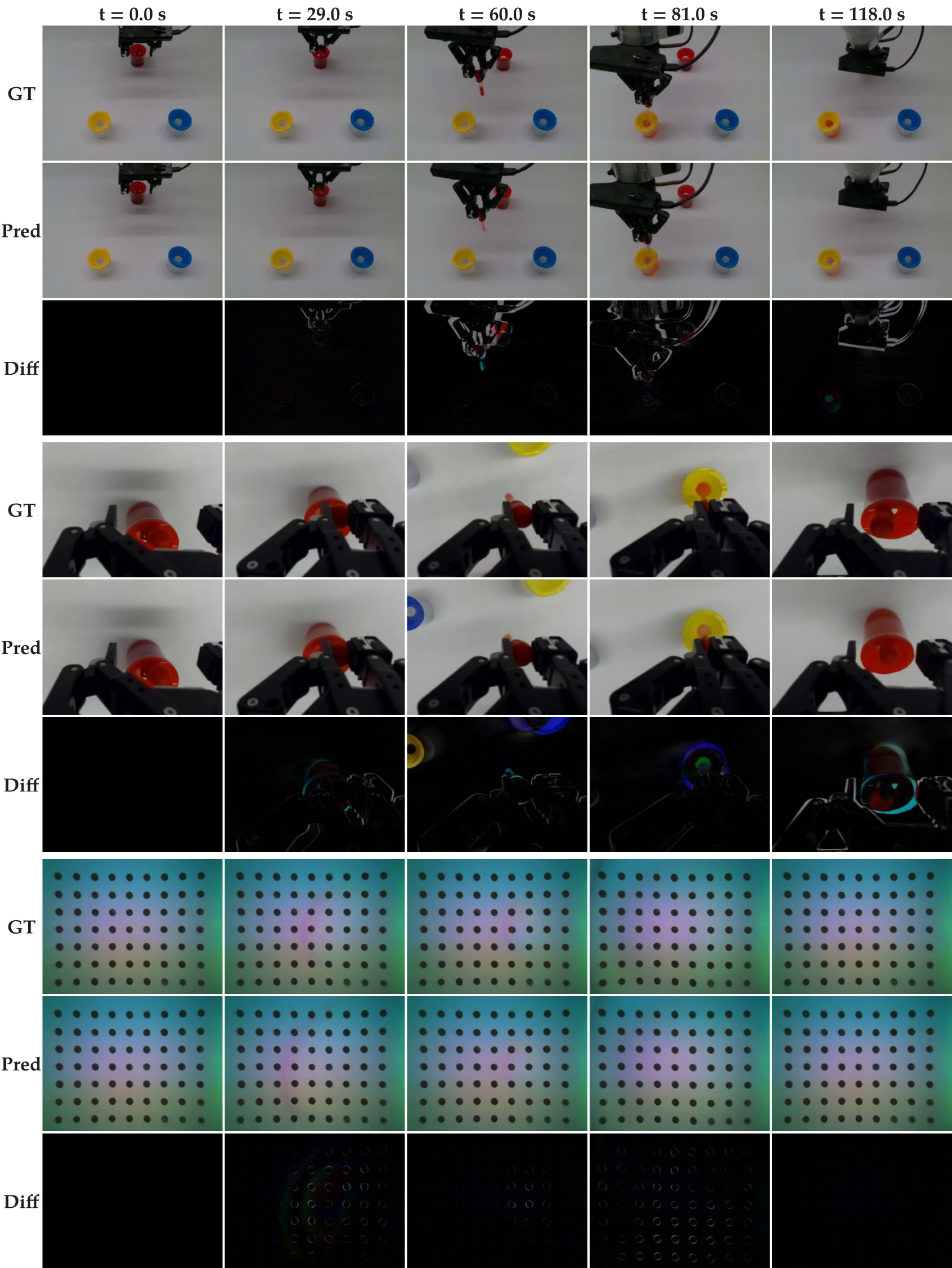}
    \caption{\textbf{\ours Visuo-Tactile World Model Rollouts For Successfully transferring Liquid to the Yellow Cup and Coming Back to the Red Cup}. We show the ground-truth rollouts, predicted rollouts for front view (top), wrist view (middle) and the gelsight Mini tactile image (bottom) as well as the differences between ground-truth and predictions.
    }
\label{fig:world_model_pipette_yellow_success}
\end{figure}

\begin{figure}
    \centering
    \includegraphics[width=\linewidth]{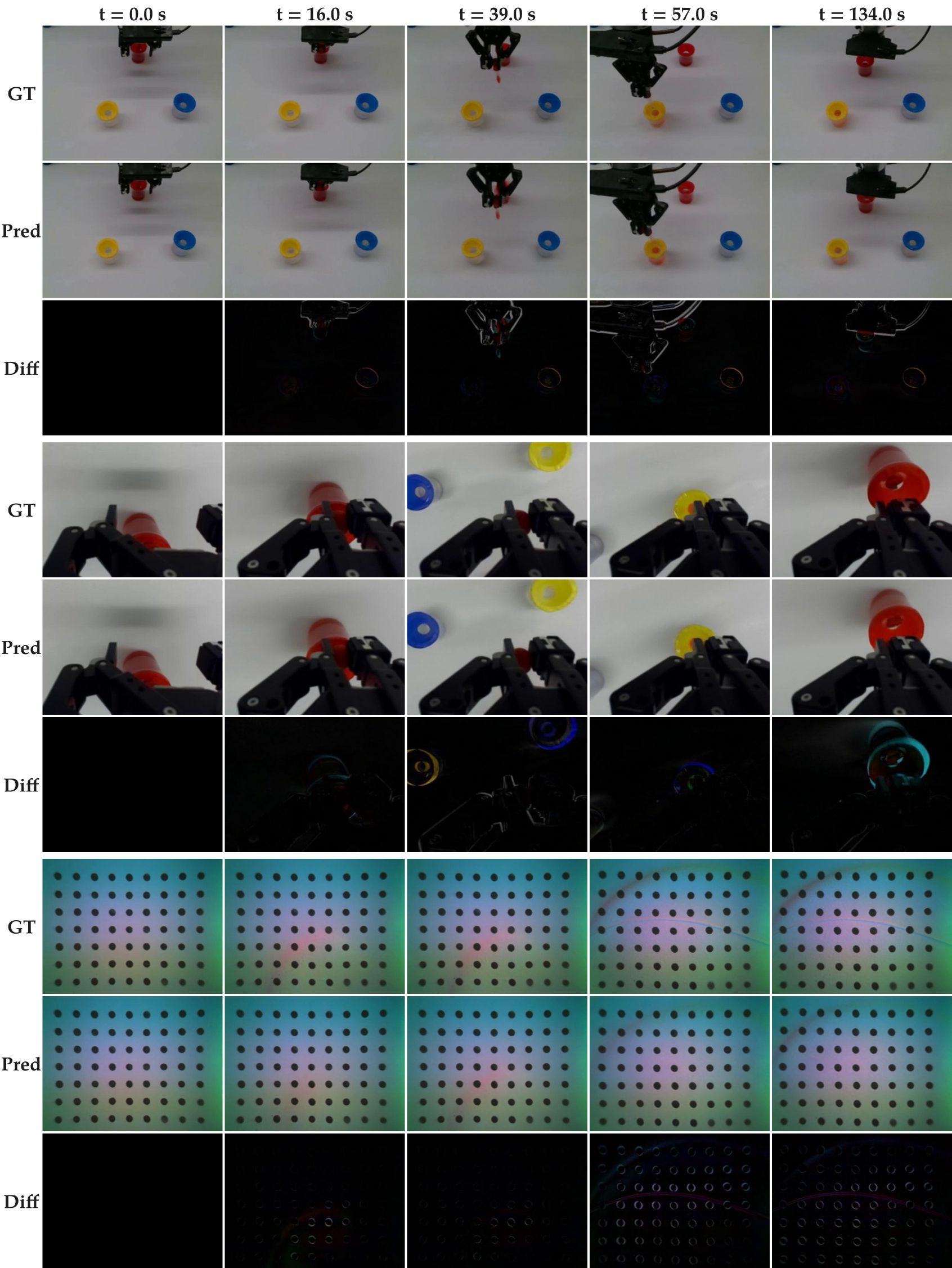}
    \caption{\textbf{\ours Visuo-Tactile World Model Rollouts For Failing to Put the Dropper Back to the Red Cup After Squeezing into the Yellow Cup}. We show the ground-truth rollouts, predicted rollouts for front view (top), wrist view (middle) and the gelsight Mini tactile image (bottom) as well as the differences between ground-truth and predictions.
    }
    \label{fig:world_model_pipette_yellow_failure_back}
\end{figure}

\begin{figure}
    \centering
    \includegraphics[width=\linewidth]{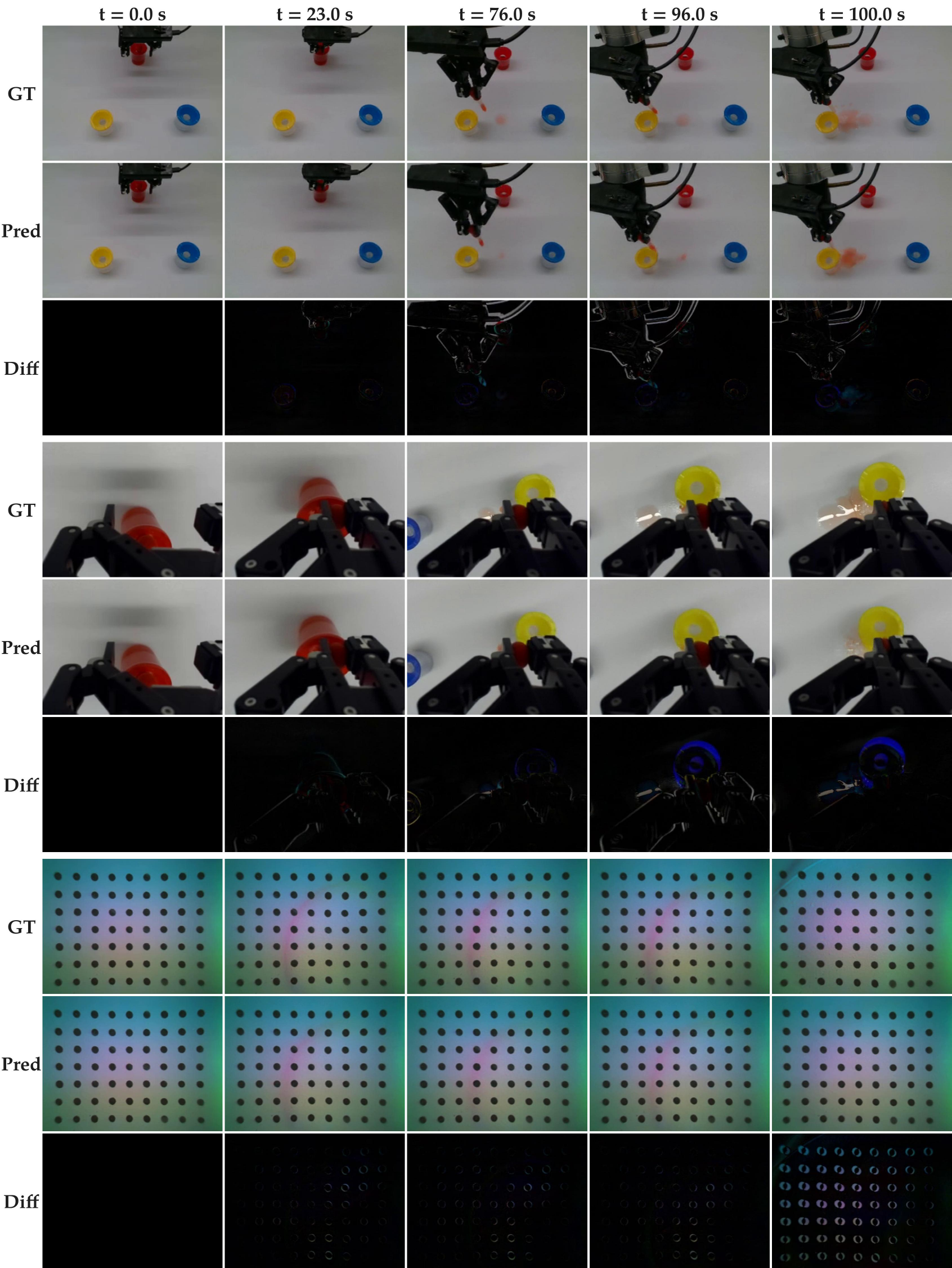}
    \caption{\textbf{\ours Visuo-Tactile World Model Rollouts For Spilling the Liquid Outside the Yellow Cup}. We show the ground-truth rollouts, predicted rollouts for front view (top), wrist view (middle) and the gelsight Mini tactile image (bottom) as well as the differences between ground-truth and predictions.
    }
    \label{fig:world_model_pipette_yellow_failure_spill}
\end{figure}

\begin{figure}
    \centering
    \includegraphics[width=\linewidth]{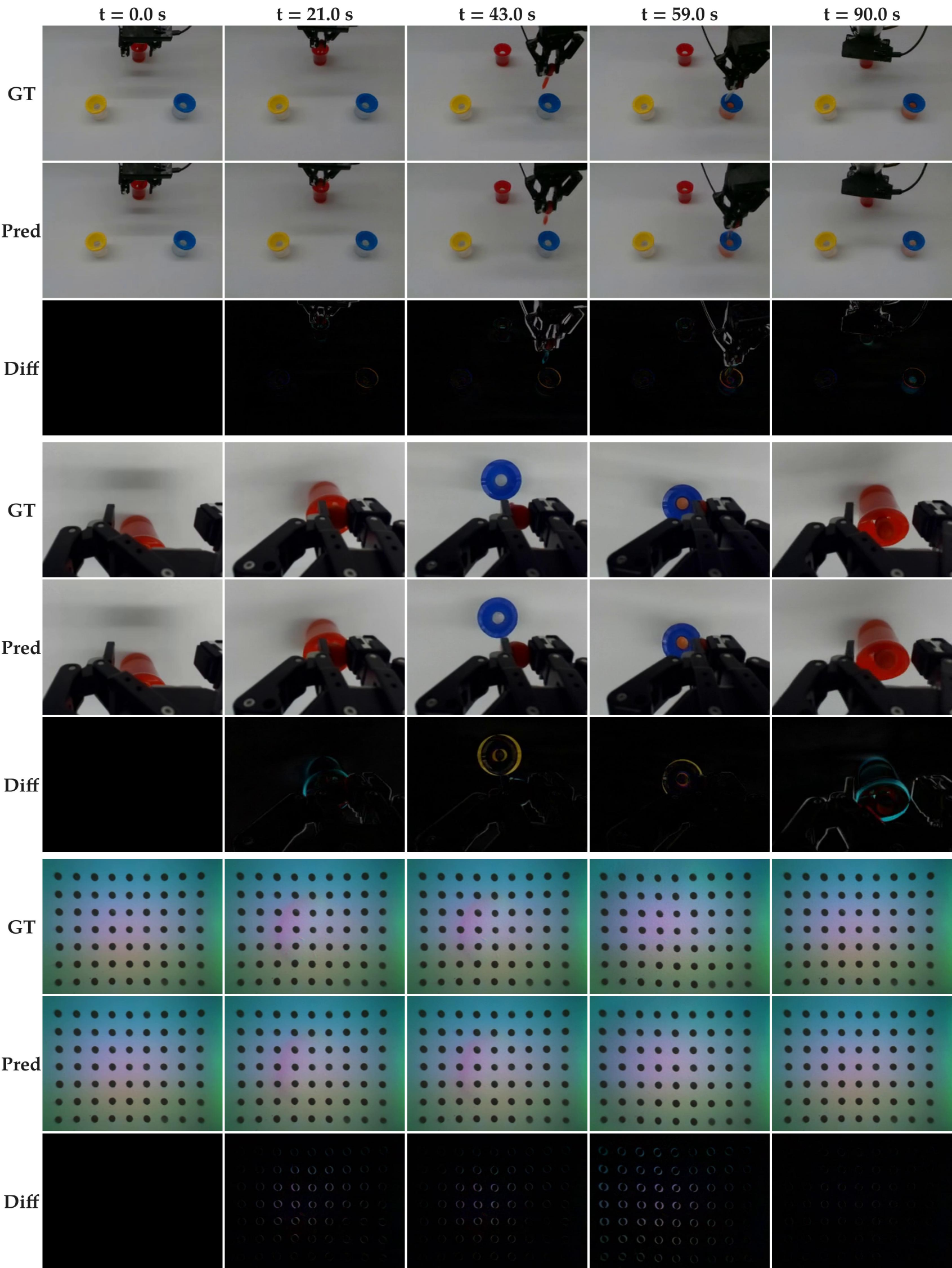}
    \caption{\textbf{\ours Visuo-Tactile World Model Rollouts For Successfully Transferring Liquid to the Blue Cup and Coming Back to the Red Cup}. We show the ground-truth rollouts, predicted rollouts for front view (top), wrist view (middle) and the gelsight Mini tactile image (bottom) as well as the differences between ground-truth and predictions.
    }
\label{fig:world_model_pipette_blue_success}
\end{figure}

\begin{figure}
    \centering
    \includegraphics[width=\linewidth]{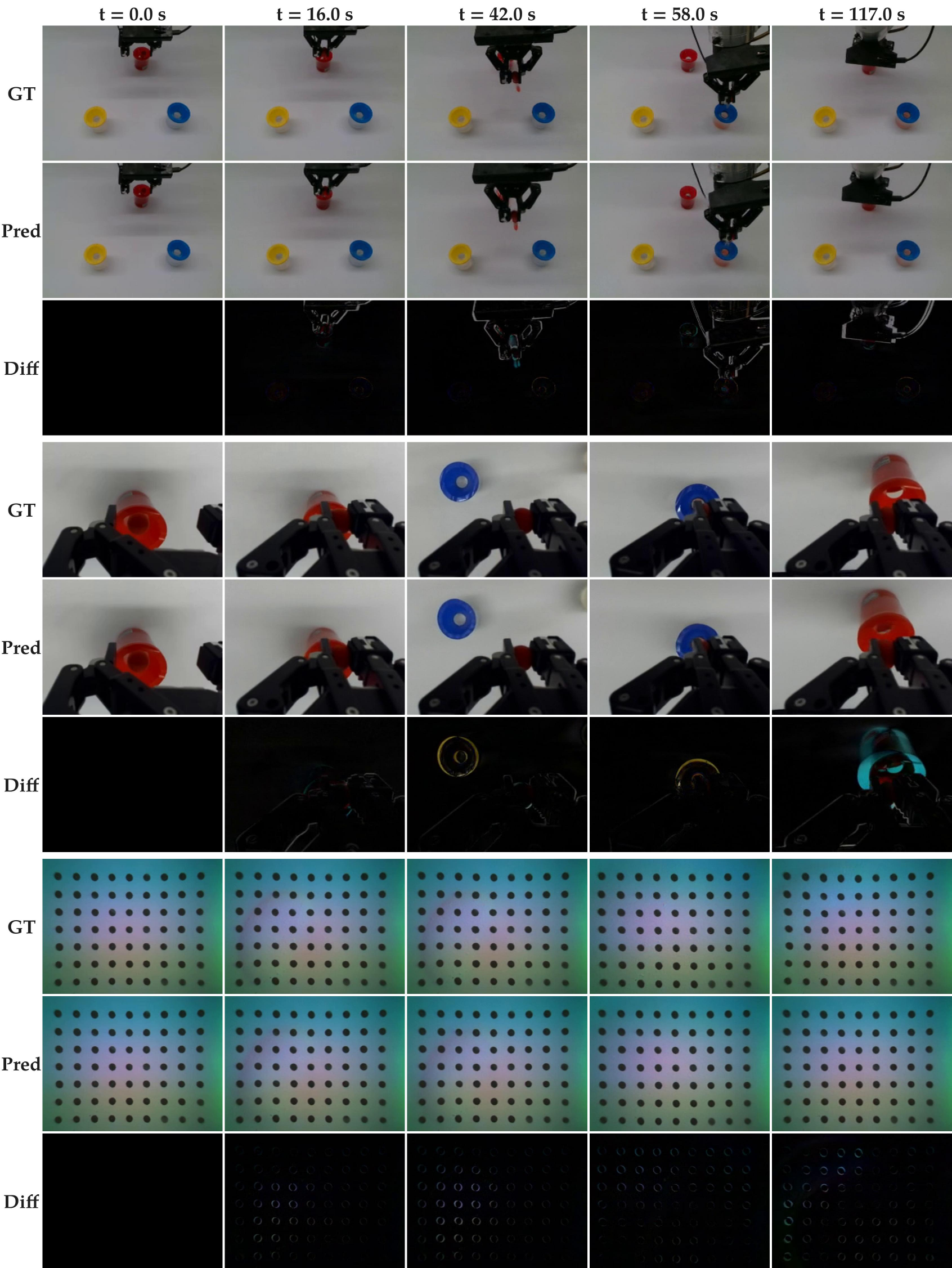}
    \caption{\textbf{\ours Visuo-Tactile World Model Rollouts For Failing to Put the Dropper Back to the Red Cup After Squeezing into the Blue Cup}. We show the ground-truth rollouts, predicted rollouts for front view (top), wrist view (middle) and the gelsight Mini tactile image (bottom) as well as the differences between ground-truth and predictions.
    }
    \label{fig:world_model_pipette_blue_failure_back}
\end{figure}

\begin{figure}
    \centering
    \includegraphics[width=\linewidth]{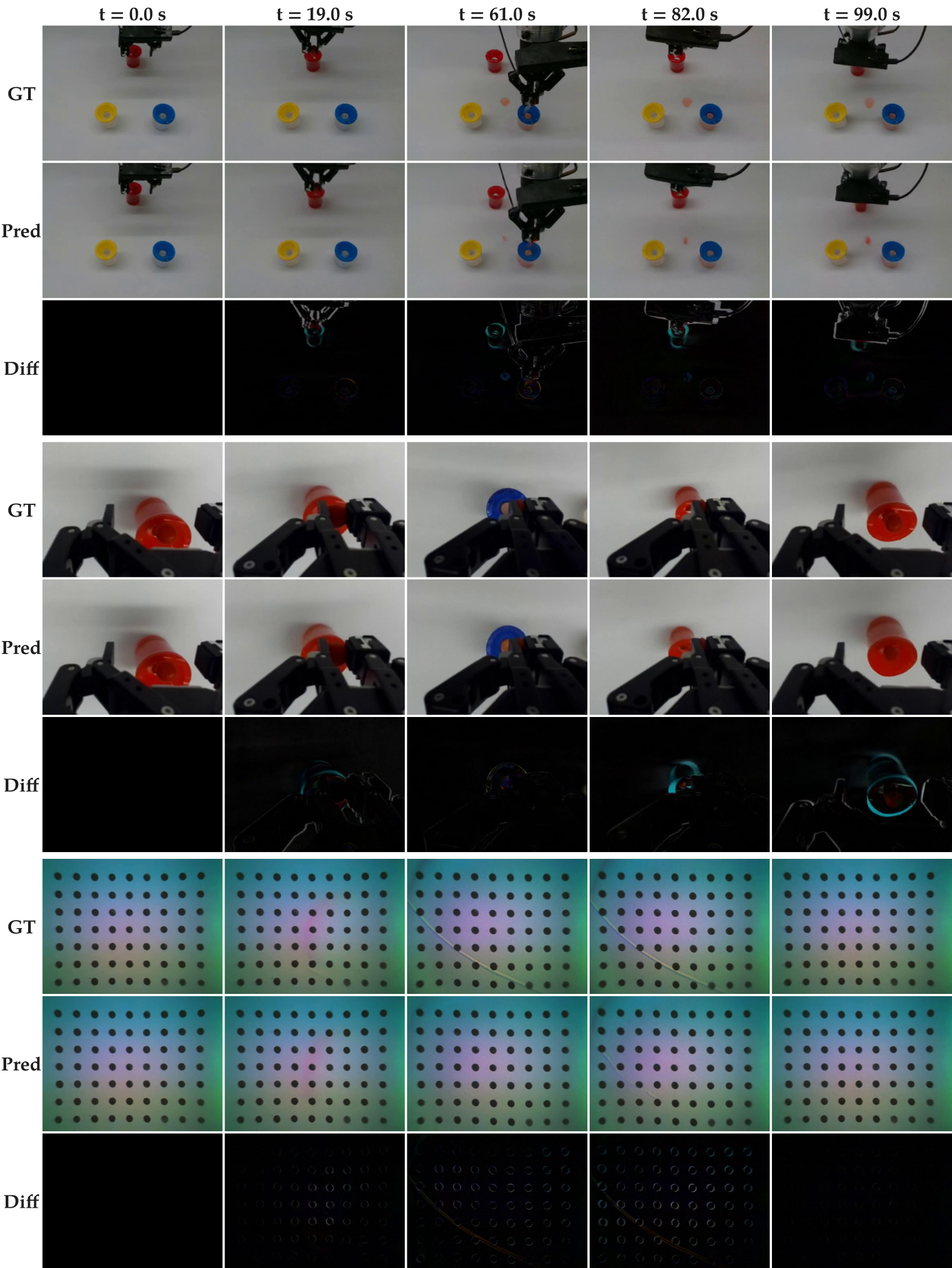}
    \caption{\textbf{\ours Visuo-Tactile World Model Rollouts For Spilling the Liquid Outside the Blue Cup}. We show the ground-truth rollouts, predicted rollouts for front view (top), wrist view (middle) and the gelsight Mini tactile image (bottom) as well as the differences between ground-truth and predictions.
    }
    \label{fig:world_model_pipette_blue_failure_spill}
\end{figure}

\begin{figure}
    \centering
    \includegraphics[width=\linewidth]{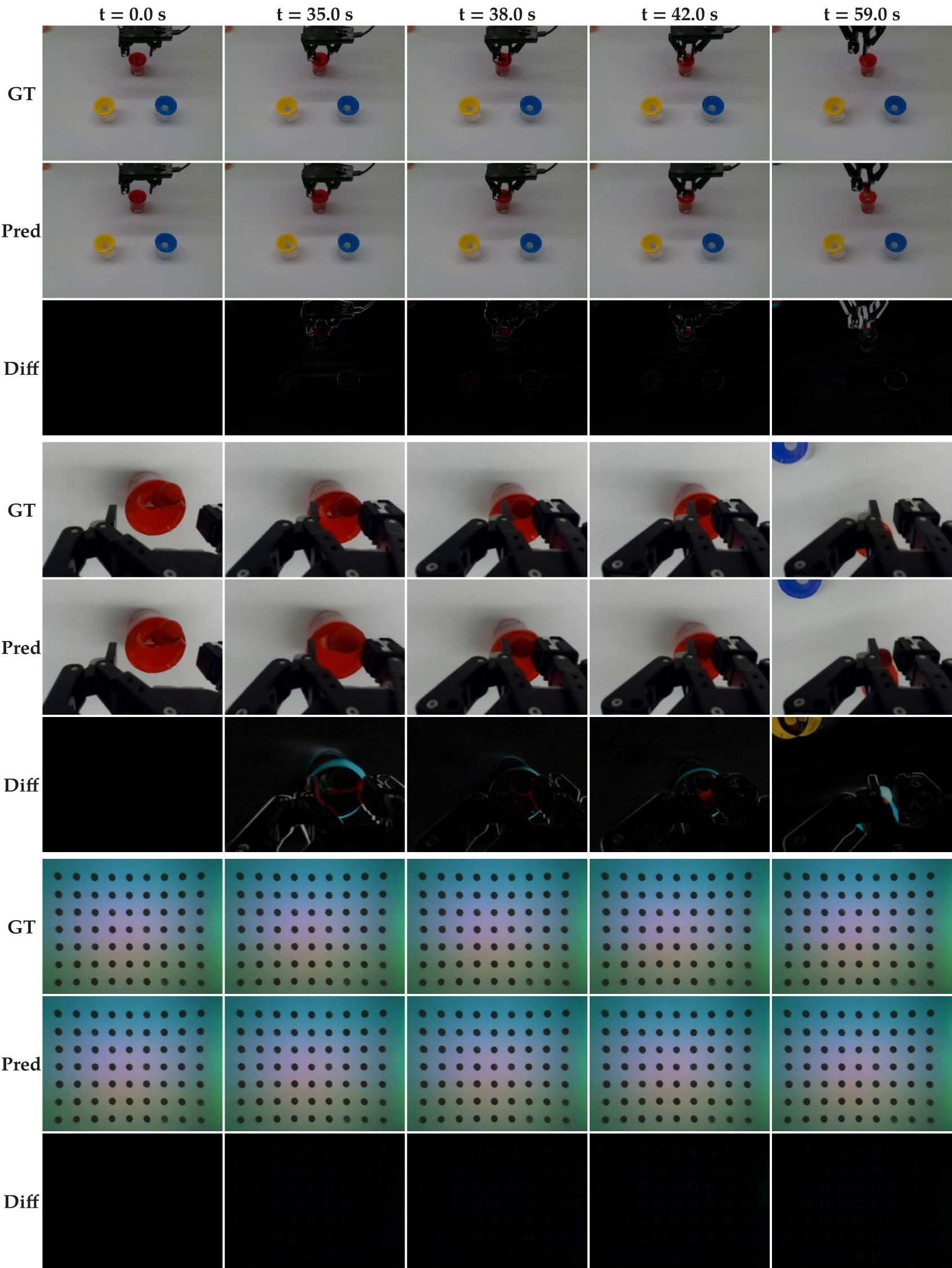}
    \caption{\textbf{\ours Visuo-Tactile World Model Rollouts For Failing to Have a Stable Grasp}. We show the ground-truth rollouts, predicted rollouts for front view (top), wrist view (middle) and the gelsight Mini tactile image (bottom) as well as the differences between ground-truth and predictions.
    }
\label{fig:world_model_pipette_failure_grasp}
\end{figure}

For qualitative evaluation, we further show that the world model can accurately imagine both successful and failed outcomes. For the wiping task, we visualize success and failure cases for wiping the orange marks in Fig.~\ref{fig:world_model_wipe_orange_success} and Fig.~\ref{fig:world_model_wipe_orange_failure}, respectively. We also show success and failure cases for wiping the black marks in Fig.~\ref{fig:world_model_wipe_black_success} and Fig.~\ref{fig:world_model_wipe_black_failure}, as well as a failure mode where the robot begins wiping but becomes stuck midway in Fig.~\ref{fig:world_model_wipe_failure_stuck}. For the insertion task, we visualize the success and failure cases for inserting in the right hole in Fig.~\ref{fig:world_model_insert_right_success} and Fig.~\ref{fig:world_model_insert_right_failure} respectively. We also show success and failure cases for inserting into the left hole in Fig.~\ref{fig:world_model_insert_left_success} and Fig.~\ref{fig:world_model_insert_left_failure}, as well as a failure mode where the robot fails to grasp the purple peg in Fig.~\ref{fig:world_model_insert_grasp_failure}. For pipette transfer task, we visualize the success and failure cases (spilling on the table or failing to put the dropper back) for transferring the liquid to the yellow cup in Fig.~\ref{fig:world_model_pipette_yellow_success}, Fig.~\ref{fig:world_model_pipette_yellow_failure_spill} and Fig.~\ref{fig:world_model_pipette_yellow_failure_back} respectively. We also show the success and failures (spilling on the table or failing to put the dropper back) for transferring the liquid to the blue cup in Fig.~\ref{fig:world_model_pipette_blue_success}, Fig.~\ref{fig:world_model_pipette_blue_failure_spill} and Fig.~\ref{fig:world_model_pipette_blue_failure_back}. We also introduce one additional failure mode of failing to grasp the dropper in Fig.~\ref{fig:world_model_pipette_failure_grasp}.

Overall, these results support the design choice of using a shared visuo-tactile world model for both visual and tactile steering. The model provides reliable long-horizon visual predictions for selecting global action modes, while also producing accurate short-horizon tactile predictions for contact refinement. This is consistent with our bi-level formulation: vision benefits from tactile context when imagining future task progress, and tactile prediction benefits from visual context when estimating upcoming contact interactions.

\begin{table}[t]
\centering
\scriptsize
\setlength{\tabcolsep}{2.5pt}
\renewcommand{\arraystretch}{1.08}
\caption{
\textbf{World Model Prediction Evaluation For Wiping Task}. We evaluate each method across 30 evaluation videos and best values are in bold. RGB metrics are lower-is-better except PSNR and SSIM; flow and height metrics are lower-is-better, while gradient cosine similarity is higher-is-better.
}
\label{tab:world_model_eval_wipe}
\resizebox{\linewidth}{!}{
\begin{tabular}{lllcccccc}
\toprule
\multicolumn{9}{c}{\textbf{RGB Prediction}} \\
\midrule
\textbf{Camera} 
& \textbf{Method} 
& \textbf{Loss Type}
& \textbf{FID}$\downarrow$ 
& \textbf{LPIPS}$\downarrow$ 
& \textbf{MAE}$\downarrow$ 
& \textbf{MSE}$\downarrow$ 
& \textbf{PSNR}$\uparrow$ 
& \textbf{SSIM}$\uparrow$ \\
\midrule
\multirow{2}{*}{Front Camera}
& RGB-only WM 
& Multi-step
& 9.0932 & 0.0995 & 0.0162 & \textbf{0.0023} & 26.9366 & 0.9293 \\
& \ours{} (RGB+Tactile WM)
& Multi-step
& \textbf{8.1015} & \textbf{0.0986} & \textbf{0.0161} & \textbf{0.0023} & \textbf{26.9702} & \textbf{0.9300} \\
\midrule
\multirow{2}{*}{Wrist Camera}
& RGB-only WM 
& Multi-step
& 24.9114 & 0.1741 & 0.0202 & \textbf{0.0029} & \textbf{26.0685} & \textbf{0.9058} \\
& \ours{} (RGB+Tactile WM) 
& Multi-step
& \textbf{22.1616} & \textbf{0.1719} & \textbf{0.0201} & 0.0030 & 25.8676 & 0.9045 \\
\bottomrule
\end{tabular}
}

\vspace{0.6em}

\resizebox{\linewidth}{!}{
\begin{tabular}{llccccc}
\toprule
\multicolumn{7}{c}{\textbf{Tactile Prediction}} \\
\midrule
\textbf{Method} 
& \textbf{Loss Type}
& \textbf{Flow AE}$\downarrow$ 
& \textbf{Flow EPE}$\downarrow$ 
& \textbf{Grad. Cos.}$\uparrow$ 
& \textbf{Height MAE}$\downarrow$ 
& \textbf{Height RMSE}$\downarrow$ \\
\midrule
Tactile-only WM 
& Multi-step
& 6.6576
& 0.1396
& \textbf{0.5907}
& 0.2380
& 0.2965 \\

\ours{} (RGB+Tactile WM)
& Multi-step
& \textbf{6.4938} 
& \textbf{0.1364} 
& 0.5829
& \textbf{0.2352} 
& \textbf{0.2959} \\
\bottomrule
\end{tabular}
}
\end{table}

\begin{table}[ht]
\centering
\scriptsize
\setlength{\tabcolsep}{2.5pt}
\renewcommand{\arraystretch}{1.08}
\caption{
\textbf{World Model Prediction Evaluation For Insertion Task}. We evaluate each method across 30 evaluation videos and best values are in bold. RGB metrics are lower-is-better except PSNR and SSIM; flow and height metrics are lower-is-better, while gradient cosine similarity is higher-is-better.
}
\label{tab:world_model_eval_insert}

\resizebox{\linewidth}{!}{
\begin{tabular}{lllcccccc}
\toprule
\multicolumn{9}{c}{\textbf{RGB Prediction}} \\
\midrule
\textbf{Camera} 
& \textbf{Method} 
& \textbf{Loss Type}
& \textbf{FID}$\downarrow$ 
& \textbf{LPIPS}$\downarrow$ 
& \textbf{MAE}$\downarrow$ 
& \textbf{MSE}$\downarrow$ 
& \textbf{PSNR}$\uparrow$ 
& \textbf{SSIM}$\uparrow$ \\
\midrule
\multirow{3}{*}{Front Camera}
& RGB-only WM 
& Multi-step
& 8.1047 & 0.1576 & 0.0122 & \textbf{0.0012} & 29.5526 & 0.9367 \\

& \ours{} (RGB+Tactile WM)
& Multi-step
& \textbf{8.0094} & \textbf{0.1566} & \textbf{0.0121} & \textbf{0.0012} & \textbf{29.6457} & \textbf{0.9372} \\
\midrule
\multirow{3}{*}{Wrist Camera}
& RGB-only WM 
& Multi-step
& 9.3695 & 0.1395 & 0.0195 & 0.0023 & 26.8440 & 0.9129 \\

& \ours{} (RGB+Tactile WM)
& Multi-step
& \textbf{8.9580} & \textbf{0.1382} & \textbf{0.0191} & \textbf{0.0022} & \textbf{27.1264} & \textbf{0.9156} \\
\bottomrule
\end{tabular}
}

\vspace{0.6em}

\resizebox{\linewidth}{!}{
\begin{tabular}{llccccc}
\toprule
\multicolumn{7}{c}{\textbf{Tactile Prediction}} \\
\midrule
\textbf{Method} 
& \textbf{Loss Type}
& \textbf{Flow AE}$\downarrow$ 
& \textbf{Flow EPE}$\downarrow$ 
& \textbf{Grad. Cos.}$\uparrow$ 
& \textbf{Height MAE}$\downarrow$ 
& \textbf{Height RMSE}$\downarrow$ \\
\midrule
Tactile-only WM 
& Multi-step
& 8.8939 & 0.2209 & 0.4977 & 0.6896 & 0.8166 \\

\ours{} (RGB+Tactile WM)
& Multi-step
& \textbf{7.5189} & \textbf{0.1742} & \textbf{0.5014} & \textbf{0.6507} & \textbf{0.7656} \\
\bottomrule
\end{tabular}
}
\end{table}

\begin{table}[!ht]
\centering
\scriptsize
\setlength{\tabcolsep}{2.5pt}
\renewcommand{\arraystretch}{1.08}
\caption{
\textbf{World Model Prediction Evaluation For Pipette Transfer Task}. We evaluate each method across a fixed evaluation set of 30 videos and best values are in bold. RGB metrics are lower-is-better except PSNR and SSIM; flow and height metrics are lower-is-better, while gradient cosine similarity is higher-is-better.
}
\label{tab:world_model_eval_pipette}
\resizebox{\linewidth}{!}{
\begin{tabular}{lllcccccc}
\toprule
\multicolumn{9}{c}{\textbf{RGB Prediction}} \\
\midrule
\textbf{Camera} 
& \textbf{Method} 
& \textbf{Loss Type}
& \textbf{FID}$\downarrow$ 
& \textbf{LPIPS}$\downarrow$ 
& \textbf{MAE}$\downarrow$ 
& \textbf{MSE}$\downarrow$ 
& \textbf{PSNR}$\uparrow$ 
& \textbf{SSIM}$\uparrow$ \\
\midrule
\multirow{2}{*}{Front Camera}
& RGB-only WM 
& Multi-step
& 12.4612 & 0.1700 & \textbf{0.0162} & \textbf{0.0017} & \textbf{27.8010} & \textbf{0.9245} \\
& \ours{} (RGB+Tactile WM) 
& Multi-step
& \textbf{11.5609} & \textbf{0.1699} & 0.0163 & 0.0019 & 27.5515 & 0.9222 \\
\midrule
\multirow{2}{*}{Wrist Camera}
& RGB-only WM 
& Multi-step
& 11.3101 & 0.1481 & \textbf{0.0173} & \textbf{0.0017} & \textbf{27.9782} & \textbf{0.9273} \\
& \ours{} (RGB+Tactile WM)
& Multi-step
& \textbf{10.5830} & \textbf{0.1480} & \textbf{0.0173} & 0.0018 & 27.7871 & 0.9266 \\
\bottomrule
\end{tabular}
}

\vspace{0.6em}

\resizebox{\linewidth}{!}{
\begin{tabular}{llccccc}
\toprule
\multicolumn{7}{c}{\textbf{Tactile Prediction}} \\
\midrule
\textbf{Method} 
& \textbf{Loss Type}
& \textbf{Flow AE}$\downarrow$ 
& \textbf{Flow EPE}$\downarrow$ 
& \textbf{Grad. Cos.}$\uparrow$ 
& \textbf{Height MAE}$\downarrow$ 
& \textbf{Height RMSE}$\downarrow$ \\
\midrule
Tactile-only WM 
& Multi-step
& 2.2372
& 0.0451
& 0.5391
& 0.6648
& 0.7642 \\

\ours{} (RGB+Tactile WM) 
& Multi-step
& \textbf{1.9514} 
& \textbf{0.0381} 
& \textbf{0.5415}
& \textbf{0.6061}
& \textbf{0.7082} \\
\bottomrule
\end{tabular}
}
\end{table}

\subsection{Multi-Modal Guidance for Visuo-Tactile Policies}
\label{app:visuo_tactile_policy}
We evaluate whether our multi-modal guidance framework can improve both visual policies and visuo-tactile policies. For each task, we compare four variants: a base visual policy, a base visuo-tactile policy, our method applied to the visual policy, and our method applied to the visuo-tactile policy.

\para{Base Policy}
 The visual policy receives wrist and external RGB observations together with proprioceptive state, while the visuo-tactile policy additionally receives the GelSight tactile observation. Both policies use the same action prediction horizon and execution protocol described in Appendix~\ref{app:wipe_task}. 

Our method is applied at inference time without changing the underlying policy weights. For each policy query, visual steering first samples candidate action sequences and selects the one with the highest predicted visual reward. Tactile steering then locally refines the selected action chunk using tactile-guided diffusion editing. This allows us to test whether guidance improves policies that do not directly observe tactile input, as well as policies that already use tactile observations during action generation.

\begin{figure}
    \centering
    \includegraphics[width=\linewidth]{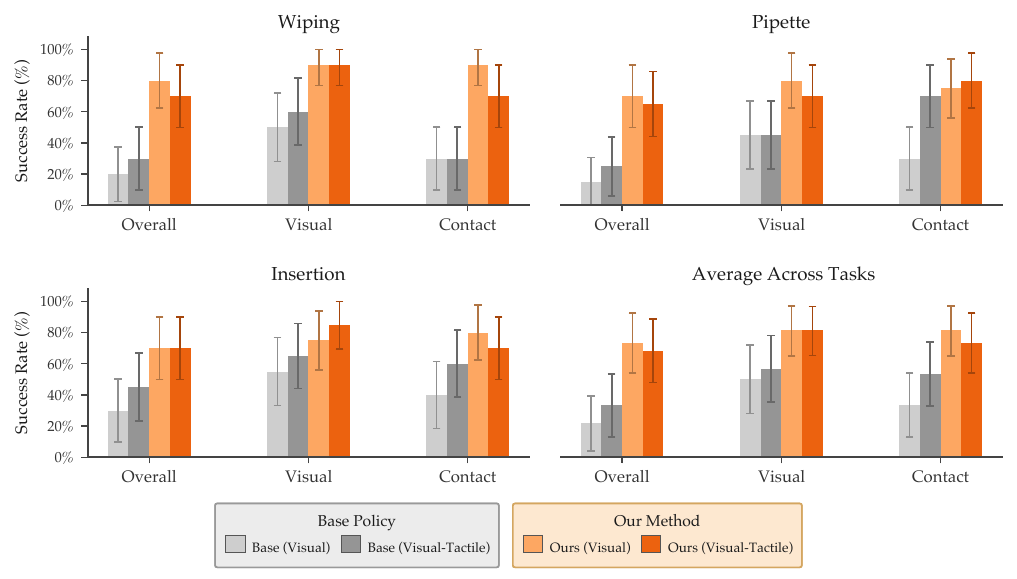}
    \caption{\textbf{Policy Steering Performance with Different Base Policies}. We ablate different base policies including policy conditioned on visual observations and policy conditioned on visuo-tactile observations. Our policy steering method \ours obtains a substantial performance gain no matter what the base policy is.}
    \label{fig:vital_visuo_tactile_policy}
\end{figure}

\para{Results}
Fig.~\ref{fig:vital_visuo_tactile_policy} reports visual success, contact success, and overall task success across the wiping, pipette transfer, and insertion tasks. Across all tasks, our method substantially improves both the visual and visuo-tactile base policies. For the visual policy, our method improves overall success from $30\%$ to $90\%$ on wiping, from $30\%$ to $75\%$ on pipette transfer, and from $40\%$ to $80\%$ on insertion. These gains indicate that multi-modal guidance can compensate for limitations of the base policy by using predicted visual futures for global goal selection and predicted tactile futures for contact refinement.

The base visuo-tactile policy generally improves contact success over the base visual policy, especially for pipette transfer and insertion, where tactile feedback is important for grasp stability and precise contact. However, adding tactile observations alone does not fully solve the tasks: overall success remains $30\%$, $70\%$, and $60\%$ for wiping, pipette transfer, and insertion, respectively. With our guidance, the visuo-tactile policy improves to $70\%$, $80\%$, and $70\%$ overall success. This shows that explicit inference-time steering remains beneficial even when the policy already receives tactile observations.

Comparing our method on visual versus visuo-tactile policies, we observe that the stronger base policy does not always lead to higher final success after guidance. For wiping and insertion, our method with the visual policy achieves the highest overall success, while for pipette transfer, our method with the visuo-tactile policy performs best. This suggests that the benefit of tactile observations in the base policy depends on the task and on how well the policy uses tactile signals during action generation. Sometimes adding the tactile observations in the policy input leads to the overfitting of the policy and the base policy distribution is much narrower compared to the visual policy, resulting in worse steering performance. In summary, our guidance framework consistently improves performance by explicitly decomposing the problem into visual mode selection and tactile contact refinement.

\section{Additional Results}
\subsection{Multi-modal Visuo-Tactile Steering Outperforms Steering with Single Modality}
\label{app:results_multi_modal}

In this section, we provide additional qualitative results for multi-modal visuo-tactile steering across the wiping, insertion, and pipette transfer tasks. These examples complement the quantitative results in Sec.~\ref{sec:exp_multimodal_guidance} and illustrate how visual and tactile guidance contribute to different aspects of task success.

\begin{figure}
    \centering
    \includegraphics[width=\linewidth]{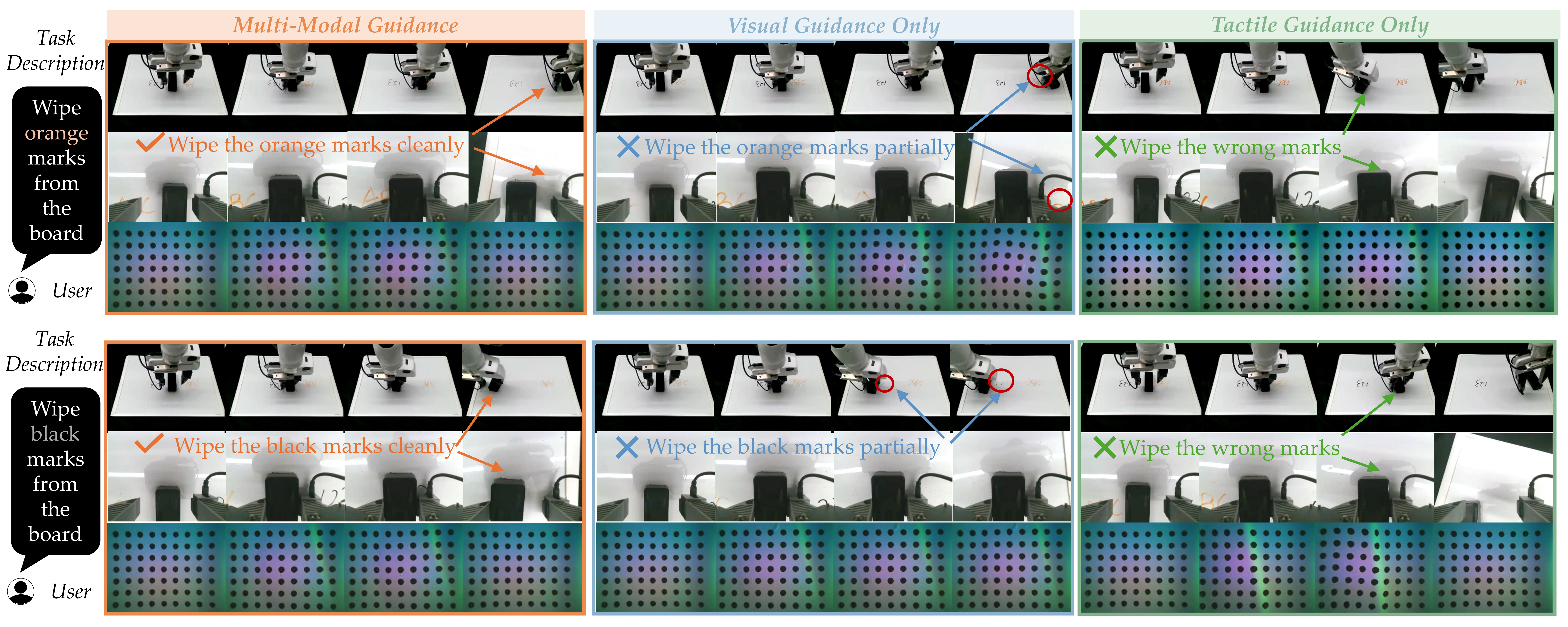}
    \caption{\textbf{Policy Steering with Different Guidance Modalities For Wiping Task}. For both task descriptions of "wiping the orange marks" and "wiping the black marks", vision-only guidance (middle) fail due to local contact errors, while tactile-only guidance (right) can fail due to global task misalignment. \ours (left) with multi-modal guidance addresses both failure modes.
    }
    \label{fig:multi_modal_wipe_all}
\end{figure}

\begin{figure}
    \centering
    \includegraphics[width=\linewidth]{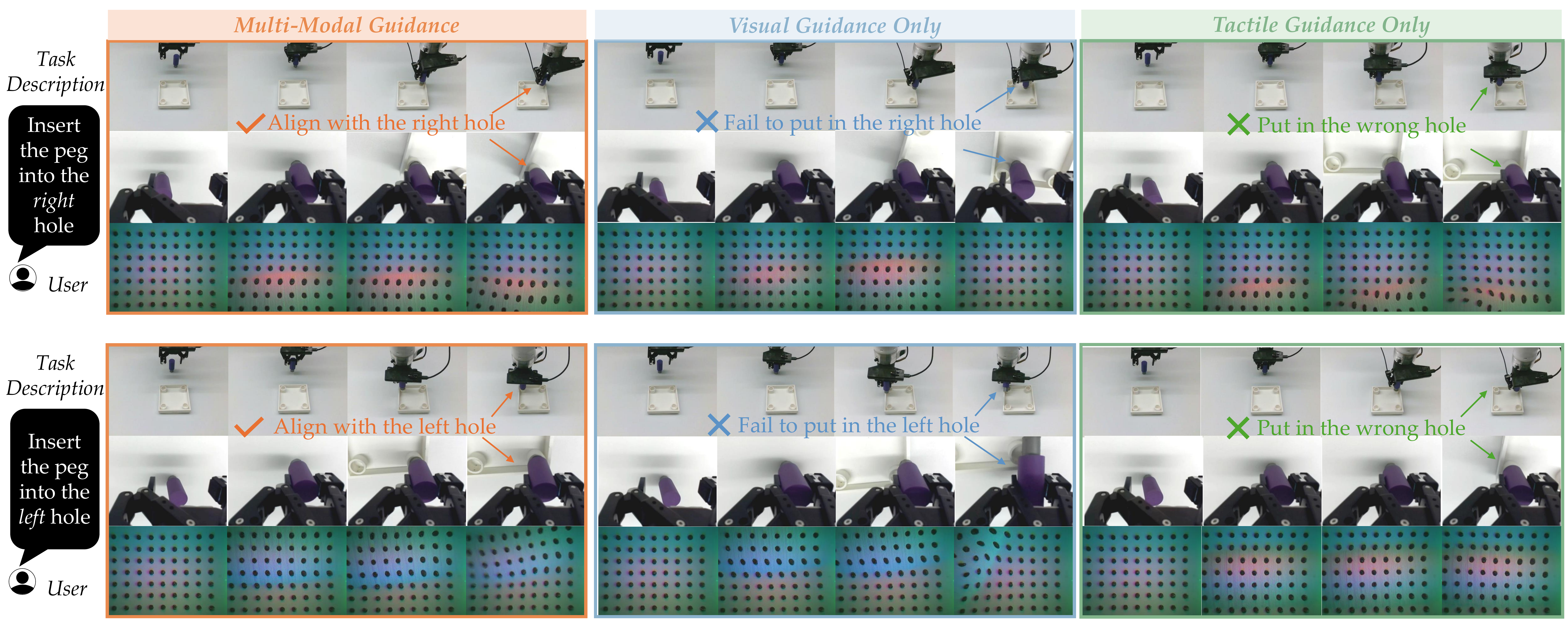}
      \caption{\textbf{Policy Steering with Different Guidance Modalities For Insertion Task}. For both task descriptions of "inserting into the left hole" and "inserting into the right hole", vision-only guidance (middle) fail due to local contact errors, while tactile-only guidance (right) can fail due to global task misalignment. \ours (left) with multi-modal guidance addresses both failure modes.
    }
    \label{fig:multi_modal_insert_all}
\end{figure}

\begin{figure}
    \centering
    \includegraphics[width=\linewidth]{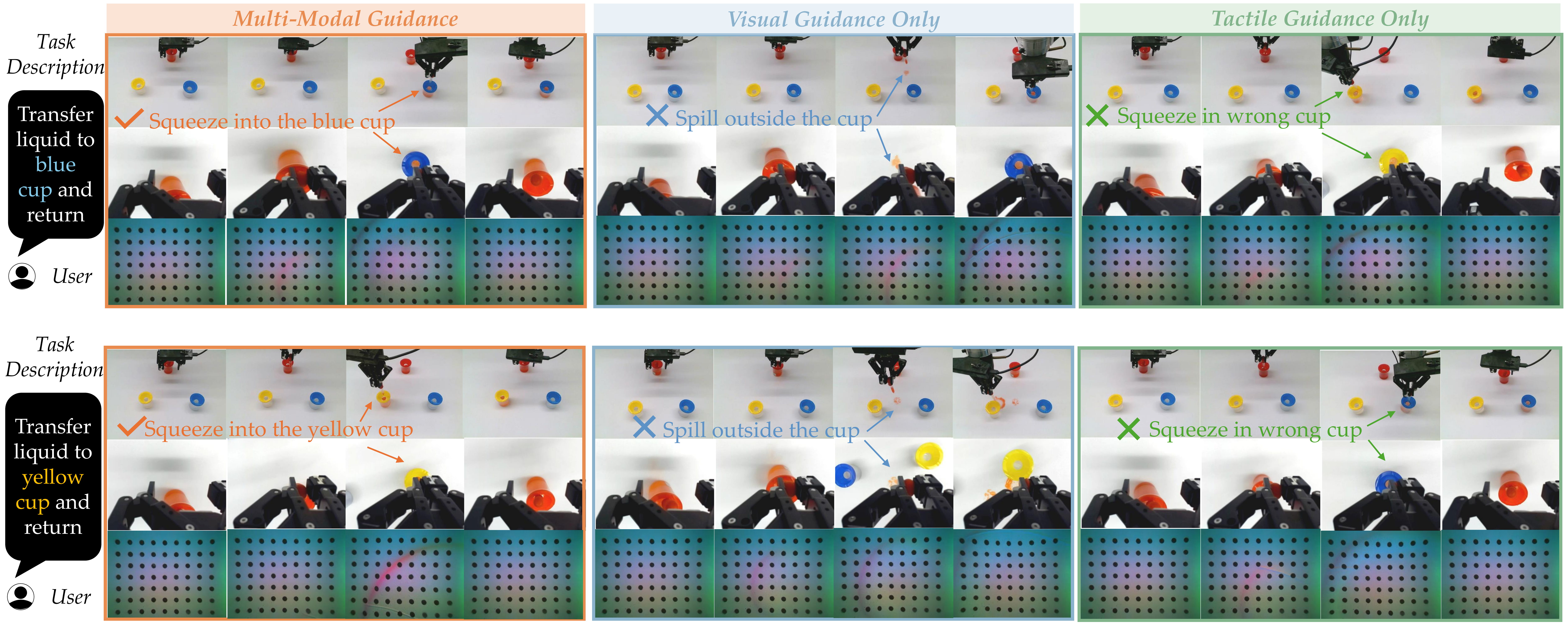}
      \caption{\textbf{Policy Steering with Different Guidance Modalities For Pipette Transfer Task}. For both task descriptions of "transfer liquid to blue cup" and "transfer liquid to right cup", vision-only guidance (middle) fails due to local contact errors, while tactile-only guidance (right) can fail due to global task misalignment. \ours (left) with multi-modal guidance addresses both failure modes.
    }
    \label{fig:multi_modal_pipette_all}
\end{figure}

\para{Success of Multi-Modal Steering}
Across all tasks, \ours{} succeeds by using visual guidance to select the correct global behavior mode and tactile guidance to refine local contact-rich execution. In the wiping task, visual steering selects the correct marked region according to the language instruction, while tactile steering improves contact pressure between the eraser and the board, allowing the robot to remove the target marks more completely as shown in Fig.~\ref{fig:multi_modal_wipe_all}. In the insertion task, visual steering selects the correct target hole, while tactile steering helps adjust the peg motion during contact, improving alignment and insertion under the tight $1$ mm tolerance as shown in Fig.~\ref{fig:multi_modal_insert_all}. In the pipette transfer task, visual steering guides the robot toward the specified blue or yellow cup and then back to the red cup, while tactile steering improves grasp stability and contact force during liquid transfer and return as shown in Fig.~\ref{fig:multi_modal_pipette_all}.

\begin{figure}
    \centering
    \includegraphics[width=0.9\linewidth]{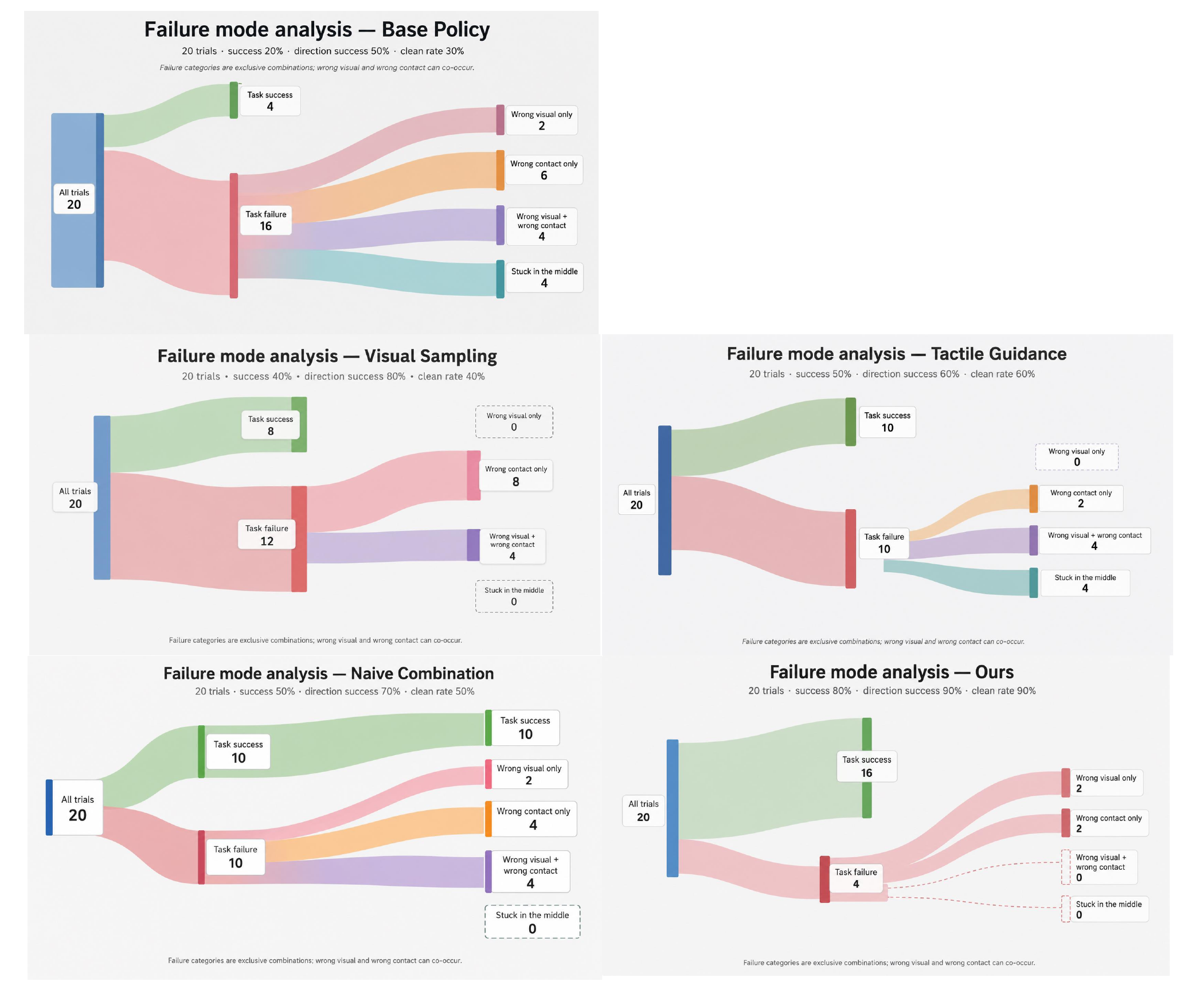}
    \caption{\textbf{Failure Mode Analysis For Wiping Task}. We demonstrate the different failure modes of \ours and baselines for the wiping task.}
    \label{fig:failure_mode_wipe}
\end{figure}

\begin{figure}
    \centering
    \includegraphics[width=\linewidth]{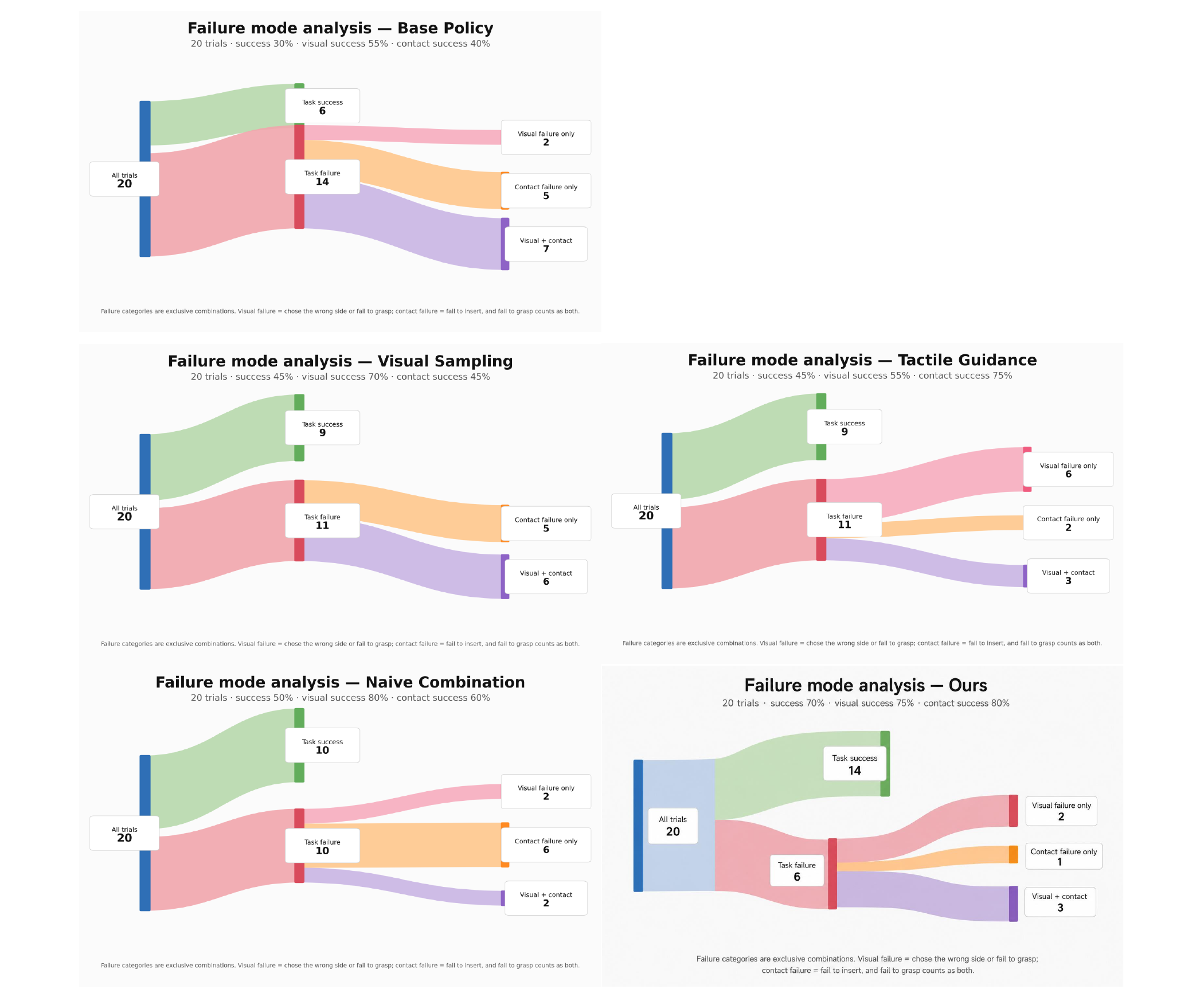}
    \caption{\textbf{Failure Mode Analysis For Insertion Task}. We demonstrate the different failure modes of \ours and baselines for the insertion task.}
    \label{fig:failure_mode_insert}
\end{figure}

\begin{figure}
    \centering
    \includegraphics[width=\linewidth]{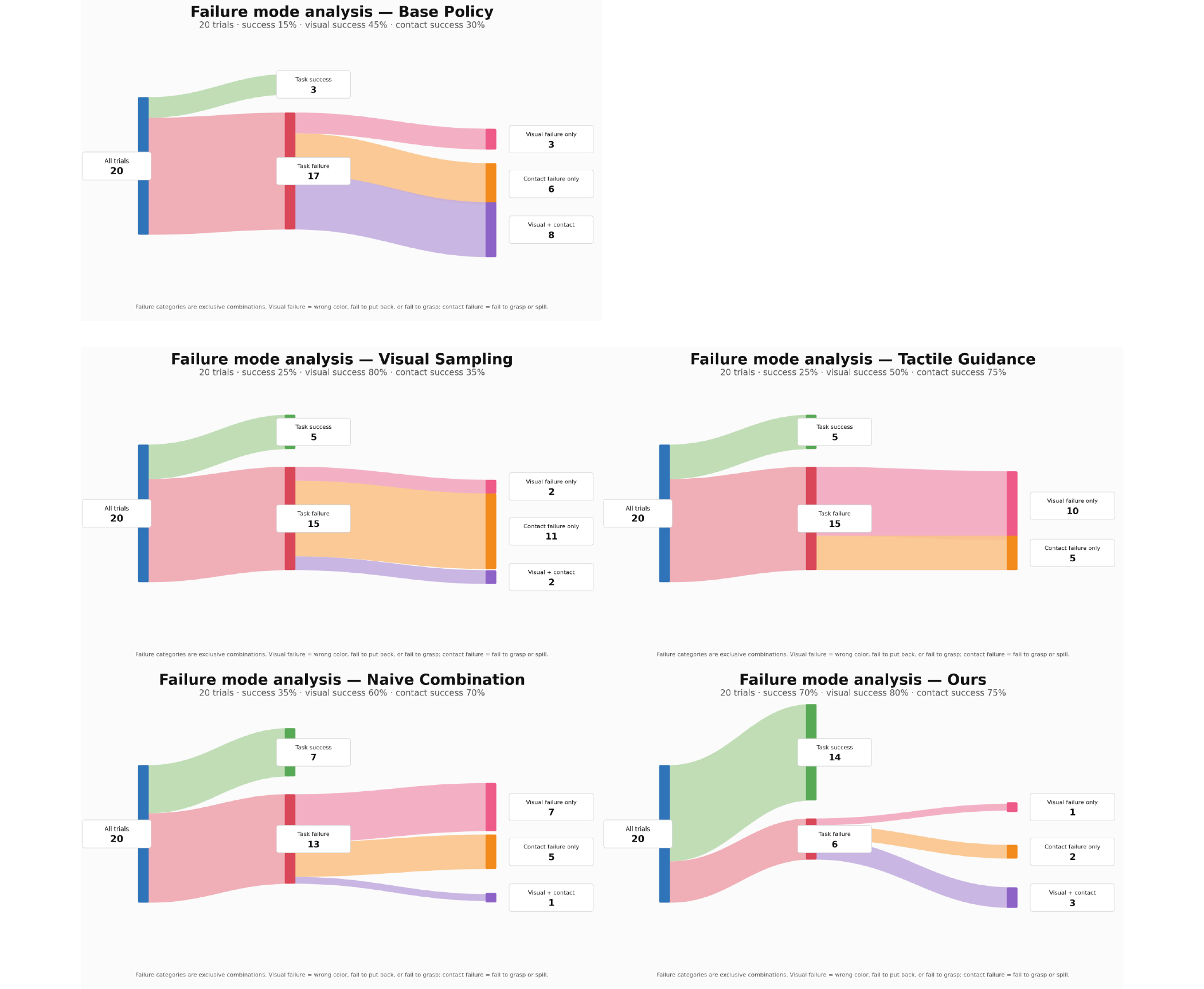}
    \caption{\textbf{Failure Mode Analysis For Pipette Transfer Task}. We demonstrate the different failure modes of \ours and baselines for the pipette transfer task.}
    \label{fig:failure_mode_pipette}
\end{figure}

\para{Failure Modes of Visual-Only Steering}
Visual-only steering primarily fails when visually promising futures do not correspond to physically reliable execution. For example, in the wiping task, visual lookahead may select an action sequence that reaches the correct mark but fails to apply sufficient pressure, resulting in incomplete erasing in Fig.~\ref{fig:multi_modal_wipe_all}. In the insertion task, the robot may move toward the correct hole but fail to insert the peg because small contact misalignments are difficult to resolve from visual predictions alone as shown in Fig.~\ref{fig:multi_modal_insert_all}. In the pipette transfer task, the robot may approach the correct target cup but grasp the dropper with too large force, causing spilling in Fig.~\ref{fig:multi_modal_pipette_all}. These failures show that visual success alone is insufficient for contact-rich tasks where local force, alignment, and grasp stability determine whether the task is completed. Detailed failure mode summaries of three tasks are in Fig.~\ref{fig:failure_mode_wipe}, Fig.~\ref{fig:failure_mode_insert} and Fig.~\ref{fig:failure_mode_pipette}.


\para{Failure Modes of Tactile-Only Steering}
Tactile-only steering exhibits the complementary limitation. Since tactile rewards emphasize local contact quality, tactile-only methods can improve grasping, pressing, or alignment behavior, but they do not reliably encode the global semantic target. In the wiping task, tactile-only guidance may produce correct contact between eraser and board but wipe the wrong region in Fig.~\ref{fig:multi_modal_wipe_all}. In the insertion task, it may encourage contact between the peg and tabletop but fail to select the instructed top-left or top-right hole as shown in Fig.~\ref{fig:multi_modal_insert_all}. In the pipette transfer task, it may improve grasp force on the dropper but move toward the wrong cup or fail to complete the return phase. These examples indicate that tactile feedback is most effective for short-horizon local refinement, but is insufficient for long-horizon semantic task selection in Fig.~\ref{fig:multi_modal_pipette_all}. Detailed failure mode summary of three tasks are in Fig.~\ref{fig:failure_mode_wipe}, Fig.~\ref{fig:failure_mode_insert} and Fig.~\ref{fig:failure_mode_pipette}.

\para{Summary}
The qualitative results reinforce the quantitative trends in Fig.~\ref{fig:policy_success_rate}. Visual-only steering is effective for choosing where to act, but not necessarily how to interact. Tactile-only steering improves how the robot interacts locally, but cannot reliably determine the correct semantic target. \ours{} combines the strengths of both modalities by using visual lookahead for global task progress and tactile diffusion editing for local contact refinement, leading to more reliable execution across diverse contact-rich manipulation tasks.

\subsection{Semantically-Aligned Vision and Tactile Verifier Enables Global and Local Adaptation}
\label{app:results_reward_world_model}
This section provides additional analysis for Sec.~\ref{sec:experiment_verifier}. Effective inference-time steering requires two properties: the world model must preserve reward-relevant information in its predicted futures, and the verifier must assign rewards that are semantically aligned with the task description. We therefore evaluate both the visual and tactile verifiers on ground-truth rollouts and world-model-predicted rollouts.

\para{Preference-Order Accuracy}
We evaluate reward quality using preference-order accuracy~\cite{tian2026position}. For each task and each modality, we construct pairs of rollouts with different human preference labels, such as successful versus failed executions or trajectories that better satisfy the language goal. Given a rollout $i$, we compute its normalized reward score by averaging the verifier reward over the rollout:
\begin{equation}
\hat{y}_i
=
\frac{1}{T_i}
\sum_{t=1}^{T_i}
R
\left(
    \hat{\latent}_{t}^{(i)};
    \lang
\right),
\label{eq:preference_reward_score}
\end{equation}
where $\hat{\latent}_{t}^{(i)}$ denotes either the ground-truth encoded observation or the world-model-predicted latent at timestep $t$, and $\lang$ is the corresponding visual or tactile language description. For each pair $(i,j)$ with different human preference labels $y_i \neq y_j$, we compare the human preference ordering with the reward-induced ordering:
\begin{equation}
s^{H}_{ij}
=
\mathrm{sign}(y_i-y_j),
\qquad
s^{R}_{ij}
=
\mathrm{sign}(\hat{y}_i-\hat{y}_j).
\label{eq:preference_signs}
\end{equation}
The preference-order accuracy is then:
\begin{equation}
A
=
\frac{1}{|\mathcal{P}|}
\sum_{(i,j)\in\mathcal{P}}
\mathbbm{1}
\left[
s^{H}_{ij}=s^{R}_{ij}
\right],
\label{eq:preference_order_accuracy}
\end{equation}
where $\mathcal{P}$ is the set of evaluated rollout pairs. Higher accuracy indicates that the verifier reward induces an ordering that better matches human judgments. We report results on both ground-truth observations (\textbf{GT}) and predicted futures (\textbf{Pred}) to measure whether the verifier remains reliable when applied to world-model imagination.

\para{Visual Verifier Performance}
Table~\ref{tab:reward_eval} shows that the adapted ROBOMETER visual verifier achieves strong preference-order accuracy across all tasks. On ground-truth observations, the visual reward obtains $80.0\%$ on wiping, $70.0\%$ on insertion, and $100.0\%$ on pipette transfer, with an average of $83.3\%$. On world-model predictions, the visual reward remains similarly reliable as $82.5\%$, $72.5\%$, and $100.0\%$, respectively, with an average of $85.0\%$.

These results indicate that the predicted visual futures preserve the semantic information needed by the VLM-based verifier. In particular, the high pipette accuracy suggests that the visual verifier can robustly distinguish whether the robot is moving toward or dispensing into the language-specified target cup. The insertion task is more challenging, likely because the ROBOMETER reward model still struggles to understand the spatial relationship of left or right compared to recognizing the color. Even in this case, the visual verifier achieves above-chance and task-useful preference accuracy, supporting its use for global mode selection rather than precise contact evaluation.

\para{Tactile Verifier Performance}
The tactile verifier also achieves strong preference-order accuracy across tasks. On ground-truth tactile observations, it obtains $85.0\%$ on wiping, $70.0\%$ on insertion, and $80.0\%$ on pipette transfer, with an average of $78.3\%$. On predicted tactile futures, the tactile reward obtains $90.0\%$, $77.5\%$, and $77.5\%$, respectively, improving the average to $81.7\%$.

These results show that the tactile verifier provides a meaningful contact-sensitive reward for both real and imagined tactile observations. The wiping task has the highest tactile accuracy, suggesting that the tactile embeddings reliably capture contact and pressure patterns. The insertion task achieves $77.5\%$ accuracy on predicted futures, indicating that the world model preserves contact cues relevant to peg-hole interaction. The pipette task remains similar accuracy between ground-truth ($80.0\%$) and predicted futures ($77.5\%$).

\para{Ground-Truth vs. Predicted Futures}
The results also demonstrate verifier accuracy on predicted futures is comparable to, and sometimes higher than, accuracy on ground-truth observations. This suggests that the visuo-tactile world model does not merely reconstruct observations at the pixel level, but preserves the semantic and contact-relevant structure required for reward evaluation. Predicted futures may also smooth high-frequency sensor noise and remove irrelevant variations in camera or tactile observations, making preference ordering easier for the verifier in some cases. This effect is visible in the improved average visual accuracy from $83.3\%$ on \textbf{GT} to $85.0\%$ on \textbf{Pred}, and tactile accuracy from $78.3\%$ on \textbf{GT} to $81.7\%$ on \textbf{Pred}.

\para{Qualitative Reward Behavior}
Fig.~\ref{fig:reward_qualitative} further illustrates the semantic alignment of the verifiers on the pipette transfer task. For visual verification, the reward depends on the language-conditioned task phase. Given the same trajectory where the robot approaches and dispenses toward the yellow cup, the verifier assigns high reward when the task description is ``transfer the liquid to the yellow cup,'' and assign lower reward when the actual task is to transfer to the blue cup. After phase switching, the visual objective changes to returning the dropper to the red cup, and the reward correspondingly shifts to favor predicted futures that move back toward the red cup.

For tactile verification, the reward captures local contact transitions. Early in the pipette task, the tactile description ``grasp lightly'' favors gentle contact suitable for approaching and positioning the dropper. Once the visual reward indicates sufficient task progress, the tactile description switches to ``grasp heavily,'' encouraging stronger contact for stable dispensing and return. The tactile reward trace aligns with marker-tracking force estimates, indicating that the language-conditioned tactile embedding captures meaningful changes in grasp force.

\para{Implications for Steering}
These results justify our use of semantically aligned verifiers in the bi-level steering framework. The visual verifier is reliable for long-horizon semantic evaluation, such as selecting the correct target mark, hole, or cup. The tactile verifier is reliable for short-horizon contact evaluation, such as determining whether the robot is applying sufficient pressure, maintaining stable grasp, or producing contact patterns consistent with insertion. Since both verifiers remain accurate on world-model predictions, \ours{} can evaluate candidate futures before execution and steer the policy toward actions that satisfy both global task goals and local contact requirements.

\subsection{Bi-level Optimization Balances Success and Steering Speed Better}
\label{app:results_bi_level}

This section provides additional qualitative and runtime analysis for Sec.~\ref{sec:experiment_bilevel}. The main experiments compare \ours{} to a naive multimodal combination baseline that sums normalized visual and tactile rewards across sampled candidates. Here, we further analyze why the bi-level structure is important.

\para{Why Naive Reward Fusion Fails}
A straightforward way to combine visual and tactile rewards is to normalize both rewards across sampled candidates and select the candidate with the highest summed score. However, this approach assumes that the two rewards are comparable over the same horizon and should be optimized simultaneously. In practice, this assumption is problematic. Visual rewards are most informative over longer horizons, where the robot's future state reveals whether it reaches the correct target. Tactile rewards are most informative over shorter horizons, where local contact changes indicate grasp force, insertion alignment, or wiping pressure. 

Because the two modalities operate on different temporal and semantic scales, naive reward fusion can select candidates that are suboptimal for full task execution. A candidate with high visual reward may dominate the combined score even if it has poor contact quality. Conversely, a candidate with strong tactile contact may be selected even if it does not make progress toward the correct visual goal. This reward imbalance leads to failures such as incomplete wiping, peg misalignment, or spilling in the pipette task, as shown in Fig.~\ref{fig:bilevel_qualitative}.

\para{Benefits of Bi-Level Steering}
Our bi-level formulation avoids this issue by assigning different roles to the two modalities. The inner visual stage performs long-horizon sampling and verification to select the global action mode:
\[
\bar{\action}_{t:t+H}
=
\arg\max_{\action_{t:t+H}}
R^v(\latentv_{t:t+H}; \lang_p^v).
\]
This stage determines the task-relevant target, such as which mark to wipe, which hole to insert into, or which cup to transfer liquid to. The outer tactile stage then refines the first $h$ steps of the selected sequence:
\[
\action^\star_{t:t+h}
=
\arg\max_{\tilde{\action}_{t:t+h}}
\log p_\theta
\left(
    \tilde{\action}_{t:t+h}
    \mid
    \bar{\action}_{t:t+h}, \obs_t
\right)
+
\beta R^\tau
\left(
    \tilde{\latent}^{\tau}_{t:t+h};
    \lang_p^\tau
\right).
\]
This keeps the final action close to the visually selected mode while allowing local edits that improve contact quality. As a result, \ours{} avoids optimizing a single mixed reward over a mismatched horizon.

\para{Qualitative Analysis}
Fig.~\ref{fig:bilevel_qualitative} shows representative examples where naive fusion fails but \ours{} succeeds. In wiping, naive fusion may move toward the correct mark but fail to maintain sufficient eraser-board contact, leaving marks partially visible. In insertion, naive fusion may choose the correct side but fail to align the peg with the hole during contact. In pipette transfer, naive fusion may approach the correct cup but spill liquid or fail to return the dropper. These failures arise because naive fusion does not explicitly separate global goal selection from local contact refinement.

By contrast, \ours{} first selects the correct global behavior using visual lookahead and then uses tactile diffusion editing to improve local physical execution. This produces actions that are both semantically correct and physically feasible.

\para{Inference-Time Trade-Off}
The bi-level design also provides a favorable efficiency trade-off as shown in Table~\ref{tab:inference_time}. Visual lookahead is performed through sampling and verification because direct gradient-based guidance through a VLM, decoder, recurrent world model, and diffusion policy would be expensive and unstable. Tactile refinement, however, is applied locally over the first $h$ actions, where classifier-based diffusion guidance is efficient and contact information is most relevant. As reported in Sec.~\ref{sec:experiment_bilevel}, \ours{} adds only a small inference-time overhead relative to naive fusion while achieving substantially higher success. This suggests that the performance improvement comes from the structure of the optimization, rather than simply from spending more computation.

\para{Summary}
The additional results show that bi-level steering is important for both effectiveness and efficiency. Rather than combining visual and tactile rewards in a single flat objective, \ours{} uses each modality at the temporal scale where it is most reliable: long-horizon vision for global task selection and short-horizon touch for local contact refinement.

\begin{table}[t]
\centering
\footnotesize
\setlength{\tabcolsep}{2.6pt}
\renewcommand{\arraystretch}{1.12}
\caption{
\textbf{Inference Time Analysis}. Runtime values are reported in milliseconds as mean $\pm$ standard deviation. The main bottleneck is visual reward inference from VLM-based ROBOMETER.
}
\label{tab:insert_runtime}
\resizebox{\linewidth}{!}{
\begin{tabular}{lcc|ccccc}
\toprule
\multirow{2}{*}{\textbf{Method}}
& \multicolumn{2}{c|}{\textbf{Configuration}}
& \multicolumn{5}{c}{\textbf{Runtime (ms)}} \\
\cmidrule(lr){2-3}
\cmidrule(lr){4-8}
& Batch Size $\mathbf{N}$
& Horizon $\mathbf{H}$
& \textbf{Sampling}
& \textbf{WM + Policy Interleaving}
& \textbf{Visual Verification}
& \textbf{Tactile Verification}
& \textbf{Total} \\
\midrule
Ours
& 10 & 16
& $69{\pm}3$
& $121{\pm}3$
& $216{\pm}10$
& $65{\pm}3$
& $471{\pm}12$ \\

Naive combination
& 10 & 16
& $69{\pm}3$
& $121{\pm}3$
& $216{\pm}10$
& $24{\pm}1$
& $429{\pm}13$ \\

Vision-only sampling (8-step)
& 10 & 8
& $69{\pm}3$
& $11{\pm}1$
& $216{\pm}10$
& --
& $295{\pm}11$ \\

Vision-only sampling (16-step)
& 10 & 16
& $69{\pm}3$
& $121{\pm}3$
& $216{\pm}10$
& --
& $405{\pm}12$ \\

Tactile-only sampling
& 10 & 8
& $69{\pm}3$
& $11{\pm}1$
& --
& $24{\pm}1$
& $103{\pm}4$ \\

Tactile-only guidance
& 1 & 8
& --
& --
& --
& $94{\pm}4$
& $94{\pm}4$ \\

Base policy
& 1 & --
& $3{\pm}1$
& --
& --
& --
& $3{\pm}1$ \\
\bottomrule
\end{tabular}
}
\vspace{0.3em}
\label{tab:inference_time}
\end{table}



\end{document}